\documentclass[11pt]{article}

\usepackage[final]{acl}

\usepackage{times}
\usepackage{latexsym}

\usepackage{hyperref}
\usepackage{url}
\usepackage{wrapfig}
\usepackage{booktabs}
\usepackage{subcaption}
\usepackage{graphicx}
\usepackage{multirow}
\usepackage{float}
\usepackage{algorithm}
\usepackage{algorithmic}
\usepackage{amsmath,amsfonts,bm}
\usepackage{xcolor}

\usepackage[T1]{fontenc}

\usepackage[utf8]{inputenc}

\usepackage{microtype}

\usepackage{inconsolata}

\usepackage{graphicx}

\title{Soft Head Selection for Injecting ICL-Derived Task Embeddings}

\author{
 \textbf{Jungwon Park$^{1,5}$},
 \textbf{Jimyeong Kim$^{2}$},
 \textbf{Changin Choi$^{3,6}$},
 \textbf{Wonjong Rhee$^{3,4}$\thanks{Corresponding Author.}}
\\
 $^1$RICS, $^2$AIIS, $^3$IPAI, $^4$Department of Intelligence and Information, 
\\
 Seoul National University
\\ 
 $^5$Daegu Gyeongbuk Institute of Science and Technology 
\\
 $^6$Samsung Advanced Institute of Technology, Samsung Electronics Co., Ltd
\\
 \small{ \texttt{\{quoded97, wlaud1001, ci2015.choi, wrhee\}@snu.ac.kr}}
}

\begin{document}
\addtocontents{toc}{\protect\setcounter{tocdepth}{-1}}
\maketitle

\begin{abstract} 
Large language models~(LLMs) are commonly adapted to downstream tasks using parameter-efficient fine-tuning~(PEFT) or in-context learning~(ICL). Recently, ICL-driven embedding-based adaptation has been proposed as a distinct task adaptation paradigm. It derives task-specific embeddings from intermediate activations using few-shot prompts and injects them during inference. Despite its conceptual appeal, this approach has not demonstrated consistent performance gains over PEFT or ICL, and its empirical advantages have been limited in practice. We propose Soft head-selection for ICL-derived Task Embeddings~(SITE), a gradient-based method that identifies task-relevant attention heads to enable effective task embedding injection. Across various types of open-ended generation, reasoning, and natural language understanding tasks, SITE significantly outperforms prior embedding-based adaptation methods and few-shot ICL, while using substantially fewer trainable parameters than PEFT. Experiments on 12 LLMs ranging from 4B to 70B parameters demonstrate the generality of our approach, and intra-task and inter-task activation patching analyses further provide new mechanistic insights by revealing strong task dependence in attention head functionality. \footnote{Our code is available at \url{https://github.com/SNU-DRL/Soft_Injection}}
\end{abstract}
\section{Introduction}

In recent years, large language models~(LLMs) have demonstrated remarkable generation capabilities across a wide range of domains, along with the ability to rapidly adapt to new tasks. Traditionally, such task adaptation has been achieved through parameter-efficient fine-tuning~(PEFT)~\citep{houlsby2019parameter, lester2021power, hu2022lora} or in-context learning~(ICL)~\citep{brown2020language}. PEFT typically yields strong task performance but requires training, whereas ICL enables training-free and flexible adaptation by incorporating input-output demonstrations into the prompt, at the cost of increased prompt length and inference overhead.

Recent studies have shown that intermediate activations of LLMs elicited by few-shot ICL encode rich task-relevant information~\citep{hendel2023context, todd2023function}. These activations can be extracted and later injected into the model, enabling task execution without explicit instructions or demonstrations in the prompt. This line of work, referred to as \emph{ICL-driven embedding-based adaptation}~(or \emph{ICL-driven activation steering}), introduces a new mechanism for conveying task information to LLMs. Most existing approaches~\citep{hendel2023context, todd2023function, zhang2024batch, huang2024multimodal, li2024implicit, wang2024elicit, cai2025beyond, liuiterative} extract \emph{task embeddings}, which are assumed to encode task-relevant information, from last-token activations at selected layers or modules of the model, and focus on designing heuristics for where and how to extract and inject them. However, despite their conceptual appeal, existing methods have not yet demonstrated clear advantages over PEFT or ICL in terms of either adaptation efficiency or task performance. The goal of this work is to develop an ICL-driven embedding-based adaptation method that achieves clear empirical advantages over these two alternatives.

\begin{figure}[t]
\begin{center}
    \includegraphics[width=0.85\columnwidth]{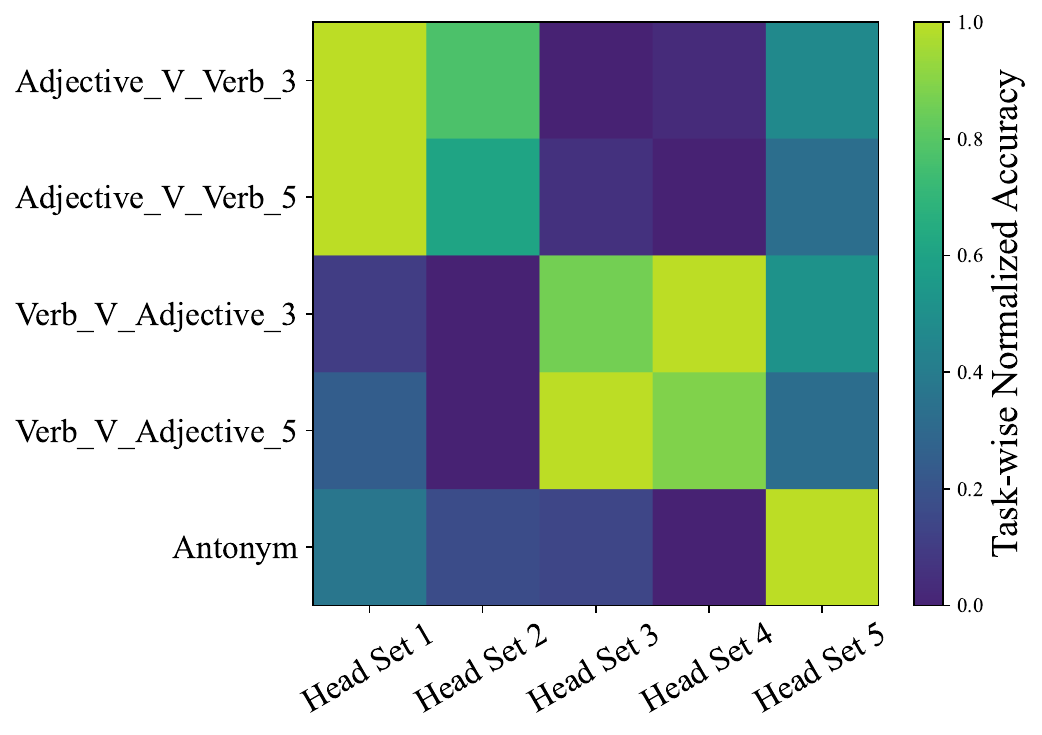}
\end{center}
\vspace{-3mm}
\caption{\textbf{Random head patching.} During zero-shot inference, activations of randomly selected attention heads are replaced~(or patched) with corresponding activations extracted from 10-shot inference, illustrating the sensitivity of task performance to the patched heads.
}
\vspace{-2.5mm}
\label{fig:random_head_patching}
\end{figure}

Attention head attribution~\citep{hao2021self, olsson2022context, todd2023function, gandelsman2023interpreting, park2024cross, zhou2024role, wu2024retrieval, elhelo2024inferring}, a line of research in mechanistic interpretability, studies the functional roles of individual attention heads in deep neural networks, including retrieval, safety, and in-context learning behaviors.  While prior studies have identified attention heads associated with in-context learning, it remains largely unexplored whether the importance of attention heads varies across tasks. In our preliminary activation patching~\citep{zhang2023towards, hendel2023context, todd2023function, bereska2024mechanistic} experiment shown in Figure~\ref{fig:random_head_patching}, where randomly selected attention head activations during zero-shot inference are replaced with those extracted from few-shot inference, we observe that task performance is highly sensitive to which heads are patched. This sensitivity suggests that task-relevant information may be unevenly distributed across attention heads and may vary substantially across tasks.

Motivated by this observation, we propose Soft head-selection for ICL-derived Task Embeddings~(SITE). SITE identifies task-relevant attention heads by formulating head selection as a continuous optimization problem, enabling efficient learning of head importance via gradient descent algorithm. For each task, our method constructs task embeddings by extracting last-token attention head activations from few-shot prompts, and learns soft head-selection parameters that linearly interpolate between the original head activations and the task embeddings. Using zero-shot prompts at inference time, SITE achieves strong performance across various open-ended generation, complex reasoning, and natural language understanding tasks, significantly outperforming prior embedding-based adaptation methods as well as few-shot ICL, while using substantially fewer trainable parameters than PEFT. We evaluate SITE on 12 LLMs spanning 4 model families, 3 model variants, and sizes ranging from 4B to 70B parameters, demonstrating its broad applicability.

Furthermore, we extend activation patching experiments using our head-selection parameters to further investigate task dependence in attention head functionality. In intra-task activation patching, we show that the selected heads effectively capture task-relevant information. In inter-task activation patching, we find that similar tasks tend to share important attention heads, whereas dissimilar tasks do not, indicating strong task specificity in head roles. These findings extend prior studies on the attention head functionalities from a task-agnostic perspective to a task-specific one. 

Overall, our contributions are summarized as follows:
\begin{itemize}
\item We propose SITE, an ICL-driven embedding-based adaptation method that significantly outperforms few-shot ICL and approaches PEFT-level performance, while requiring substantially fewer trainable parameters.
\item We demonstrate that soft head selection enables effective identification and utilization of task-relevant attention heads across diverse tasks and models.
\item We provide new mechanistic insights into task-specific functional roles of attention heads, showing that \emph{head importance varies across tasks and reflects task similarity}.
\end{itemize}


\section{Related work}
\label{sec:related_work}
\paragraph{Task adaptation of LLMs.} Task adaptation methods for LLMs can be broadly categorized into prompt-based, weight-based, and embedding-based approaches, according to where task-relevant information is encoded. Prompt-based methods~\citep{zhou2022large, liu2022makes, lu2022fantastically, yang2023large} include in-context learning and instruction optimization, which improve task performance by selecting, ordering, or designing input-output exemplars or task instructions within the prompt. Weight-based approaches comprise weight-based PEFT techniques such as Adapters~\citep{houlsby2019parameter}, LoRA~\citep{hu2022lora}, and (IA)$^3$~\citep{liu2022few}, which adapt models by introducing lightweight trainable modules or updating a small subset of model weights. Embedding-based methods~\citep{lester2021power, li2021prefix, li2023inference, hendel2023context, todd2023function} inject task-specific embeddings, which are substantially smaller than PEFT modules, into intermediate activations, enabling task execution without using in-prompt instructions or demonstrations at inference time.

\begin{figure*}[t]
\begin{center}
    \includegraphics[width=\linewidth]{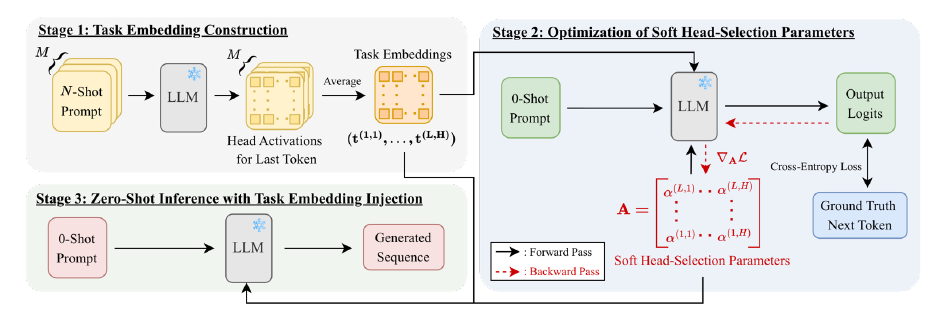}
\end{center}
\vspace{-3mm}
\caption{\textbf{Method Overview.} Our method consists of three stages: (1) constructing task embeddings by averaging last-token attention head activations across $M$ few-shot ICL prompts with $N$ input-output pairs; (2) optimizing soft head-selection parameters to identify where task embeddings should be injected during zero-shot inference; and (3) applying the task embeddings and learned selection parameters at inference time to perform tasks without in-prompt examples. $L$ and $H$ denote the number of attention layers and attention heads per layer, respectively, in the LLM. Stages 1 and 2 are executed once per task as precomputation, and the resulting task embeddings and soft head-selection parameters are then repeatedly used during zero-shot inference in Stage 3.
}
\label{fig:method_overview}
\end{figure*}

\paragraph{ICL-driven embedding-based adaptation.} A class of embedding-based adaptation methods~\citep{lester2021power, peng2024live, saglam2025learning, li2025m2iv, kang2025adaptive, li2025towards, yang2025task} directly optimizes task embeddings via gradient-based training, but these approaches typically require tens of thousands of optimization steps and often suffer from training instability. In contrast, ICL-driven embedding-based methods~\citep{hendel2023context, todd2023function, zhang2024batch, huang2024multimodal, li2024implicit, wang2024elicit, cai2025beyond, liuiterative} avoid tuning task embeddings and instead focus on identifying effective locations for extracting and injecting ICL-drived task embeddings. Early methods such as FV~\citep{todd2023function} and TV~\citep{hendel2023context} rely on layer- or head-wise activation patching, which incurs substantial search overhead. More recent approaches, including MTV~\citep{huang2024multimodal} and I2CL~\citep{li2024implicit}, reduce this cost through more efficient optimization strategies, but still rely on reinforcement learning-based optimization~\citep{williams1992simple} or restricted search spaces, and often fail to consistently outperform few-shot ICL. Moreover, most prior work focuses on simple classification tasks and omits comparisons with PEFT, limiting rigorous empirical evaluation. In contrast, our method identifies task embedding injection locations at the level of individual attention heads using gradient-based soft head-selection, and achieves performance comparable to PEFT across diverse task types.

\section{Method}
\label{sec:method}
Our method consists of three stages. (1) First, we construct task embeddings using few-shot ICL prompts. (2) Next, we optimize soft head-selection parameters via gradient descent. (3) Finally, we apply the pre-computed task embeddings and the learned head-selection parameters at inference time to enable task execution without in-prompt examples. Stages 1 and 2 are performed once per task, while Stage 3 is executed for each query within the task. Figure~\ref{fig:method_overview} provides an overview of our method.

\paragraph{Stage 1: Task embedding construction.}
Let $P_1, P_2, \dots, P_M$ denote the $M$ few-shot ICL prompts, each containing $N$ input-output examples sampled from the training set. For each prompt $P_m$, we perform a forward pass through the model and extract the output of every attention head at every layer. Specifically, for head $h \in \{ 1, 2, \dots, H \}$ in layer $l \in \{ 1, 2, \dots, L \}$, the output of head $h$ at layer $l$ for prompt $P_m$ is given by:
\begin{equation}
\begin{aligned}
\mathbf{t}_m^{(l, h)} 
= \text{softmax} \left( \frac{ \mathbf{Q}_m^{(l, h)} ( \mathbf{K}_m^{(l, h)} )^T }{\sqrt{d_k}}  \right) \mathbf{V}_m^{(l, h)}& \\
\in \mathbb{R}^{S_m \times d_v}, &
\end{aligned}
\end{equation}
where $\mathbf{Q}_m^{(l, h)}, \mathbf{K}_m^{(l, h)}, \mathbf{V}_m^{(l, h)}$ are the query, key, and value matrices, $d_k$ is the key/query dimension, $d_v$ is the value dimension, and $S_m$ is the number of tokens in the tokenized sequence of prompt $P_m$. 
We extract the activations corresponding to the last token and average them across all $M$ prompts to obtain a task embedding for each head:
\begin{equation}
\mathbf{t}^{(l, h)} = \frac{1}{M} \sum_{m = 1}^M \mathbf{t}_m^{(l, h)} [-1, :] \in \mathbb{R}^{d_v}
\end{equation}
The collection $\{ \mathbf{t}^{(l, h)} \}_{l=1, \dots, L;\, h=1, \dots, H}$ constitutes the task embeddings for the given task.

\paragraph{Stage 2: Optimization of soft head-selection parameters.}
To effectively identify task-relevant attention heads, we formulate head selection as a continuous optimization problem. For each task, we introduce a learnable matrix $ \mathbf{A} $, where each entry $ \alpha^{(l, h)}$ serves as a \emph{soft head-selection parameter} for attention head $h \in \{ 1, 2, \dots, H \}$ in layer $l \in \{ 1, 2, \dots, L \}$:
\begin{equation}
\mathbf{A} = \begin{bmatrix}
\alpha^{(L,1)} & \cdots & \alpha^{(L,H)} \\
\vdots & \ddots & \vdots \\
\alpha^{(1,1)} & \cdots & \alpha^{(1,H)}
\end{bmatrix} \in [0, 1]^{L \times H}
\end{equation}
Each $\alpha^{(l, h)}$ controls the degree to which task-specific information is injected into the corresponding attention head. Let $\mathbf{o}^{(l,h)} \in \mathbb{R}^{d_v}$ denote the original last-token activation of head $h$ in layer $l$ during optimization. We inject task embeddings by linearly interpolating between the original activation $\mathbf{o}^{(l,h)}$ and the task embedding $\mathbf{t}^{(l,h)}$:
\begin{equation}
\label{eqn:soft injection}
\mathbf{o}^{(l,h)} \leftarrow (1 - \alpha^{(l, h)}) \cdot \mathbf{o}^{(l,h)} + \alpha^{(l, h)} \cdot \mathbf{t}^{(l, h)},
\end{equation}
for all $l \in \{ 1, 2, \dots, L \}$ and $h \in \{ 1, 2, \dots, H \}$. During optimization, the LLM is kept frozen and only $\mathbf{A}$ is updated. We optimize $\mathbf{A}$ for a few hundred gradient descent steps by minimizing the cross-entropy loss for next-token prediction, computed from the output logits of the modified forward passes. At each optimization step, inference is performed using a zero-shot prompt, with task embeddings injected according to the current values of $\mathbf{A}$. Each $\alpha^{(l, h)}$ is parameterized as the sigmoid of an unconstrained scalar to ensure the values in $[0,1]$. The pseudocode for this procedure is provided in Algorithm~\ref{algorithm:optimization} of Appendix~\ref{subsec:implementation details for FV, MTV, and SITE}.

\paragraph{Stage 3: Zero-shot inference with task embedding injection.}
After optimization, we apply the learned soft head-selection parameters $ \mathbf{A} $ with the task embeddings $\{ \mathbf{t}^{(l, h)} \}_{l, h}$ to guide the LLM in performing tasks without in-prompt exemplars. The soft injection is applied in the same manner as during the optimization stage~(Eq.~(\ref{eqn:soft injection})) but only once, at the last token of the initial input prompt, assuming KV caching~\citep{pope2023efficiently} is enabled during autoregressive decoding. No further interventions are applied during subsequent decoding steps. This single-step injection embeds task-relevant information into the KV cache at the start of generation, allowing the model to produce the remaining tokens without additional intervention. While some prior methods inject task embeddings at multiple token positions~\citep{huang2024multimodal, li2024implicit}, we intervene only at a single-token activation to reduce intervention complexity and avoid excessive undesirable steering during text generation.

\section{Experiments}
\label{sec:experiments}

\subsection{Experimental setup}
\label{sub_sec:Experimental setup}

\paragraph{Models.} We use Llama-3.1-8B to compare multiple methods across a wide range of tasks, and apply our method on an additional 11 LLMs to assess its generality. In total, the 12 LLMs used in this study span 4 model families of Llama 3.1~\citep{grattafiori2024llama}, Mistral~\citep{jiang2023mistral7b, jiang2024mixtral}, Qwen3~\citep{yang2025qwen3}, and Gemma-3~\citep{team2025gemma}, covering 3 variation types and sizes ranging from 4B to 70B parameters. Detailed information about the 12 LLMs is provided in Table~\ref{tab:Experiments_model_list} of Appendix~\ref{subsec:12 models used for evaluation}.

\paragraph{Tasks.} We evaluate our method on 57 ICL tasks from the official Function Vectors~(FV)~\citep{todd2023function} repository, as well as three additional benchmarks: ANLI~\citep{nie2020adversarial}, MMLU-Pro~\citep{wang2024mmlu}, and Big-Bench Hard (BBH)~\citep{suzgun2023challenging}. The FV benchmark comprises 29 abstractive and 28 extractive open-ended generation tasks spanning diverse problem types, including closely related and contrasting variants, making it well suited for studying in-context learning behaviors. ANLI is a natural language inference benchmark that is more difficult than earlier datasets of SNLI~\citep{bowman2015large} and MNLI~\citep{williams2018broad}. MMLU-Pro and BBH consist of tasks requiring diverse and complex reasoning, where MMLU-Pro extends the original MMLU~\citep{hendrycks2020measuring} with more challenging and realistic question sets.

\begin{table*}[]
\resizebox{\textwidth}{!}{%
\begin{tabular}{ll|c|rrrr|c}
\toprule
\multicolumn{1}{l}{Category}     & \multicolumn{1}{l}{Method} & \multicolumn{1}{|c|}{\begin{tabular}{@{}c@{}}\# Trainable \\ Parameters\end{tabular}} & \multicolumn{1}{c}{\begin{tabular}{@{}c@{}} FV Benchmark \\ (57 tasks) \end{tabular}} & \multicolumn{1}{c}{\begin{tabular}{@{}c@{}} ANLI \\ (3 tasks) \end{tabular}} & \multicolumn{1}{c}{\begin{tabular}{@{}c@{}} MMLU-Pro \\ (14 tasks) \end{tabular}} & \multicolumn{1}{c|}{\begin{tabular}{@{}c@{}} Big-Bench Hard \\ (27 tasks) \end{tabular}} & \multicolumn{1}{c}{Average} \\
\midrule
\midrule
\multirow{2}{*}{Parameter-based} & LoRA          & 3407.87K                                                          & \underline{86.76} ± (0.24)                                                                        & 45.82 ± (1.20)                                                               & \underline{41.04} ± (0.34)                                                                    & \underline{60.39} ± (0.48)                                                                          & \underline{58.50}   \\
                                 & (IA)$^3$      & 524.29K                                                           & 81.97 ± (0.43)                                                                        & \underline{47.03} ± (0.14)                                                               & 40.87 ± (0.42)                                                                    & 60.29 ± (0.70)                                                                          & 57.54   \\
\midrule
\multirow{2}{*}{ICL}             & 0-shot        & -                                                                 & 9.85 ± (0.00)                                                                         & 0.00 ± (0.00)                                                                & 17.96 ± (0.00)                                                                    & 16.41 ± (0.00)                                                                          & 11.06   \\
                                 & 10-shot       & -                                                                 & \underline{76.76} ± (0.09)                                                                        & \underline{43.96} ± (0.34)                                                               & \underline{36.47} ± (0.48)                                                                    & \underline{47.17} ± (0.98)                                                                          & \underline{51.09}   \\
\midrule
\multirow{9}{*}{Embedding-based} & Prompt Tuning & 81.92K                                                            & 75.06 ± (0.35)                                                                        & 33.22 ± (0.13)                                                               & 10.60 ± (0.66)                                                                    & 33.45 ± (0.81)                                                                          & 38.13   \\
                                 & FV            & -                                                                 & 33.16 ± (0.08)                                                                        & 0.00 ± (0.00)                                                                & 1.14 ± (0.04)                                                                     & 17.82 ± (0.37)                                                                          & 13.03   \\
                                 & TV            & -                                                                 & 59.07 ± (0.83)                                                                        & 33.17 ± (0.04)                                                               & 31.60 ± (0.25)                                                                    & 42.01 ± (0.71)                                                                          & 41.46   \\
                                 & MTV           & 1.02K                                                             & 73.24 ± (1.46)                                                                        & 34.70 ± (0.88)                                                               & 34.07 ± (0.88)                                                                    & 42.54 ± (0.71)                                                                          & 46.14   \\
                                 & LIVE          & 131.10K                                                           & 23.34 ± (1.03)                                                                        & 0.00 ± (0.00)                                                                & 20.51 ± (2.07)                                                                    & 12.89 ± (2.32)                                                                          & 14.19   \\
                                 & I2CL          & 0.13K                                                             & 79.89 ± (0.56)                                                                        & 28.01 ± (3.94)                                                               & 27.14 ± (0.32)                                                                    & 50.60 ± (1.12)                                                                          & 46.41   \\
                                 & IV            & -                                                                 & 42.52 ± (0.13)                                                                        & 31.52 ± (0.44)                                                               & 9.84 ± (0.51)                                                                     & 26.25 ± (1.53)                                                                          & 27.53   \\
                                 & Ours ($M=1$)  & 1.02K                                                             & 89.67 ± (0.60)                                                                        & 46.35 ± (0.51)                                                               & 37.24 ± (0.84)                                                                    & 56.76 ± (1.42)                                                                          & 57.50   \\
                                 & Ours ($M=50$) & 1.02K                                                             & \textbf{90.02} ± (0.19)                                                                        & \textbf{47.31} ± (0.15)                                                               & \textbf{38.78} ± (0.41)                                                                    & \textbf{58.04} ± (0.72)                                                                          & \textbf{58.54}  \\
\bottomrule
\end{tabular}%
}
\caption{\textbf{Comparison on four benchmarks using Llama-3.1-8B.} We compare our method~(SITE) with parameter-based methods, in-context learning~(ICL), and prior embedding-based approaches. The number of trainable parameters is reported where applicable. Best embedding-based results are shown in \textbf{bold}, and best parameter-based and ICL results are \underline{underlined}.}
\vspace{-2mm}
\label{tab:main_results_llama_3.1_8b_new}
\end{table*}

\paragraph{Implementation details.} For each task, the dataset is split into training, validation, and test sets following the split ratio used in FV; only the training and validation sets are used to construct task embeddings and optimize soft head-selection parameters. Unless otherwise specified, task embeddings are constructed using $M = 50$ few-shot prompts, each containing $N = 10$ input-output exemplars that are randomly sampled from the training set, with possible overlaps across prompts. The soft head-selection parameters are initialized to 0.5 and optimized using the Adam~\citep{kingma2014adam} optimizer with a learning rate of 0.2 for 400 iterations, without regularization or model-specific hyperparameter tuning, across all 12 LLMs. Checkpoints are selected based on validation loss, evaluated every 50 iterations. We use greedy decoding to minimize randomness. Implementation details for other methods are provided in Appendix~\ref{subsec:implementation details for FV, MTV, and SITE}.

\subsection{Experimental results}
\label{subsec:Experimental results}
Across four benchmarks, we compare our method with parameter-based methods~(LoRA~\citep{hu2022lora}, (IA)$^3$~\citep{liu2022few}), in-context learning~(ICL), and prior embedding-based approaches~(Prompt Tuning~\citep{lester2021power}, FV~\citep{todd2023function}, TV~\citep{hendel2023context}, MTV~\citep{huang2024multimodal}, LIVE~\citep{peng2024live}, I2CL~\citep{li2024implicit}, IV~\citep{liuiterative}). We report results of our method using two values of $M$~($1$ and $50$) to demonstrate the robustness of our method to the number of prompts used for task embedding construction. Table~\ref{tab:main_results_llama_3.1_8b_new} reports the mean accuracy and standard deviation over three random seeds using Llama-3.1-8B.

\begin{figure}[t]
\begin{center}
    \includegraphics[width=0.99\linewidth]{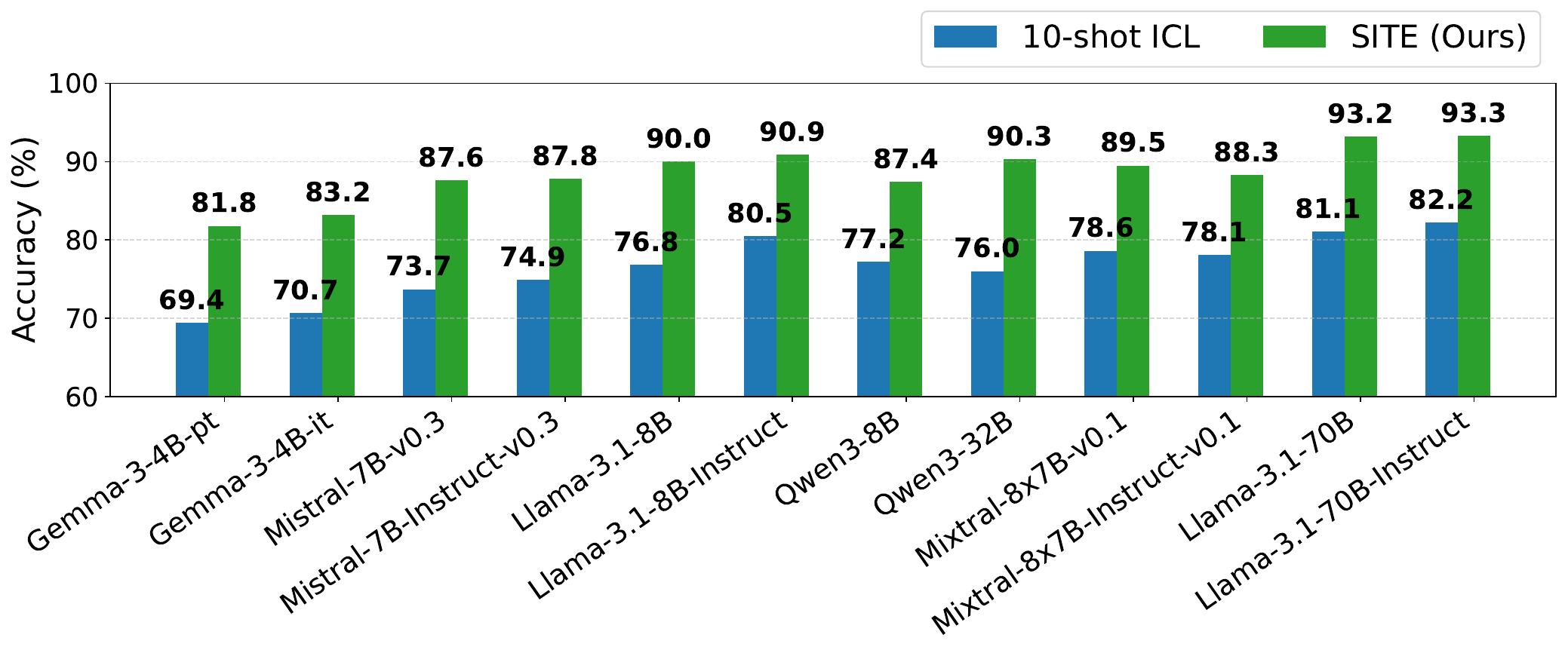}
\end{center}
\vspace{-4mm}
\caption{\textbf{Average performance on the FV benchmark for 12 backbone LLMs.} We report results for 10-shot ICL and our method, with average accuracies annotated above each bar.
}
\label{fig:performance_of_various_models}
\vspace{-3mm}
\end{figure}

Overall, our method significantly outperforms all embedding-based baselines and ICL across benchmarks. For example, SITE with $M=50$ outperforms 10-shot ICL by 13.26\% on the FV benchmark and by 7.45\% on average across all four benchmarks. Compared to parameter-based methods such as LoRA and (IA)$^3$, our method achieves higher performance on the FV benchmark and ANLI, and comparable results on MMLU-Pro and BBH. These two latter benchmarks require complex reasoning or expert-level knowledge, suggesting that ICL-derived task embeddings may have limited representational capacity or require additional adaptation for such complex tasks. Notably, our method achieves these results with substantially fewer trainable parameters than PEFT, as it freezes both the backbone LLM and the task embeddings and optimizes only the injection locations. We further observe that increasing $M$ slightly improves performance by reducing instance-specific noise while preserving task-level information, resulting in more stable head-selection optimization.

To evaluate generality across backbone LLMs, we further apply our method to 12 LLMs spanning four model families, three model variants, and sizes ranging from 4B to 70B parameters. Figure~\ref{fig:performance_of_various_models} presents results on the 57 FV tasks, comparing our method with 10-shot ICL. Across all models, our method achieves an average improvement of 10.2-14.3\% over 10-shot ICL, demonstrating strong generalization across diverse LLMs. Task-level results for all 12 models are reported in Tables~\ref{tab:main_results_gemma-3-4b-pt}-\ref{tab:main_results_llama-3.1-70b-instruct} of Appendix~\ref{sec:task-wise performance across 11 additional LLMs}.

\section{Activation patching analysis with binarized head-selection parameters}
\label{sec:analysis}
Activation patching~\citep{meng2022locating, zhang2023towards, hendel2023context, todd2023function, bereska2024mechanistic} is a widely used technique in mechanistic interpretability for analyzing the functional roles of model components. By replacing the activations of specific components with alternative representations, it enables direct assessment of their causal contributions through changes in model outputs. In this section, we extend the activation patching experiment introduced in Figure~\ref{fig:random_head_patching} by leveraging binarized head-selection parameters.

As shown in Figure~\ref{fig:analysis_optimized_head_selection_line_plot}, the learned soft head-selection parameters exhibit near-binary patterns, motivating the use of their binarized counterparts for activation patching. Using these binarized parameters, we conduct two types of activation patching experiments. In both settings, we patch selected attention head activations during zero-shot inference using task embeddings extracted from 10-shot prompts. The two experiments differ in whether the head-selection parameters are derived from the same task as zero-shot inference or from a different task. Through these experiments, we examine whether the heads selected by our method capture task-relevant information and whether attention head functionality exhibits task-specific property, such that similar tasks share important heads while dissimilar tasks do not. While the analyses in this section are based on Llama-3.1-8B, the findings generalize to other larger LLMs, as demonstrated in Appendix~\ref{subsec:Optimized soft head-selection values for larger language models} and Appendix~\ref{subsec:cross-task results for larger language models}.

\subsection{Intra-task activation patching}
\label{subsec:identification of task-relevant attention heads}
\begin{figure}[t]
\begin{center}
    \includegraphics[width=\linewidth]{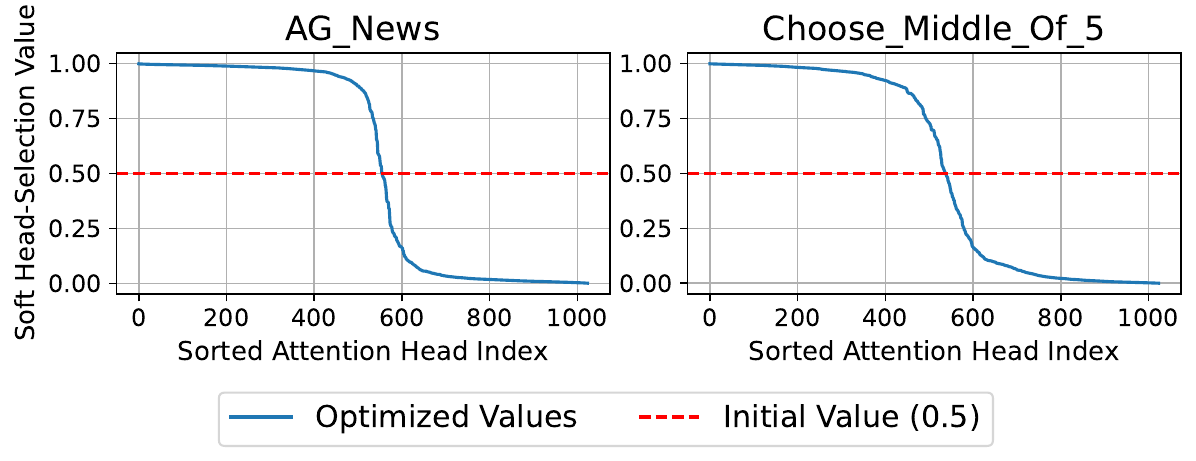}
\end{center}
\vspace{-3mm}
\caption{\textbf{Optimized values of the soft head-selection parameters for two FV tasks.} Each plot shows the optimized soft head-selection values for all 1024 attention heads in Llama-3.1-8B, sorted in descending order. Dashed lines indicate the initial value of 0.5 assigned to all soft head-selection parameters at the start of training. Results for all 57 tasks in the FV benchmark are provided in Figures~\ref{fig:analysis_head_visualization_line_plot_grid_1}-\ref{fig:analysis_head_visualization_line_plot_grid_3} of Appendix~\ref{subsec:Optimized soft head-selection values for all 57 tasks}.
}
\label{fig:analysis_optimized_head_selection_line_plot}
\end{figure}

In intra-task activation patching, the task embeddings~($\mathbf{t}^{(l, h)}$), the head-selection parameters~($\alpha^{(l, h)}$), and the zero-shot inference input are all derived from the same task. We compare patching high-$\alpha$ heads~($\alpha^{(l, h)} > 0.5$) with patching low-$\alpha$ heads~($\alpha^{(l, h)} \leq 0.5$) to assess whether the heads selected by our method capture task-relevant information. High-$\alpha$ heads correspond to attention heads assigned larger interpolation weights by the learned head-selection parameters~(see Eq.~(\ref{eqn:soft injection})), reflecting the heads prioritized by our method for task embedding injection. As shown in Figure~~\ref{fig:analysis_optimized_head_selection_line_plot} and Figures~\ref{fig:analysis_head_visualization_line_plot_grid_1}-\ref{fig:analysis_head_visualization_line_plot_grid_3} in Appendix~\ref{subsec:Optimized soft head-selection values for all 57 tasks}, the numbers of high-$\alpha$ and low-$\alpha$ heads are comparable across tasks, each accounting for approximately half of the total attention heads. 

Table~\ref{tab:intra-task analysis} reports the zero-shot performance~(before activation patching), the performance after patching activations of high-$\alpha$ heads, and the performance after patching activations of low-$\alpha$ heads. Patching high-$\alpha$ head activations leads to substantial accuracy improvements across all four benchmarks. In contrast, patching low-$\alpha$ head activations consistently degrades performance, even though the task embeddings are derived from the same 10-shot prompts and differ only in which attention heads are patched. These results indicate that the learned head-selection parameters effectively identify task-relevant attention heads.


\definecolor{darkgreen}{RGB}{0,100,0}
\definecolor{darkred}{RGB}{180,0,0}
\begin{table}[]
\resizebox{\columnwidth}{!}{%
\begin{tabular}{l|ccc}
\toprule
Benchmark     &  \begin{tabular}[c]{@{}c@{}}Baseline  \\ (0-shot)\end{tabular} & \begin{tabular}[c]{@{}c@{}}High-$\alpha$ Heads  \\ ($\alpha^{(l, h)} > 0.5$)\end{tabular} & \begin{tabular}[c]{@{}c@{}}Low-$\alpha$ Heads  \\ ($\alpha^{(l, h)} \leq 0.5$)\end{tabular} \\
\midrule
FV Benchmark   & 9.9  & 88.9~(\textcolor{darkgreen}{+79.0}) & 3.7~(\textcolor{darkred}{-6.2})   \\
ANLI           & 0.0  & 46.9~(\textcolor{darkgreen}{+46.9}) & 0.0~(\textcolor{darkred}{-0.0})   \\
MMLU-Pro       & 18.0 & 37.8~(\textcolor{darkgreen}{+19.8}) & 7.5~(\textcolor{darkred}{-10.5})   \\
Big-Bench Hard & 16.4 & 55.7~(\textcolor{darkgreen}{+39.3}) & 13.7~(\textcolor{darkred}{-2.7})   \\ \midrule
Average        & 11.1 & 57.3~(\textcolor{darkgreen}{+46.2}) & 6.2~(\textcolor{darkred}{-4.9})   \\
\bottomrule
\end{tabular}%
}
\caption{\textbf{Intra-task activation patching analysis}. Comparison of 0-shot baseline performance with results after patching high-$\alpha$ and low-$\alpha$ attention heads, where the task embeddings and head-selection parameters derived from the same task.}
\label{tab:intra-task analysis}
\end{table}


\begin{table*}[ht]
\centering
\resizebox{0.9\textwidth}{!}{%
\begin{tabular}{l|l|l|l}
\toprule
\textbf{Evaluation Task}          & \textbf{Task Description}                                                                                                     & \textbf{\begin{tabular}[c]{@{}r@{}}Top-3 Head-Selection Tasks\\~(Accuracy, \%) \end{tabular}}                                                                                                     & \textbf{\begin{tabular}[c]{@{}r@{}}Bottom-3 Head-Selection Tasks\\~(Accuracy, \%) \end{tabular}}                                                                                               \\ \toprule
Adjective\_V\_Verb\_5    & \begin{tabular}[c]{@{}l@{}}Select the only adjective \\ from a list of 5 words \\ (1 adjective, 4 verbs)\end{tabular}       & \begin{tabular}[c]{@{}l@{}}Adjective\_V\_Verb\_3 (96.7)\\ Adjective\_V\_Verb\_5 (94.3)\\ Animal\_V\_Object\_5 (78.6)\end{tabular}  & \begin{tabular}[c]{@{}l@{}}Verb\_V\_Adjective\_3 (3.3)\\ Verb\_V\_Adjective\_5 (4.8)\\ English\_French (5.7)\end{tabular}                   \\
\midrule
Verb\_V\_Adjective\_5    & \begin{tabular}[c]{@{}l@{}}Select the only verb \\ from a list of 5 words \\ (1 verb, 4 adjectives)\end{tabular}            & \begin{tabular}[c]{@{}l@{}}Verb\_V\_Adjective\_3 (98.6)\\ Verb\_V\_Adjective\_5 (98.1)\\ Color\_V\_Animal\_5 (81.4)\end{tabular}   & \begin{tabular}[c]{@{}l@{}}Adjective\_V\_Verb\_3 (1.0)\\ Antonym (7.6)\\ Park\_Country (7.6)\end{tabular}                                   \\
\midrule
Alphabetically\_First\_5 & \begin{tabular}[c]{@{}l@{}}Choose the word that comes \\ first in alphabetical order \\ from a list of 5 words\end{tabular} & \begin{tabular}[c]{@{}l@{}}Alphabetically\_First\_5 (84.8)\\ Alphabetically\_First\_3 (42.4)\\ Commonsense\_QA (29.5)\end{tabular} & \begin{tabular}[c]{@{}l@{}}Alphabetically\_Last\_5 (5.7)\\ Alphabetically\_Last\_3 (8.6)\\ Choose\_Middle\_Of\_3 (13.3)\end{tabular}        \\
\midrule
Alphabetically\_Last\_5  & \begin{tabular}[c]{@{}l@{}}Choose the word that comes \\ last in alphabetical order \\ from a list of 5 words\end{tabular}  & \begin{tabular}[c]{@{}l@{}}Alphabetically\_Last\_5 (39.0)\\ Alphabetically\_Last\_3 (31.4)\\ Commonsense\_QA (25.7)\end{tabular}   & \begin{tabular}[c]{@{}l@{}}Alphabetically\_First\_5 (0.0)\\ Alphabetically\_First\_3 (8.1)\\ Capitalize\_Second\_Letter (13.3)\end{tabular} \\
\midrule
English\_French          & \begin{tabular}[c]{@{}l@{}}Translate the given \\ English word into French\end{tabular}                                     & \begin{tabular}[c]{@{}l@{}}English\_French (81.2)\\ English\_German (80.7)\\ English\_Spanish (80.0)\end{tabular}                  & \begin{tabular}[c]{@{}l@{}}Person\_Instrument (34.0)\\ Next\_Capital\_Letter (38.9)\\ Object\_V\_Concept\_3 (39.3)\end{tabular}             \\
\midrule
English\_German          & \begin{tabular}[c]{@{}l@{}}Translate the given \\ English word into German\end{tabular}                                     & \begin{tabular}[c]{@{}l@{}}English\_French (71.4)\\ English\_Spanish (68.3)\\ English\_German (68.2)\end{tabular}                  & \begin{tabular}[c]{@{}l@{}}Prev\_Item (23.6)\\ Next\_Capital\_Letter (26.2)\\ Person\_Instrument (30.4)\end{tabular}                        \\
\midrule
English\_Spanish         & \begin{tabular}[c]{@{}l@{}}Translate the given \\ English word into Spanish\end{tabular}                                    & \begin{tabular}[c]{@{}l@{}}English\_French (84.4)\\ English\_German (83.4)\\ English\_Spanish (82.9)\end{tabular}                  & \begin{tabular}[c]{@{}l@{}}Person\_Instrument (39.1)\\ Next\_Capital\_Letter (40.8)\\ Object\_V\_Concept\_3 (44.7)
\end{tabular}            \\
\bottomrule
\end{tabular}%
}
\caption{\textbf{Inter-task activation patching analysis for seven evaluation tasks.} For each evaluation task, we report performance after patching high-$\alpha$ attention heads, where task embeddings are fixed to the evaluation task and head-selection parameters are derived from different head-selection tasks. Among the 57 FV tasks, we report the top-3 and bottom-3 head-selection tasks ranked by post-patching accuracy. For clarity, brief task descriptions for all 57 FV tasks are provided in Tables~\ref{tab:Appendix_task_description_1}-\ref{tab:Appendix_task_description_4} of Appendix~\ref{subsec:task descriptions for all 57 ICL tasks}. More results are provided in Tables~\ref{tab:Appendix_additional_cross-task_accuracy}-\ref{tab:Appendix_cross-task_accuracy_llama-3.1-70b} of Appendix~\ref{sec:additional results on cross-task analysis}.}
\label{tab:Analysis_cross-task_accuracy}
\end{table*}
\vspace{-2.5mm}

\subsection{Inter-task activation patching}
\label{subsec:Task-specific vs. task-agnostic: which better explains head roles?}
In inter-task activation patching, we fix both the task of zero-shot inference and the task embeddings used for activation patching, and vary only the head-selection parameters across tasks. Specifically, we select a single \emph{evaluation task}, from which the zero-shot inference input and the task embeddings are derived, and apply head-selection parameters learned from different tasks in the FV benchmark. We refer to these tasks as \emph{head-selection tasks}, as they determine which attention heads~(i.e., high-$\alpha$ heads) are selected for patching. We focus on the 57 FV tasks, which span diverse problem types with closely related and contrasting variants, making them well suited for analyzing task-specific in-context learning behavior.

By keeping the task embeddings fixed and varying only the patching locations using head-selection parameters from different tasks, any differences in model outputs can be attributed solely to task-specific head selection. If attention head functionality is task-specific, then applying head-selection parameters derived from different tasks should lead to substantially different model outputs, despite using identical task embeddings. This experiment therefore tests whether high-$\alpha$ heads reflect task-specific functional roles.

Table~\ref{tab:Analysis_cross-task_accuracy} reports inter-task activation patching results for seven evaluation tasks, along with brief task descriptions for clarity. For each evaluation task, we report the top-3 and bottom-3 head-selection tasks based on post-patching performance. For example, for the evaluation task Adjective\_V\_Verb\_5, the top-performing head-selection tasks include semantically similar tasks such as Adjective\_V\_Verb\_5, Adjective\_V\_Verb\_3, while the bottom-performing ones include semantically dissimilar tasks such as Verb\_V\_Adjective\_3, Verb\_V\_Adjective\_5. Similar patterns are observed for Verb\_V\_Adjective\_5, Alphabetically\_First\_5, and Alphabetically\_Last\_5. For translation tasks, English-French, English-German, and English-Spanish, the top and bottom head-selection tasks are largely consistent across the board. Notably, using high-$\alpha$ heads from semantically similar translation tasks often matches or even exceeds the performance obtained using those from the evaluation task itself, suggesting that closely related tasks share important attention heads. 

Overall, the intra-task and inter-task activation patching analyses reveal strong task specificity in attention head functionality, extending prior studies from a task-agnostic to a task-specific perspective. We hope these findings motivate future work on characterizing task-dependent head roles and exploiting such structure for effective and controllable inference-time LLM adaptations.

\section{Discussion and efficiency analysis}
\label{sec:discussion}

\begin{figure}[t]
\begin{center}
    \includegraphics[width=\linewidth]{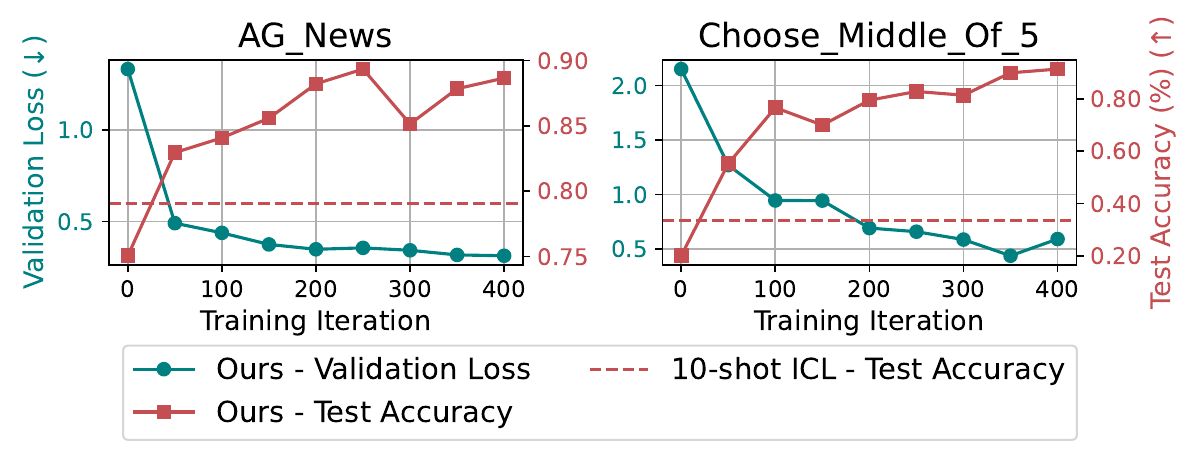}
\end{center}
\vspace{-3mm}
\caption{\textbf{Training dynamics of soft head-selection parameters for two FV tasks.} Validation loss~(left y-axis) and test accuracy~(right y-axis) are plotted over 400 training iterations. Dashed lines indicate the 10-shot ICL accuracies for reference. Plots for all 57 FV tasks are provided in Figures~\ref{fig:analysis_training_dynamics_grid_1}-\ref{fig:analysis_training_dynamics_grid_3} of Appendix~\ref{subsec:training_dynamics-full results for all 57 tasks}.
}
\label{fig:analysis_training_dynamics}
\vspace{-1.0mm}
\end{figure}

\paragraph{Training dynamics of head selection.}
\label{subsec:training dynamics of head selection}
Figure~\ref{fig:analysis_training_dynamics} shows the validation loss and test accuracy of our method during optimization of the soft head-selection parameters for Llama-3.1-8B. The figure shows training dynamics for two FV tasks; results for all 57 FV tasks are provided in Appendix~\ref{subsec:training_dynamics-full results for all 57 tasks}, and some selected plots for larger LLMs are provided in Appendix~\ref{subsec:training dynamics-results with larger language models}. While the optimization trajectories vary slightly across tasks, the overall trend is consistent: validation loss decreases and test accuracy improves over training iterations. This trend indicates that gradient descent effectively tunes the head-selection parameters to identify meaningful head positions for embedding injection.

\begin{figure}[t]
\begin{center}
    \includegraphics[width=\linewidth]{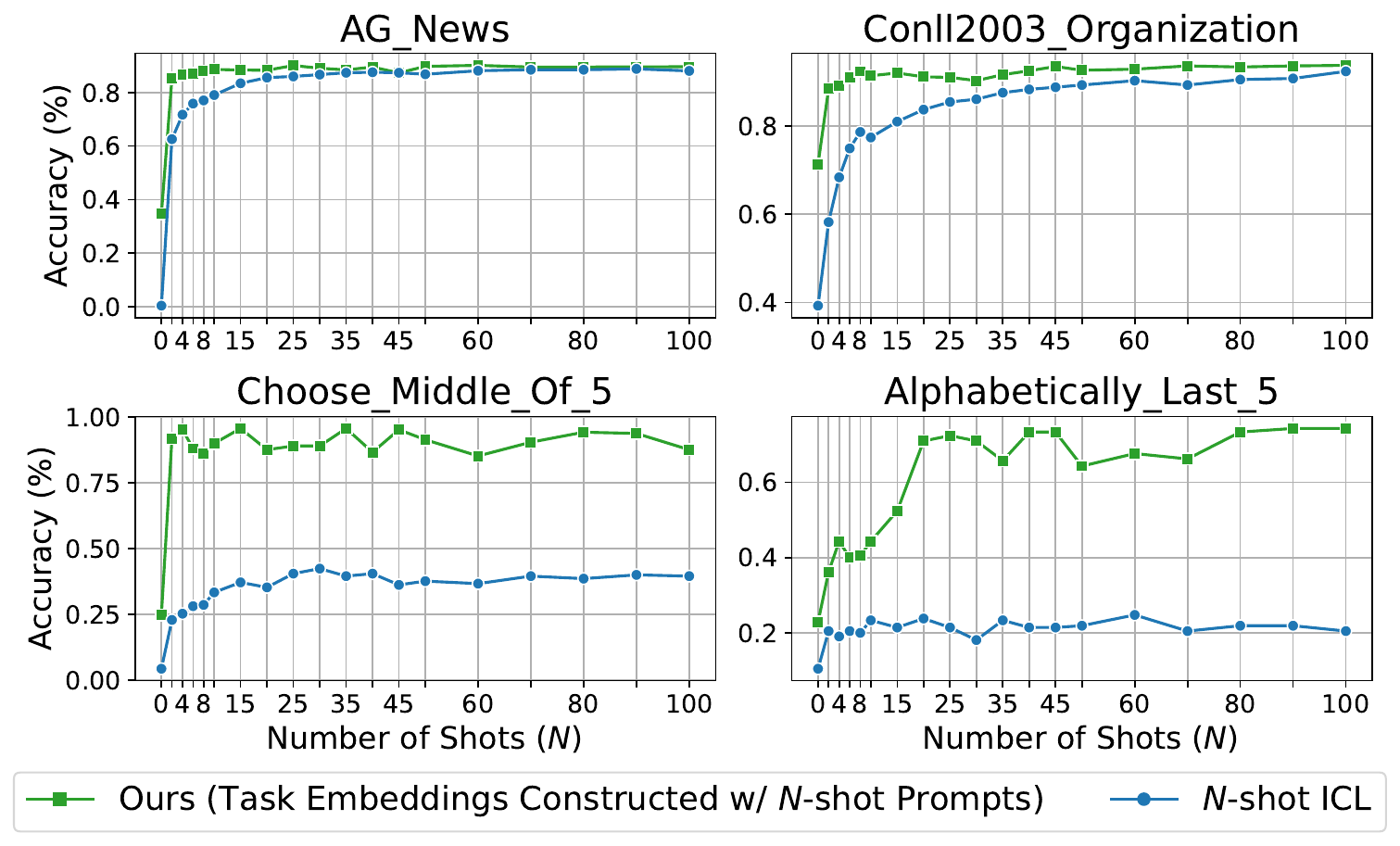}
\end{center}
\vspace{-2.5mm}
\caption{\textbf{Impact of shot count on task performance for four FV tasks.} Performance of standard $N$-shot ICL and our method as the number of shots~($N$) increases from 0 to 100. For our method, $N$ denotes the number of examples per prompt used in task embedding construction. Results are shown for Llama-3.1-8B.
}
\label{fig:analysis_shot_variation}
\end{figure}

\paragraph{Impact of shot count on task performance.}
Increasing the number of in-prompt examples~(or shots) is a common strategy for improving few-shot ICL performance. While some studies~\citep{agarwal2024many, bertsch2025context} report continued gains as the number of shots increases from \emph{a few to many}, others~\citep{zhang2025more} observe that performance often plateaus after only a small number of examples. A plausible explanation, also suggested by recent work~\citep{zou2024retrieval}, is that the benefit of additional shots varies across tasks. Motivated by these observations, we study the effect of shot count on both standard ICL and our task embedding-based adaptation. Specifically, we examine whether many-shot ICL can close the performance gap with our method, and whether increasing the shots of the few-shot prompts used in task embedding construction improves our method by yielding higher-quality ICL-derived task embeddings. Figure~\ref{fig:analysis_shot_variation} reports results on four FV tasks, varying the shots ($N$) for both the $N$-shot ICL and the $N$-shot prompts used in task embedding construction. For AG\_News, and Conll2003\_Organization, many-shot ICL matches the performance of our method at $N=100$. In contrast, tasks such as Choose\_Middle\_Of\_5 and Alphabetically\_Last\_5 exhibit limited gains even with many-shot prompting, indicating that our method remains effective on tasks where many-shot ICL struggles. Moreover, for Alphabetically\_Last\_5, our method benefits from increasing the shots used in task embedding construction, suggesting that some tasks require richer few-shot context to produce effective ICL-derived task embeddings.

\begin{table}[t]
\centering
\setlength{\tabcolsep}{4pt}
\resizebox{\linewidth}{!}{\begin{tabular}{ccccccc}
\toprule
\textbf{\# Test Prompts} & \textbf{0-shot} & \textbf{10-shot} & \textbf{FV} & \textbf{MTV} & \textbf{I2CL} & \textbf{Ours} \\
\midrule
1000  & \textbf{15.7}  & 24.8  & 269.8 & 41.9  & 29.9 & \underline{22.0}  \\
5000  & \textbf{78.3}  & 124.2 & 332.7 & 169.4 & 96.1 & \underline{88.6}  \\
10000 & \textbf{156.7} & 248.4 & 411.2 & 328.8 & 178.7 & \underline{172.0} \\
\bottomrule
\end{tabular}}
\caption{\textbf{Runtime comparison as the number of test prompts increases.} Total runtime~(in minutes) is reported for 1000, 5000, and 10000 prompts on AG\_News using Llama-3.1-8B. Our method scales efficiently as the number of test prompts increases. All runtimes were measured on a single NVIDIA A6000 GPU.}
\label{tab:Discussion_runtime_comparison}
\vspace{-4mm}
\end{table}

\paragraph{Computational efficiency.} Table~\ref{tab:Discussion_runtime_comparison} compares the total runtime of ICL and several embedding-based adaptation methods as the number of test prompts increases. In our method, task embeddings and soft head-selection parameters are computed once~(Stages~1 and~2) and reused across all test inputs, so their cost does not scale with the number of prompts. Consequently, our runtime remains close to the zero-shot baseline and is lower than that of 10-shot ICL. In contrast, FV incurs significant overhead due to extensive head-wise activation patching, while MTV is slowed by a suboptimal loop structure in its original implementation; even after code-level optimization, its runtime remains slightly higher than ours. I2CL achieves total runtime comparable to our method. Our approach also has minimal memory overhead, requiring only about 0.5~MB~(in float32 precision) to store task embeddings and head-selection parameters for Llama-3.1-8B. Overall, our method achieves strong task performance while maintaining the time and memory efficiency of zero-shot inference, particularly as the number of test prompts grows.



\section{Conclusion}
This paper introduces SITE, an ICL-driven embedding-based adaptation method that identifies task-relevant attention heads via a continuous, gradient-based optimization framework. Extensive experiments across four diverse benchmarks and 12 LLMs show that SITE consistently outperforms prior embedding-based adaptation methods and few-shot ICL, while approaching PEFT-level performance with substantially fewer trainable parameters. Beyond empirical gains, SITE offers new mechanistic insights into task-dependent attention head functionality through intra-task and inter-task activation patching analyses.




\section*{Limitations}
While SITE achieves strong performance compared to few-shot ICL and prior embedding-based adaptation methods, it has several limitations. First, SITE requires a modest amount of labeled data to construct ICL-derived task embeddings and optimize the soft head-selection parameters. Although not included in our main experiments, we observed that 30-50 labeled examples are generally sufficient to achieve strong performance, likely because SITE optimizes only a small number of parameters (specifically, $L \times H$ head-selection scalars; e.g., 1024 for Llama-3.1-8B). Nevertheless, acquiring even this amount of labeled data may be challenging for low-resource or newly defined tasks. A promising direction is to augment limited data with LLM-generated synthetic examples, as recent work suggests such synthetic data can rival or even surpass human-curated datasets~\citep{long2024llms, yehudai2024genie, nadas2025synthetic}. Second, SITE requires access to internal model activations, specifically attention head outputs, which restricts its applicability to open-source LLMs and precludes deployment on proprietary models such as GPT-5~\citep{openai2025} or Gemini 2.5~\citep{comanici2025gemini}. Finally, we do not evaluate SITE on multi-step reasoning or long-context generation tasks. Extending the method to these settings and evaluating it on a broader range of benchmarks are important directions for future work.
\section*{Ethical considerations}
Like other embedding-based adaptation methods, our method uses pre-computed ICL-drived task embeddings, instead of in-prompt input-output examples at inference time. In real-world applications, sharing pre-computed task embeddings rather than raw examples can offer benefits such as improved inference efficiency, enhanced privacy protection, and stronger data security. However, such embeddings could also be misused to conceal harmful information and be distributed to users without their awareness. To mitigate these risks, we recommend implementing safeguards, such as pre-release embedding screening, well-defined usage policies, and runtime output filtering tied to embedding identities, prior to the deployment of systems that rely on pre-computed embeddings.


\section*{Acknowledgments}

This work was supported by Institute of Information \& communications Technology Planning \& Evaluation (IITP) grant funded by the Korea government (MSIT) ([NO.RS-2021-II211343, Artificial Intelligence Graduate School Program (Seoul National University)], [No.RS-2023-00235293, Development of autonomous driving big data processing, management, search, and sharing interface technology to provide autonomous driving data according to the purpose of usage]) and the InnoCORE program of the Ministry of Science and ICT (26-InnoCORE-01).

\bibliography{bibliography}

\clearpage
\appendix

\addtocontents{toc}{\protect\setcounter{tocdepth}{2}}
\tableofcontents
\clearpage
\section{Additional experimental setup}
\label{sec:additional experimental setup}

\subsection{12 large language models used for evaluation}
\label{subsec:12 models used for evaluation}
Our evaluation covers 12 large language models in total. From the Llama-3.1 family~\citep{grattafiori2024llama}, we include Llama-3.1-8B, Llama-3.1-8B-Instruct, Llama-3.1-70B, and Llama-3.1-70B-Instruct. From the Mistral family~\citep{jiang2023mistral7b, jiang2024mixtral}, we include Mistral-7B-v0.3, Mistral-7B-Instruct-v0.3, Mixtral-8x7B-v0.1, and Mixtral-8x7B-Instruct-v0.1. From the Qwen3 family~\citep{yang2025qwen3}, we include Qwen3-8B and Qwen3-32B. From the Gemma-3 family~\citep{team2025gemma}, we include Gemma-3-4B-pt and Gemma-3-4B-it. The corresponding variation type, the number of attention layers~($L$), and the number of heads per layer~($H$) for each model are provided in Table~\ref{tab:Experiments_model_list}.

\begin{table*}[ht]
\centering
\resizebox{\textwidth}{!}{\begin{tabular}{l|l|l|c|c}
\toprule
\textbf{Model~Family}               & \textbf{Model~Name}                 & \textbf{Variation~Type}               & \textbf{Attention~Layers~($L$)}  &  \textbf{Heads~per~Layer~($H$)}                                                \\ \midrule
\multirow{4}{*}{Llama-3.1} & Llama-3.1-8B               & Pretrained (Base)                                       & 32  & 32           \\
                           & Llama-3.1-8B-Instruct      & Instruction-tuned~(Chat/Alignment)                      &  32 & 32           \\
                           & Llama-3.1-70B              & Pretrained (Base)                                       & 80  & 64           \\
                           & Llama-3.1-70B-Instruct     & Instruction-tuned~(Chat/Alignment)                      & 80  & 64           \\ \midrule
\multirow{4}{*}{Mistral}   & Mistral-7B-v0.3            & Pretrained (Base)                                       & 32  & 32           \\
                           & Mistral-7B-Instruct-v0.3   & Instruction-tuned~(Chat/Alignment)                      & 32 & 32           \\
                           & Mixtral-8x7B-v0.1          & Mixture of Experts, Pretrained            & 32  & 32           \\
                           & Mixtral-8x7B-Instruct-v0.1 & Mixture of Experts, Instruction-tuned     & 32 & 32           \\ \midrule
\multirow{2}{*}{Qwen3}     & Qwen3-8B                   & Instruction-tuned~(Chat/Alignment)                      & 36  & 32           \\
                           & Qwen3-32B                  & Instruction-tuned~(Chat/Alignment)                      & 64  & 64           \\ \midrule
\multirow{2}{*}{Gemma-3}   & Gemma-3-4B-pt              & Pretrained (Base)                                       & 34  & 8           \\
                           & Gemma-3-4B-it              & Instruction-tuned~(Chat/Alignment)                      & 34  & 8           \\ 
                           \bottomrule
\end{tabular}}
\caption{\textbf{Models used for evaluation.} We consider 12 large language models spanning four model families, three variation types, and model sizes ranging from 4B to 70B parameters.}
\label{tab:Experiments_model_list}
\end{table*}

\subsection{Implementation details and baselines}
\label{subsec:implementation details for FV, MTV, and SITE}
This section describes the implementation details of SITE and provides concise descriptions and configurations of the baseline methods used in our experiments. For baseline methods, we largely follow the default configurations and settings provided in the original papers and official codebases, with minor adjustments noted below for compatibility or fairness.
\begin{itemize}
\item \textbf{SITE~(Ours)}. Algorithm~\ref{algorithm:optimization} details the optimization of the soft head-selection parameters~(Stage~2 of our method, shown in Figure~\ref{fig:method_overview}). We run the optimization for $J=400$ iterations for all tasks and all 12 LLMs. Checkpoints are selected based on the lowest validation loss, evaluated every 50 iterations using up to 100 validation examples. All methods are evaluated using exact match between the ground-truth answer and the initial tokens of the model output. For tasks unrelated to capitalization or lowercasing, we allow case-insensitive matches, following prior work~\citep{huang2024multimodal}, as some tasks contain inconsistent capitalization in the ground-truth labels.
\item \textbf{LoRA}~\citep{hu2022lora}. This method inserts trainable low-rank matrices into the weight matrices of a frozen LLM. We apply LoRA to all query and key projection layers, using rank 8, scaling factor $\alpha=8$, and dropout rate 0.05. A learning rate of 0.0001 is used for all tasks, as higher learning rates occasionally led to unstable training. 
\item \textbf{(IA)$^3$}~\citep{liu2022few}. This method rescales key and value activations in attention layers, as well as the intermediate activations within multilayer perceptron~(MLP) blocks, via element-wise multiplication with learnable vectors. We use a learning rate of 0.002 for all tasks, which is lightly tuned to improve performance.
\item \textbf{Prompt Tuning}~\citep{lester2021power}. This method prepends learnable continuous embeddings to the input embedding sequence of a frozen LLM. We use 20 virtual tokens and a learning rate of 0.05, which is lightly tuned to improve performance.
\item \textbf{Function Vectors~(FV)}~\citep{todd2023function}. This method performs activation patching on attention head outputs using clean few-shot prompts and corrupted~(shuffled) few-shot prompts, and selects heads whose patching most improves performance. The selected head outputs are summed to form a single task embedding, which is added to the activation of a heuristically chosen layer~(typically around one-third of the model depth) during zero-shot inference. We adopt the hyperparameters specified for Llama-2-7B in the official repository and apply them to Llama-3.1-8B, as both models share the same number of layers and attention heads.
\item \textbf{Task Vectors~(TV)}~\citep{hendel2023context}. This method extracts the last-token activation from the selected layer during few-shot inference and injects it by replacement into the same layer during zero-shot inference. The extraction~/~injection layer is selected via a validation sweep over all layers.
\item \textbf{MTV}~\citep{huang2024multimodal}. This method learns a head-sampling distribution using the REINFORCE~\citep{williams1992simple} algorithm. Task embeddings are constructed from attention head activations sampled according to this distribution and injected by replacement during zero-shot inference. While the original implementation uses 100 prompts with 4 demonstrations each for task embedding construction, we instead use 50 prompts with 10 demonstrations per prompt to match our $M=50$ setting. 
\item \textbf{LIVE}~\citep{peng2024live}. This method learns task embeddings by introducing and optimizing trainable layer-wise vectors with associated scaling factors, which are added to the corresponding layer activations. These parameters are optimized end-to-end using a combination of cross-entropy and KL-divergence losses on the output logits. We follow the settings used for VQAv2~\citep{goyal2017making} in the original work, except for increasing the batch size from 2 to 4 and training each task for 50 epochs on its full training set.
\item \textbf{I2CL}~\citep{li2024implicit}. This method constructs task embeddings by averaging last-token activations from multiple 1-shot prompts, separately for the multi-head attention~(MHA) and multilayer perceptron~(MLP) blocks at each layer. These embeddings are added to the corresponding layer activations, with the injection strength learned end-to-end. Although originally designed for classification tasks, we adapt I2CL to open-ended generation and increase the number of prompts used for task embedding construction to 50.
\item \textbf{Iterative Vectors~(IV)}~\citep{liuiterative}. This method defines iterative vectors as differences between attention layer activations with and without demonstrations. These vectors are accumulated over multiple minibatches and injected additively into the corresponding layer activations during inference.
\end{itemize}

\begin{algorithm*}[ht]
    \caption{Optimization of Soft Head-Selection Parameters}\label{algorithm:optimization}
    \begin{algorithmic}[1]
        \REQUIRE $L$: Number of attention layers in LLM
        \REQUIRE $H$: Number of attention heads per layer
        \REQUIRE $J$: Number of training iterations
        \REQUIRE $\mathbb{T}$: Training set of tuples (0-shot prompt, ground-truth answer)
        \REQUIRE $\{ \mathbf{t}^{(l, h)} \}_{l=1, \dots, L;\, h=1, \dots, H}$: Task embeddings for a given task
        \REQUIRE $\text{SA}^{(l,h)}(\cdot)$: Self-attention at head $h$ in layer $l$ (before output projection)
        \REQUIRE $W_o^{(l)}$: Output projection of the multi-head self-attention block in layer $l$
        \REQUIRE $\text{MLP}^{(l)}(\cdot)$: MLP block of layer $l$
        \STATE Initialize soft head-selection parameters $\mathbf{A} = [\alpha^{(l,h)}]_{l=1,\dots, L;\,h=1,\dots,H} \in \mathbb{R}^{L \times H}$ with $\alpha^{(l,h)}=\sigma(0)=0.5$ for all $l, h$, where $\sigma(\cdot)$ denotes the sigmoid function
        \FOR{each iteration $j = 1, 2, \dots, J$}
            \STATE Sample $(P_j, a_j) \sim \mathbb{T}$
            \STATE $\mathbf{e} \leftarrow \text{Tokenize}(P_j)$
            \STATE $\mathbf{v}_1 \leftarrow \text{Embed}(\mathbf{e})$
            \FOR{all $l = 1, 2, \dots, L$}
                \STATE $[\mathbf{u}^{(l,1)}, \mathbf{u}^{(l,2)}, \dots, \mathbf{u}^{(l,H)}] \leftarrow [\text{SA}^{(l,1)}(\mathbf{v}_1), \text{SA}^{(l,2)}(\mathbf{v}_1), \dots, \text{SA}^{(l,H)}(\mathbf{v}_1)]$
                \FOR{all $h = 1, 2, \dots, H$}
                    \STATE $\mathbf{u}^{(l,h)}[-1,:] \leftarrow (1 - \alpha^{(l, h)}) \cdot \mathbf{u}^{(l,h)}[-1,:] + \alpha^{(l, h)} \cdot \mathbf{t}^{(l, h)}$
                \ENDFOR
                \STATE $\mathbf{v} \leftarrow (\mathbf{u}^{(l,1)}\oplus\mathbf{u}^{(l,2)}\oplus\cdots\oplus\mathbf{u}^{(l,H)}) W_o^{(l)}$
                \STATE $\mathbf{v}_2 \leftarrow \mathbf{v}_1 + \mathbf{v}$
                \STATE $\mathbf{v} \leftarrow \text{MLP}^{(l)}(\mathbf{v}_2)$
                \STATE $\mathbf{v}_1 \leftarrow \mathbf{v}_2 + \mathbf{v}$
            \ENDFOR
            \STATE Compute output logits: $\mathbf{p}_j \leftarrow \text{OutputProj}(\mathbf{v}_1)$
            \STATE $\mathcal{L}_j \leftarrow \text{CrossEntropy} (\mathbf{p}_j, a_j)$
            \STATE Update $\mathbf{A}$ with the Adam optimizer: $\mathbf{A} \leftarrow \text{Adam} (\mathbf{A}, \nabla_{\mathbf{A}} \mathcal{L}_j)$
        \ENDFOR
        \STATE \textbf{Return:} Optimized soft head-selection parameters $\mathbf{A} = [\alpha^{(l,h)}]_{l=1,\dots, L;\,h=1,\dots,H} \in \mathbb{R}^{L \times H}$
        \STATE \textit{*Note:} $\oplus$ denotes concatenation.
        \STATE \textit{*Note:} $\mathbf{o}^{(l,h)}$ in Eq.~\ref{eqn:soft injection} corresponds to $\mathbf{o}^{(l,h)}=\mathbf{u}^{(l,h)}[-1,:] \in \mathbb{R}^{d_v}$.
        \STATE \textit{*Note:} Although Lines~8–10 are written with a loop for clarity, they are implemented as a vectorized operation in practice.
    \end{algorithmic}
\end{algorithm*}

\subsection{Overview of all 57 tasks in the FV benchmark}
\label{subsec:task descriptions for all 57 ICL tasks}
In Section~\ref{subsec:Task-specific vs. task-agnostic: which better explains head roles?}, we conduct inter-task activation patching analysis using all 57 tasks from the FV~\citep{todd2023function} benchmark. This benchmark contains both closely related and contrasting task variants, making it well suited for analyzing task-specific in-context learning behavior. It comprises 29 abstractive and 28 extractive tasks, where abstractive tasks require generating information not explicitly present in the prompt, while extractive tasks involve retrieving answers directly from it. Many tasks were introduced by FV, while others originate from prior work and were subsequently filtered or reformatted by FV, including AG\_News~\citep{zhang2015character}, Antonym~\citep{nguyen2017distinguishing}, Synonym~\citep{nguyen2017distinguishing}, Commonsense\_QA~\citep{talmor2018commonsenseqa}, English-French~\citep{conneau2017word}, English-German~\citep{conneau2017word}, English-Spanish~\citep{conneau2017word}, Landmark-Country~\citep{hernandez2023linearity}, Person-Instrument~\citep{hernandez2023linearity}, Person-Occupation~\citep{hernandez2023linearity}, Person-Sport~\citep{hernandez2023linearity}, Product-Company~\citep{hernandez2023linearity}, Sentiment~\citep{socher2013recursive, honovich2022instruction}, Conll2003\_Location~\citep{sang2003introduction}, Conll2003\_Organization~\citep{sang2003introduction}, and Conll2003\_Person~\citep{sang2003introduction}. For completeness and clarity, we present task descriptions and input-output examples for all 57 FV tasks in Tables~\ref{tab:Appendix_task_description_1}-\ref{tab:Appendix_task_description_4}.

\begin{table*}[ht]
\centering
\resizebox{0.9\textwidth}{!}{\begin{tabular}{l|ll}
\toprule
\multirow[c]{2}{*}{\textbf{Task Name}} & \textbf{Task Description}                                                                                                                                                                                                                         &  \\ \cmidrule(l){2-2}
                           & \textbf{Input-Output Example}                                                                                                                                                                                                                                  &  \\ \midrule
\multirow[c]{2}{*}{AG\_News}  & \begin{tabular}[t]{@{}l@{}}Classify a news article based on its headline and opening sentences into one of:\\ 
\textit{World}, \textit{Sports}, \textit{Business}, or \textit{Science/Technology}.\end{tabular} &  \\ \cmidrule(l){2-2}
                           & \begin{tabular}[c]{@{}l@{}}\textbf{Input:} \begin{tabular}[t]{@{}l@{}}Surviving Biotech's Downturns Charly Travers offers advice on \\ withstanding the volatility of the biotech sector. \end{tabular}\\ \textbf{Output:} Business\end{tabular}                                              &  \\ \midrule
\multirow[c]{2}{*}{Antonym}   & Generate the antonym of a given word. &  \\ \cmidrule(l){2-2}
                           & \begin{tabular}[c]{@{}l@{}}\textbf{Input:} overnight\\ \textbf{Output:} daytime\end{tabular} &  \\ \midrule
\multirow[c]{2}{*}{Capitalize} & Capitalize the given word. &  \\ \cmidrule(l){2-2}
                           & \begin{tabular}[c]{@{}l@{}}\textbf{Input:} without\\ \textbf{Output:} Without\end{tabular} &  \\ \midrule
\multirow[c]{2}{*}{Capitalize\_First\_Letter} & Generate the first letter of a given word in capital form. &  \\ \cmidrule(l){2-2}
                           & \begin{tabular}[c]{@{}l@{}}\textbf{Input:} deliver\\ \textbf{Output:} D\end{tabular}  &  \\ \midrule
\multirow[c]{2}{*}{Capitalize\_Last\_Letter} & Generate the last letter of a given word in capital form. &  \\ \cmidrule(l){2-2}
                           & \begin{tabular}[c]{@{}l@{}}\textbf{Input:} clean\\ \textbf{Output:} N\end{tabular}  &  \\ \midrule
\multirow[c]{2}{*}{Capitalize\_Second\_Letter} & Generate the second letter of a given word in capital form. &  \\ \cmidrule(l){2-2}
                           & \begin{tabular}[c]{@{}l@{}}\textbf{Input:} amazing\\ \textbf{Output:} M\end{tabular}  &  \\  \midrule
\multirow[c]{2}{*}{Commonsense\_QA}  & Select the most plausible answer to a commonsense question from five given options. &  \\ \cmidrule(l){2-2}
                           & \begin{tabular}[c]{@{}l@{}}\textbf{Input:} \begin{tabular}[t]{@{}l@{}}Sammy wanted to go to where the people were. Where might he go? \\ a: race track $\;$ b: populated areas $\;$ c: the desert $\;$ d: apartment $\;$ e: roadblock \end{tabular}\\ \textbf{Output:} b\end{tabular}                                              &  \\ \midrule
\multirow[c]{2}{*}{Country-Capital}   & Generate the capital city of a given country. &  \\ \cmidrule(l){2-2}
                           & \begin{tabular}[c]{@{}l@{}}\textbf{Input:} United States of America\\ \textbf{Output:} Washington, D.C.\end{tabular} &  \\ \midrule
\multirow[c]{2}{*}{Country-Currency} & Generate the currency used in a given country. &  \\ \cmidrule(l){2-2}
                           & \begin{tabular}[c]{@{}l@{}}\textbf{Input:} Singapore\\ \textbf{Output:} Singapore Dollar (SGD)\end{tabular} &  \\ \midrule
\multirow[c]{2}{*}{English-French} & Translate the given English word into French. &  \\ \cmidrule(l){2-2}
                           & \begin{tabular}[c]{@{}l@{}}\textbf{Input:} know\\ \textbf{Output:} savoir\end{tabular}  &  \\ \midrule
\multirow[c]{2}{*}{English-German} & Translate the given English word into German. &  \\ \cmidrule(l){2-2}
                           & \begin{tabular}[c]{@{}l@{}}\textbf{Input:} drink\\ \textbf{Output:} trinken\end{tabular}  &  \\ \midrule
\multirow[c]{2}{*}{English-Spanish} & Translate the given English word into Spanish. &  \\ \cmidrule(l){2-2}
                           & \begin{tabular}[c]{@{}l@{}}\textbf{Input:} sometimes\\ \textbf{Output:} a veces\end{tabular}  &  \\ \midrule
\multirow[c]{2}{*}{Landmark-Country}   & Generate the country of a given landmark. &  \\ \cmidrule(l){2-2}
                           & \begin{tabular}[c]{@{}l@{}}\textbf{Input:} South East Forests National Park\\ \textbf{Output:} Australia\end{tabular} &  \\ \midrule
\multirow[c]{2}{*}{Lowercase\_First\_Letter} & Generate the first letter of a given word in lowercase. &  \\ \cmidrule(l){2-2}
                           & \begin{tabular}[c]{@{}l@{}}\textbf{Input:} CLEVER\\ \textbf{Output:} c\end{tabular} &  \\ \midrule
\multirow[c]{2}{*}{Lowercase\_Last\_Letter} & Generate the last letter of a given word in lowercase. &  \\ \cmidrule(l){2-2}
                           & \begin{tabular}[c]{@{}l@{}}\textbf{Input:} PILLOW\\ \textbf{Output:} w\end{tabular}  &  \\ 
\bottomrule
\end{tabular}}
\caption{\textbf{Task descriptions and input-output examples for 57 FV tasks (Part 1 of 4).} 
This table provides task names, descriptions, and representative input-output examples for the FV tasks used in our experiments. The remaining tasks are provided in Tables~\ref{tab:Appendix_task_description_2}-\ref{tab:Appendix_task_description_4}.}
\label{tab:Appendix_task_description_1}
\end{table*}

\clearpage
\begin{table*}[ht]
\centering
\resizebox{0.9\textwidth}{!}{\begin{tabular}{l|ll}
\toprule
\multirow[c]{2}{*}{\textbf{Task Name}} & \textbf{Task Description}                                                                                                                                                                                                                         &  \\ \cmidrule(l){2-2}
                           & \textbf{Input-Output Example}                                                                                                                                                                                                                                  &  \\ \midrule
\multirow[c]{2}{*}{National\_Parks} & Generate the U.S. state of a given national park unit. &  \\ \cmidrule(l){2-2}
                           & \begin{tabular}[c]{@{}l@{}}\textbf{Input:} Glacier Bay National Park\\ \textbf{Output:} Alaska\end{tabular}  &  \\ \midrule
\multirow[c]{2}{*}{Next\_Capital\_Letter} & Generate the next capital letter of the first letter in a given word. &  \\ \cmidrule(l){2-2}
                           & \begin{tabular}[c]{@{}l@{}}\textbf{Input:} microphone\\ \textbf{Output:} N\end{tabular}  &  \\ \midrule
\multirow[c]{2}{*}{Next\_Item} & Generate the next item in a known sequence~(e.g., days, months, letters, or numbers). &  \\ \cmidrule(l){2-2}
                           & \begin{tabular}[c]{@{}l@{}}\textbf{Input:} Friday\\ \textbf{Output:} Saturday\end{tabular} &  \\ \midrule
\multirow[c]{2}{*}{Park-Country} & Generate the country of a given national park. &  \\ \cmidrule(l){2-2}
                           & \begin{tabular}[c]{@{}l@{}}\textbf{Input:} Dartmoor National Park\\ \textbf{Output:} United Kingdom\end{tabular}  &  \\ \midrule
\multirow[c]{2}{*}{Person-Instrument} & Generate the musical instrument played by a given musician. &  \\ \cmidrule(l){2-2}
                           & \begin{tabular}[c]{@{}l@{}}\textbf{Input:} Andor Toth\\ \textbf{Output:} violin\end{tabular}  &  \\ \midrule
\multirow[c]{2}{*}{Person-Occupation} & Generate the occupation of a given individual. &  \\ \cmidrule(l){2-2}
                           & \begin{tabular}[c]{@{}l@{}}\textbf{Input:} Li Yining\\ \textbf{Output:} economist\end{tabular}  &  \\ \midrule
\multirow[c]{2}{*}{Person-Sport}  & Generate the sport played by a given athlete. &  \\ \cmidrule(l){2-2}
                           & \begin{tabular}[c]{@{}l@{}}\textbf{Input:} Andrea Pirlo\\ \textbf{Output:} soccer\end{tabular}                                              &  \\ \midrule
\multirow[c]{2}{*}{Present-Past}   & Generate the past-tense form of a given present-tense verb. &  \\ \cmidrule(l){2-2}
                           & \begin{tabular}[c]{@{}l@{}}\textbf{Input:} write\\ \textbf{Output:} wrote\end{tabular} &  \\ \midrule
\multirow[c]{2}{*}{Prev\_Item} & Generate the previous item in a known sequence~(e.g., days, months, letters, or numbers). &  \\ \cmidrule(l){2-2}
                           & \begin{tabular}[c]{@{}l@{}}\textbf{Input:} april\\ \textbf{Output:} march\end{tabular} &  \\ \midrule
\multirow[c]{2}{*}{Product-Company} & Generate the company associated with a given commercial product. &  \\ \cmidrule(l){2-2}
                           & \begin{tabular}[c]{@{}l@{}}\textbf{Input:} Wii Balance Board\\ \textbf{Output:} Nintendo\end{tabular}  &  \\ \midrule
\multirow[c]{2}{*}{Sentiment} & Generate the sentiment of a given movie review. &  \\ \cmidrule(l){2-2}
                           & \begin{tabular}[c]{@{}l@{}}\textbf{Input:} Very well-written and very well-acted.\\ \textbf{Output:} positive\end{tabular}  &  \\ \midrule
\multirow[c]{2}{*}{Singular-Plural} & Generate the plural form of a given singular noun. &  \\ \cmidrule(l){2-2}
                           & \begin{tabular}[c]{@{}l@{}}\textbf{Input:} island\\ \textbf{Output:} islands\end{tabular}  &  \\ \midrule
\multirow[c]{2}{*}{Synonym}   & Generate the synonym of a given word. &  \\ \cmidrule(l){2-2}
                           & \begin{tabular}[c]{@{}l@{}}\textbf{Input:} identify\\ \textbf{Output:} recognize\end{tabular} &  \\ \midrule
\multirow[c]{2}{*}{Word\_Length} & Generate the number of letters in a given word. &  \\ \cmidrule(l){2-2}
                           & \begin{tabular}[c]{@{}l@{}}\textbf{Input:} discuss\\ \textbf{Output:} 7\end{tabular} &  \\ \midrule
\multirow[c]{2}{*}{Adjective\_V\_Verb\_3} & Select the adjective from a list of 3 words~(1 adjective, 2 verbs). &  \\ \cmidrule(l){2-2}
                           & \begin{tabular}[c]{@{}l@{}}\textbf{Input:} prepare, faithful, develop\\ \textbf{Output:} faithful\end{tabular}  &  \\ 
\bottomrule
\end{tabular}}
\caption{\textbf{Task descriptions and input-output examples for 57 FV tasks (Part 2 of 4).} This table continues from Table~\ref{tab:Appendix_task_description_1}, providing task names, descriptions, and representative input-output examples for the FV tasks used in our experiments. The remaining tasks are provided in Tables~\ref{tab:Appendix_task_description_3}-\ref{tab:Appendix_task_description_4}.}
\label{tab:Appendix_task_description_2}
\end{table*}
\clearpage
\begin{table*}[ht]
\centering
\resizebox{0.84\textwidth}{!}{\begin{tabular}{l|ll}
\toprule
\multirow[c]{2}{*}{\textbf{Task Name}} & \textbf{Task Description}                                                                                                                                                                                                                         &  \\ \cmidrule(l){2-2}
                           & \textbf{Input-Output Example}                                                                                                                                                                                                                                  &  \\ \midrule
\multirow[c]{2}{*}{Adjective\_V\_Verb\_5} & Select the adjective from a list of 5 words~(1 adjective, 4 verbs). &  \\ \cmidrule(l){2-2}
                           & \begin{tabular}[c]{@{}l@{}}\textbf{Input:} remember, teach, knowledgeable, doubt, write\\ \textbf{Output:} knowledgeable\end{tabular}  &  \\ \midrule
\multirow[c]{2}{*}{Alphabetically\_First\_3} & Select the word that comes first in alphabetical order from a list of 3 words. &  \\ \cmidrule(l){2-2}
                           & \begin{tabular}[c]{@{}l@{}}\textbf{Input:} grapefruit, thoughtful, diligent\\ \textbf{Output:} diligent\end{tabular}  &  \\ \midrule
\multirow[c]{2}{*}{Alphabetically\_First\_5} & Select the word that comes first in alphabetical order from a list of 5 words. &  \\ \cmidrule(l){2-2}
                           & \begin{tabular}[c]{@{}l@{}}\textbf{Input:} test, prepare, hammer, beyond, pigeon\\ \textbf{Output:} beyond\end{tabular} &  \\ \midrule
\multirow[c]{2}{*}{Alphabetically\_Last\_3} & Select the word that comes last in alphabetical order from a list of 3 words. &  \\ \cmidrule(l){2-2}
                           & \begin{tabular}[c]{@{}l@{}}\textbf{Input:} sample, garlic, cream\\ \textbf{Output:} sample\end{tabular}  &  \\ \midrule
\multirow[c]{2}{*}{Alphabetically\_Last\_5} & Select the word that comes last in alphabetical order from a list of 5 words. &  \\ \cmidrule(l){2-2}
                           & \begin{tabular}[c]{@{}l@{}}\textbf{Input:} about, navy, gentle, duster, green\\ \textbf{Output:} navy\end{tabular}  &  \\ \midrule
\multirow[c]{2}{*}{Animal\_V\_Object\_3} & Select the animal from a list of 3 words~(1 animal, 2 non-animals). &  \\ \cmidrule(l){2-2}
                           & \begin{tabular}[c]{@{}l@{}}\textbf{Input:} lettuce, basketball, dog\\ \textbf{Output:} dog\end{tabular}  &  \\ \midrule
\multirow[c]{2}{*}{Animal\_V\_Object\_5}  & Select the animal from a list of 5 words~(1 animal, 4 non-animals). &  \\ \cmidrule(l){2-2}
                           & \begin{tabular}[c]{@{}l@{}}\textbf{Input:} soda, rice, potato, snorkel, sloth\\ \textbf{Output:} sloth\end{tabular}                                              &  \\ \midrule
\multirow[c]{2}{*}{Choose\_First\_Of\_3}   & Select the first word from a list of 3 words. &  \\ \cmidrule(l){2-2}
                           & \begin{tabular}[c]{@{}l@{}}\textbf{Input:} ostrich, since, out\\ \textbf{Output:} ostrich\end{tabular} &  \\ \midrule
\multirow[c]{2}{*}{Choose\_First\_Of\_5} & Select the first word from a list of 5 words. &  \\ \cmidrule(l){2-2}
                           & \begin{tabular}[c]{@{}l@{}}\textbf{Input:} reach, puzzle, passionate, silver, complete\\ \textbf{Output:} reach\end{tabular} &  \\ \midrule
\multirow[c]{2}{*}{Choose\_Last\_Of\_3} & Select the last word from a list of 3 words. &  \\ \cmidrule(l){2-2}
                           & \begin{tabular}[c]{@{}l@{}}\textbf{Input:} salmon, rice, socks\\ \textbf{Output:} socks\end{tabular}  &  \\ \midrule
\multirow[c]{2}{*}{Choose\_Last\_Of\_5} & Select the last word from a list of 5 words. &  \\ \cmidrule(l){2-2}
                           & \begin{tabular}[c]{@{}l@{}}\textbf{Input:} spicy, cowardly, hoop, komodo, toward\\ \textbf{Output:} toward\end{tabular}  &  \\ \midrule
\multirow[c]{2}{*}{Choose\_Middle\_Of\_3} & Select the middle word from a list of 3 words. &  \\ \cmidrule(l){2-2}
                           & \begin{tabular}[c]{@{}l@{}}\textbf{Input:} garlic, candle, argue\\ \textbf{Output:} candle\end{tabular}  &  \\ \midrule
\multirow[c]{2}{*}{Choose\_Middle\_Of\_5}   & Select the middle word from a list of 5 words. &  \\ \cmidrule(l){2-2}
                           & \begin{tabular}[c]{@{}l@{}}\textbf{Input:} table, qualify, airplane, harmonious, happy\\ \textbf{Output:} airplane\end{tabular} &  \\ \midrule
\multirow[c]{2}{*}{Color\_V\_Animal\_3} & Select the color from a list of 3 words~(1 color, 2 animals). &  \\ \cmidrule(l){2-2}
                           & \begin{tabular}[c]{@{}l@{}}\textbf{Input:} camel, penguin, brown\\ \textbf{Output:} brown\end{tabular} &  \\ \midrule
\multirow[c]{2}{*}{Color\_V\_Animal\_5} & Select the color from a list of 5 words~(1 color, 4 animals). &  \\ \cmidrule(l){2-2}
                           & \begin{tabular}[c]{@{}l@{}}\textbf{Input:} salamander, chinchilla, flamingo, black, tiger\\ \textbf{Output:} black\end{tabular}  &  \\ 
\bottomrule
\end{tabular}}
\caption{\textbf{Task descriptions and input-output examples for 57 FV tasks (Part 3 of 4).} This table continues from Tables~\ref{tab:Appendix_task_description_1}-\ref{tab:Appendix_task_description_2}, providing task names, descriptions, and representative input-output examples for the FV tasks used in our experiments. The remaining tasks are provided in Table~\ref{tab:Appendix_task_description_4}.}
\label{tab:Appendix_task_description_3}
\end{table*}
\clearpage
\begin{table*}[ht]
\centering
\resizebox{0.85\textwidth}{!}{\begin{tabular}{l|ll}
\toprule
\multirow[c]{2}{*}{\textbf{Task Name}} & \textbf{Task Description}                                                                                                                                                                                                                         &  \\ \cmidrule(l){2-2}
                           & \textbf{Input-Output Example}                                                                                                                                                                                                                                  &  \\ \midrule
\multirow[c]{2}{*}{Concept\_V\_Object\_3} & Select the concept from a list of 3 words~(1 abstract concept, 2 concrete entities). &  \\ \cmidrule(l){2-2}
                           & \begin{tabular}[c]{@{}l@{}}\textbf{Input:} radio, whimsical, robot\\ \textbf{Output:} whimsical\end{tabular}  &  \\ \midrule
\multirow[c]{2}{*}{Concept\_V\_Object\_5} & Select the concept from a list of 5 words~(1 abstract concept, 4 concrete entities). &  \\ \cmidrule(l){2-2}
                           & \begin{tabular}[c]{@{}l@{}}\textbf{Input:} towel, map, hammock, read, blanket\\ \textbf{Output:} read\end{tabular}  &  \\ \midrule
\multirow[c]{2}{*}{Conll2003\_Location} & Select the location entity from a given sentence. &  \\ \cmidrule(l){2-2}
                           & \begin{tabular}[c]{@{}l@{}}\textbf{Input:} Clinton arrives in Chicago on day of re-nomination.\\ \textbf{Output:} Chicago\end{tabular} &  \\ \midrule
\multirow[c]{2}{*}{Conll2003\_Organization} & Select the organization entity from a given sentence. &  \\ \cmidrule(l){2-2}
                           & \begin{tabular}[c]{@{}l@{}}\textbf{Input:} Advertising revenues at The Times grew 20 percent.\\ \textbf{Output:} The Times\end{tabular}  &  \\ \midrule
\multirow[c]{2}{*}{Conll2003\_Person} & Select the person entity from a given sentence. &  \\ \cmidrule(l){2-2}
                           & \begin{tabular}[c]{@{}l@{}}\textbf{Input:} They contained \$ 650,000 in jewelry and \$ 40,000 in cash, Andrews said.\\ \textbf{Output:} Andrews\end{tabular}  &  \\ \midrule
\multirow[c]{2}{*}{Fruit\_V\_Animal\_3} & Select the fruit from a list of 3 words~(1 fruit, 2 animals). &  \\ \cmidrule(l){2-2}
                           & \begin{tabular}[c]{@{}l@{}}\textbf{Input:} pineapple, iguana, leopard\\ \textbf{Output:} pineapple\end{tabular}  &  \\ \midrule
\multirow[c]{2}{*}{Fruit\_V\_Animal\_5}  & Select the fruit from a list of 5 words~(1 fruit, 4 animals). &  \\ \cmidrule(l){2-2}
                           & \begin{tabular}[c]{@{}l@{}}\textbf{Input:} walrus, lizard, panther, lion, cranberry\\ \textbf{Output:} cranberry\end{tabular}                                              &  \\ \midrule
\multirow[c]{2}{*}{Object\_V\_Concept\_3}   & Select the concrete entity from a list of 3 words~(1 concrete entity, 2 abstract concepts). &  \\ \cmidrule(l){2-2}
                           & \begin{tabular}[c]{@{}l@{}}\textbf{Input:} need, lamp, beneath\\ \textbf{Output:} lamp\end{tabular} &  \\ \midrule
\multirow[c]{2}{*}{Object\_V\_Concept\_5} & Select the concrete entity from a list of 5 words~(1 concrete entity, 4 abstract concepts). &  \\ \cmidrule(l){2-2}
                           & \begin{tabular}[c]{@{}l@{}}\textbf{Input:} passionate, jigsaw, remove, expensive, fearless\\ \textbf{Output:} jigsaw\end{tabular} &  \\ \midrule
\multirow[c]{2}{*}{Squad\_Val} & Retrieve the answer to a given question based on a provided context paragraph. &  \\ \cmidrule(l){2-2}
                           & \begin{tabular}[c]{@{}l@{}}\textbf{Input:} \begin{tabular}[t]{@{}l@{}} The Panthers offense, which led the NFL in scoring (500 points), was loaded \\with talent, boasting six Pro Bowl selections. Pro Bowl quarterback Cam \\Newton had one of his best seasons, throwing for 3,837 yards and rushing for \\636, while recording a career-high and league-leading 45 total touchdowns \\(35 passing, 10 rushing), a career-low 10 interceptions, and a career-best \\quarterback rating of 99.4. Newton's leading receivers were tight end Greg \\Olsen, who caught a career-high 77 passes for 1,104 yards and seven \\touchdowns, and wide receiver Ted Ginn, Jr., who caught 44 passes for 739 \\yards and 10 touchdowns; Ginn also rushed for 60 yards and returned 27 \\punts for 277 yards. Other key receivers included veteran Jerricho Cotchery \\(39 receptions for 485 yards), rookie Devin Funchess (31 receptions for 473 \\yards and five touchdowns), and second-year receiver Corey Brown (31 \\receptions for 447 yards). The Panthers backfield featured Pro Bowl running \\back Jonathan Stewart, who led the team with 989 rushing yards and six \\touchdowns in 13 games, along with Pro Bowl fullback Mike Tolbert, who \\rushed for 256 yards and caught 18 passes for another 154 yards. Carolina's \\offensive line also featured two Pro Bowl selections: center Ryan Kalil \\and guard Trai Turner. \\What position does Jerricho Cotchery play? \end{tabular}\\ \textbf{Output:} receivers\end{tabular}  &  \\ \midrule
\multirow[c]{2}{*}{Verb\_V\_Adjective\_3} & Select the verb from a list of 3 words~(1 verb, 2 adjectives). &  \\ \cmidrule(l){2-2}
                           & \begin{tabular}[c]{@{}l@{}}\textbf{Input:} dirty, dance, diligent\\ \textbf{Output:} dance\end{tabular}  &  \\ \midrule
\multirow[c]{2}{*}{Verb\_V\_Adjective\_5} & Select the verb from a list of 5 words~(1 verb, 4 adjectives). &  \\ \cmidrule(l){2-2}
                           & \begin{tabular}[c]{@{}l@{}}\textbf{Input:} heavy, overcome, quick, modern, dazzling\\ \textbf{Output:} overcome\end{tabular}  &  \\
\bottomrule
\end{tabular}}
\caption{\textbf{Task descriptions and input-output examples for 57 FV tasks (Part 4 of 4).} This table concludes the series from Tables~\ref{tab:Appendix_task_description_1}-\ref{tab:Appendix_task_description_3}, providing task names, descriptions, and representative input-output examples for the FV tasks used in our experiments.}
\label{tab:Appendix_task_description_4}
\end{table*}
\clearpage

\subsection{Ablation study on prompt templates}
\label{subsec:ablation study on prompt templates}
For all experiments, we adopt FV's default ICL prompt template: \texttt{`Q:\{$x_{i1}$\}\textbackslash nA:\{$y_{i1}$\}\textbackslash n\textbackslash n \dots~Q:\{$x_{iN}$\}\textbackslash nA:\{$y_{iN}$\}\textbackslash n\textbackslash nQ:\{$x_{iq}$\}\textbackslash nA:'}, where each $\{x_{ik}\}$ and $\{y_{ik}\}$ is replaced with the corresponding input-output pair. To assess the robustness of our method to prompt formatting, we conduct an ablation study using five prompt templates, each provided by FV~\citep{todd2023function}, including the default template used in all other experiments. These templates are listed in Table~\ref{tab:Appendix_list_of_prompt_templates}. For each template, we evaluate our method along with 0-shot and 10-shot ICL across all 57 FV tasks using Llama-3.1-8B, with results reported in Table~\ref{tab:Appendix_prompt_template_ablation}. Across all five templates, our method consistently achieves strong performance, with average accuracies ranging from 89.0\% to 91.2\%, significantly outperforming the 10-shot ICL~(76.8\%-77.8\%). These results demonstrate the robustness of our method to variations in prompt format.

\begin{table}[ht]
\centering
\resizebox{\linewidth}{!}{%
\begin{tabular}{l|l}
\toprule
\textbf{Prompt Template} & \textbf{Format of a single $(\{x_{ik}\}, \{y_{ik}\})$ pair} \\
\midrule
Template~1 (Default) & \texttt{Q:\{$x_{ik}$\}\textbackslash nA:\{$y_{ik}$\}\textbackslash n\textbackslash n} \\
Template~2           & \texttt{question:\{$x_{ik}$\}\textbackslash nanswer:\{$y_{ik}$\}\textbackslash n\textbackslash n} \\
Template~3           & \texttt{A:\{$x_{ik}$\}\textbackslash nB:\{$y_{ik}$\}\textbackslash n\textbackslash n} \\
Template~4           & \texttt{\{$x_{ik}$\} $\rightarrow$\{$y_{ik}$\}\textbackslash n\textbackslash n} \\
Template~5           & \texttt{input:\{$x_{ik}$\} output:\{$y_{ik}$\}\textbackslash n} \\
\bottomrule
\end{tabular}
}
\caption{\textbf{Prompt templates used in the ablation study.} Each template shows how a single input-output pair $(\{x_{ik}\}, \{y_{ik}\})$ is formatted. All templates are sourced from FV~\citep{todd2023function}. Template~1 serves as the default prompt format used in all main experiments.}
\label{tab:Appendix_list_of_prompt_templates}
\end{table}

\begin{table*}[ht]
\centering
\setlength{\tabcolsep}{3pt}
\resizebox{\textwidth}{!}{%
\begin{tabular}{l|ccc|ccc|ccc|ccc|ccc}
\toprule
\multirow{2}{*}{\textbf{Task Type}}          & \multicolumn{3}{|c|}{\textbf{Template 1~(default)}} & \multicolumn{3}{c|}{\textbf{Template 2}} & \multicolumn{3}{c|}{\textbf{Template 3}} & \multicolumn{3}{c|}{\textbf{Template 4}} & \multicolumn{3}{c}{\textbf{Template 5}} \\ \cmidrule(lr){2-4} \cmidrule(lr){5-7} \cmidrule(lr){8-10} \cmidrule(lr){11-13} \cmidrule(lr){14-16}
      & \textbf{0-shot}       & \textbf{10-shot}       & \textbf{Ours}      & \textbf{0-shot}    & \textbf{10-shot}   & \textbf{Ours} \textbf{}  & \textbf{0-shot}    & \textbf{10-shot}   & \textbf{Ours}   & \textbf{0-shot}    & \textbf{10-shot}   & \textbf{Ours}   & \textbf{0-shot}    & \textbf{10-shot}   & \textbf{Ours}   \\ \toprule
Abstractive    & 2.7          & 75.9          & 87.7      & 3.6       & 77.0      & 87.8   & 3.7       & 75.1      & 87.4   & 3.3       & 75.9      & 87.6   & 2.4       & 76.4      & 86.3   \\
Extractive     & 17.3         & 77.6          & 92.4      & 26.9      & 76.6      & 90.9   & 23.2      & 79.2      & 93.0   & 5.0       & 78.7      & 94.8   & 25.0      & 79.2      & 91.7   \\ \midrule
All~(57 FV tasks) & 9.9          & \underline{76.8}          & \textbf{90.0}      & 15.0      & \underline{76.8}      & \textbf{89.3}   & 13.3      & \underline{77.1}      & \textbf{90.2}   & 4.2       & \underline{77.3}      & \textbf{91.2}   & 13.5      & \underline{77.8}      & \textbf{89.0}  \\
\bottomrule
\end{tabular}}
\caption{\textbf{Results of prompt template ablation using Llama-3.1-8B.} Average accuracies for our method and the 0-shot/10-shot ICL baselines are reported across five prompt templates. Results are shown for 29 abstractive tasks, 28 extractive tasks, and all 57 FV tasks, respectively. Our method consistently demonstrates strong performance across all templates.
}
\label{tab:Appendix_prompt_template_ablation}
\end{table*}


\section{Task-wise performance of the FV benchmark across 12 LLMs}
\label{sec:task-wise performance across 11 additional LLMs}
Figure~\ref{fig:performance_of_various_models} in Section~\ref{subsec:Experimental results} shows the average performance of our method across all 57 FV tasks, compared to the 10-shot ICL, for all 12 LLMs listed in Table~\ref{tab:Experiments_model_list}. 
Task-wise results for all 12 LLMs are provided in Tables~\ref{tab:main_results_gemma-3-4b-pt}–\ref{tab:main_results_llama-3.1-70b-instruct}.

\clearpage
\begin{table*}[ht]
\centering
\setlength{\tabcolsep}{4.5pt}
\begin{subtable}[t]{0.495\textwidth}
\centering
\scriptsize
\resizebox{0.99\textwidth}{!}{\begin{tabular}{lccc}
\toprule
\textbf{Task Name}                  & \textbf{0-shot} & \textbf{10-shot} & \textbf{Ours}  \\ \midrule
AG\_News                   & 0.3    & 79.7    & 84.3  \\
Antonym                    & 0.4    & 66.3    & 66.7  \\
Capitalize                 & 1.2    & 99.4    & 99.4  \\
Capitalize\_First\_Letter  & 1.2    & 99.4    & 99.4  \\
Capitalize\_Last\_Letter   & 0.6    & 15.8    & 71.9  \\
Capitalize\_Second\_Letter & 0.0    & 25.5    & 64.2  \\
Commonsense\_QA            & 30.5   & 67.5    & 59.8  \\
Country-Capital            & 0.0    & 92.9    & 88.1  \\
Country-Currency           & 0.0    & 81.0    & 78.6  \\
English-French             & 0.5    & 81.2    & 75.7  \\
English-German             & 0.4    & 75.5    & 69.9  \\
English-Spanish            & 0.3    & 84.3    & 81.6  \\
Landmark-Country           & 1.7    & 82.3    & 78.3  \\
Lowercase\_First\_Letter   & 0.0    & 100.0   & 100.0 \\
Lowercase\_Last\_Letter    & 0.6    & 37.4    & 94.7  \\
National\_Parks            & 7.4    & 79.0    & 79.0  \\
Next\_Capital\_Letter      & 2.3    & 1.8     & 35.7  \\
Next\_Item                 & 0.0    & 89.4    & 91.5  \\
Park-Country               & 21.7   & 82.2    & 77.1  \\
Person-Instrument          & 0.9    & 65.4    & 69.2  \\
Person-Occupation          & 0.0    & 52.3    & 64.5  \\
Person-Sport               & 0.0    & 94.0    & 98.5  \\
Present-Past               & 1.6    & 100.0   & 100.0 \\
Prev\_Item                 & 0.0    & 66.0    & 83.0  \\
Product-Company            & 2.8    & 78.9    & 78.9  \\
Sentiment                  & 0.0    & 95.9    & 94.7  \\
Singular-Plural            & 2.3    & 100.0   & 100.0 \\
Synonym                    & 6.8    & 50.0    & 52.3  \\
Word\_Length               & 0.0    & 18.1    & 28.1  \\ \midrule
Average                    & 2.9    & \underline{71.1}    & \textbf{78.1}         \\
\bottomrule
\end{tabular}}
\caption{Abstractive task results}
\end{subtable}
\hfill
\begin{subtable}[t]{0.495\textwidth}
\centering
\scriptsize
\resizebox{0.99\textwidth}{!}{\begin{tabular}{lccccc}
\toprule
\textbf{Task Name}                & \textbf{0-shot} & \textbf{10-shot} & \textbf{Ours}  \\ \midrule
Adjective\_V\_Verb\_3    & 16.7   & 74.3    & 95.2  \\
Adjective\_V\_Verb\_5    & 15.2   & 66.7    & 94.3  \\
Alphabetically\_First\_3 & 24.3   & 29.5    & 33.3  \\
Alphabetically\_First\_5 & 22.9   & 23.3    & 28.6  \\
Alphabetically\_Last\_3  & 21.9   & 30.5    & 37.1  \\
Alphabetically\_Last\_5  & 11.9   & 19.1    & 29.5  \\
Animal\_V\_Object\_3     & 10.5   & 70.5    & 98.1  \\
Animal\_V\_Object\_5     & 16.7   & 64.8    & 95.7  \\
Choose\_First\_Of\_3     & 65.2   & 99.5    & 100.0 \\
Choose\_First\_Of\_5     & 82.9   & 98.6    & 100.0 \\
Choose\_Last\_Of\_3      & 4.8    & 96.7    & 99.5  \\
Choose\_Last\_Of\_5      & 0.0    & 92.4    & 100.0 \\
Choose\_Middle\_Of\_3    & 1.4    & 55.7    & 94.8  \\
Choose\_Middle\_Of\_5    & 0.5    & 21.9    & 51.9  \\
Color\_V\_Animal\_3      & 12.9   & 83.8    & 100.0 \\
Color\_V\_Animal\_5      & 11.0   & 84.3    & 99.1  \\
Concept\_V\_Object\_3    & 20.5   & 70.5    & 94.8  \\
Concept\_V\_Object\_5    & 17.1   & 62.9    & 95.7  \\
Conll2003\_Location      & 9.3    & 82.1    & 91.3  \\
Conll2003\_Organization  & 12.6   & 75.6    & 88.4  \\
Conll2003\_Person        & 12.9   & 87.9    & 95.7  \\
Fruit\_V\_Animal\_3      & 6.2    & 74.8    & 98.6  \\
Fruit\_V\_Animal\_5      & 3.3    & 71.0    & 99.5  \\
Object\_V\_Concept\_3    & 15.2   & 71.4    & 97.1  \\
Object\_V\_Concept\_5    & 14.8   & 61.9    & 96.2  \\
Squad\_Val               & 53.1   & 85.8    & 86.0  \\
Verb\_V\_Adjective\_3    & 11.4   & 67.1    & 98.1  \\
Verb\_V\_Adjective\_5    & 5.2    & 71.0    & 98.6  \\ \midrule
Average                  & 17.9   & \underline{67.6}    & \textbf{85.6}                  \\
\bottomrule
\end{tabular}}
\caption{Extractive task results}
\end{subtable}
\caption{\textbf{Task-wise performance on 57 FV tasks using Gemma-3-4B-pt.} Our method is evaluated along with 0-shot and 10-shot ICL baselines. (a) Results on 29 abstractive tasks. (b) Results on 28 extractive tasks. The best results are shown in \textbf{bold}, and the second-best results are \underline{underlined}.}
\label{tab:main_results_gemma-3-4b-pt}
\end{table*}

\clearpage
\begin{table*}[ht]
\centering
\setlength{\tabcolsep}{4.5pt}
\begin{subtable}[t]{0.495\textwidth}
\centering
\scriptsize
\resizebox{0.99\textwidth}{!}{\begin{tabular}{lccc}
\toprule
\textbf{Task Name}                  & \textbf{0-shot} & \textbf{10-shot} & \textbf{Ours}  \\ \midrule
AG\_News                   & 0.0    & 75.4    & 84.9  \\
Antonym                    & 0.0    & 66.3    & 70.2  \\
Capitalize                 & 0.0    & 99.4    & 99.4  \\
Capitalize\_First\_Letter  & 4.7    & 96.5    & 100.0 \\
Capitalize\_Last\_Letter   & 2.9    & 18.1    & 79.0  \\
Capitalize\_Second\_Letter & 7.3    & 18.8    & 93.9  \\
Commonsense\_QA            & 59.6   & 68.8    & 61.6  \\
Country-Capital            & 0.0    & 90.5    & 90.5  \\
Country-Currency           & 0.0    & 69.1    & 78.6  \\
English-French             & 0.3    & 81.4    & 71.4  \\
English-German             & 0.1    & 75.7    & 64.2  \\
English-Spanish            & 0.0    & 84.2    & 81.2  \\
Landmark-Country           & 0.0    & 76.6    & 74.9  \\
Lowercase\_First\_Letter   & 0.0    & 99.4    & 100.0 \\
Lowercase\_Last\_Letter    & 0.0    & 39.8    & 90.6  \\
National\_Parks            & 0.0    & 69.5    & 70.5  \\
Next\_Capital\_Letter      & 2.3    & 4.7     & 65.5  \\
Next\_Item                 & 2.1    & 97.9    & 95.7  \\
Park-Country               & 0.0    & 76.4    & 72.6  \\
Person-Instrument          & 0.0    & 48.6    & 57.0  \\
Person-Occupation          & 0.0    & 39.5    & 62.8  \\
Person-Sport               & 0.0    & 94.0    & 97.0  \\
Present-Past               & 0.0    & 100.0   & 98.4  \\
Prev\_Item                 & 2.1    & 74.5    & 87.2  \\
Product-Company            & 2.8    & 72.5    & 66.1  \\
Sentiment                  & 0.0    & 91.0    & 94.3  \\
Singular-Plural            & 0.0    & 100.0   & 97.7  \\
Synonym                    & 6.0    & 50.3    & 51.7  \\
Word\_Length               & 0.0    & 38.0    & 68.4  \\ \midrule
Average                    & 3.1    & \underline{69.5}    & \textbf{80.2}         \\
\bottomrule
\end{tabular}}
\caption{Abstractive task results}
\end{subtable}
\hfill
\begin{subtable}[t]{0.495\textwidth}
\centering
\scriptsize
\resizebox{0.99\textwidth}{!}{\begin{tabular}{lccccc}
\toprule
\textbf{Task Name}                & \textbf{0-shot} & \textbf{10-shot} & \textbf{Ours}  \\ \midrule
Adjective\_V\_Verb\_3    & 3.8    & 77.6    & 97.6  \\
Adjective\_V\_Verb\_5    & 15.7   & 72.9    & 95.7  \\
Alphabetically\_First\_3 & 21.9   & 28.1    & 31.9  \\
Alphabetically\_First\_5 & 22.9   & 21.4    & 26.7  \\
Alphabetically\_Last\_3  & 12.9   & 39.1    & 39.5  \\
Alphabetically\_Last\_5  & 16.7   & 25.7    & 29.1  \\
Animal\_V\_Object\_3     & 21.9   & 81.0    & 98.6  \\
Animal\_V\_Object\_5     & 23.3   & 91.0    & 97.1  \\
Choose\_First\_Of\_3     & 31.0   & 99.5    & 100.0 \\
Choose\_First\_Of\_5     & 25.7   & 99.5    & 100.0 \\
Choose\_Last\_Of\_3      & 10.0   & 99.1    & 100.0 \\
Choose\_Last\_Of\_5      & 15.7   & 94.8    & 100.0 \\
Choose\_Middle\_Of\_3    & 7.6    & 70.0    & 94.3  \\
Choose\_Middle\_Of\_5    & 12.4   & 35.2    & 66.7  \\
Color\_V\_Animal\_3      & 17.1   & 71.4    & 100.0 \\
Color\_V\_Animal\_5      & 13.8   & 79.5    & 99.1  \\
Concept\_V\_Object\_3    & 27.6   & 61.9    & 97.6  \\
Concept\_V\_Object\_5    & 34.3   & 75.2    & 93.3  \\
Conll2003\_Location      & 7.7    & 86.4    & 92.4  \\
Conll2003\_Organization  & 19.7   & 77.8    & 88.2  \\
Conll2003\_Person        & 23.1   & 93.4    & 96.4  \\
Fruit\_V\_Animal\_3      & 11.4   & 74.8    & 97.6  \\
Fruit\_V\_Animal\_5      & 4.8    & 82.9    & 100.0 \\
Object\_V\_Concept\_3    & 3.3    & 77.6    & 97.6  \\
Object\_V\_Concept\_5    & 3.3    & 67.6    & 96.2  \\
Squad\_Val               & 71.7   & 87.9    & 86.4  \\
Verb\_V\_Adjective\_3    & 3.8    & 69.5    & 95.7  \\
Verb\_V\_Adjective\_5    & 6.7    & 71.4    & 99.1  \\ \midrule
Average                  & 17.5   & \underline{71.9}    & \textbf{86.3}        \\
\bottomrule
\end{tabular}}
\caption{Extractive task results}
\end{subtable}
\caption{\textbf{Task-wise performance on 57 FV tasks using Gemma-3-4B-it.} Our method is evaluated along with 0-shot and 10-shot ICL baselines. (a) Results on 29 abstractive tasks. (b) Results on 28 extractive tasks. The best results are shown in \textbf{bold}, and the second-best results are \underline{underlined}.}
\label{tab:main_results_gemma-3-4b-it}
\end{table*}
\clearpage
\begin{table*}[ht]
\centering
\setlength{\tabcolsep}{4.5pt}
\begin{subtable}[t]{0.495\textwidth}
\centering
\scriptsize
\resizebox{0.99\textwidth}{!}{\begin{tabular}{lccc}
\toprule
\textbf{Task Name}                  & \textbf{0-shot} & \textbf{10-shot} & \textbf{Ours}  \\ \midrule
AG\_News                   & 0.4    & 81.0    & 88.9  \\
Antonym                    & 7.9    & 68.3    & 67.9  \\
Capitalize                 & 6.5    & 100.0   & 100.0 \\
Capitalize\_First\_Letter  & 6.5    & 90.0    & 99.4  \\
Capitalize\_Last\_Letter   & 0.6    & 33.9    & 86.6  \\
Capitalize\_Second\_Letter & 0.6    & 27.9    & 96.4  \\
Commonsense\_QA            & 21.1   & 70.8    & 59.0  \\
Country-Capital            & 4.8    & 90.5    & 88.1  \\
Country-Currency           & 0.0    & 78.6    & 78.6  \\
English-French             & 0.3    & 79.8    & 77.7  \\
English-German             & 1.4    & 74.0    & 63.2  \\
English-Spanish            & 0.3    & 84.6    & 79.6  \\
Landmark-Country           & 0.0    & 85.1    & 82.9  \\
Lowercase\_First\_Letter   & 0.0    & 83.0    & 100.0 \\
Lowercase\_Last\_Letter    & 0.0    & 49.1    & 95.3  \\
National\_Parks            & 1.1    & 79.0    & 77.9  \\
Next\_Capital\_Letter      & 0.6    & 5.3     & 98.8  \\
Next\_Item                 & 0.0    & 97.9    & 97.9  \\
Park-Country               & 0.0    & 87.3    & 79.6  \\
Person-Instrument          & 0.0    & 75.7    & 76.6  \\
Person-Occupation          & 0.0    & 59.9    & 70.0  \\
Person-Sport               & 0.0    & 92.5    & 97.0  \\
Present-Past               & 1.6    & 98.4    & 100.0 \\
Prev\_Item                 & 0.0    & 91.5    & 95.7  \\
Product-Company            & 0.9    & 82.6    & 80.7  \\
Sentiment                  & 0.0    & 94.7    & 93.9  \\
Singular-Plural            & 2.3    & 97.7    & 97.7  \\
Synonym                    & 1.7    & 51.2    & 47.9  \\
Word\_Length               & 0.0    & 31.6    & 63.7  \\ \midrule
Average                    & 2.0    & \underline{73.8}    & \textbf{84.2}    \\
\bottomrule
\end{tabular}}
\caption{Abstractive task results}
\end{subtable}
\hfill
\begin{subtable}[t]{0.495\textwidth}
\centering
\scriptsize
\resizebox{0.99\textwidth}{!}{\begin{tabular}{lccccc}
\toprule
\textbf{Task Name}                & \textbf{0-shot} & \textbf{10-shot} & \textbf{Ours}  \\ \midrule
Adjective\_V\_Verb\_3    & 34.8   & 76.7    & 98.1  \\
Adjective\_V\_Verb\_5    & 16.7   & 79.1    & 97.1  \\
Alphabetically\_First\_3 & 31.0   & 32.9    & 53.8  \\
Alphabetically\_First\_5 & 20.5   & 22.9    & 86.2  \\
Alphabetically\_Last\_3  & 24.8   & 27.1    & 45.7  \\
Alphabetically\_Last\_5  & 11.0   & 18.6    & 51.4  \\
Animal\_V\_Object\_3     & 24.3   & 71.9    & 96.7  \\
Animal\_V\_Object\_5     & 24.3   & 88.1    & 99.1  \\
Choose\_First\_Of\_3     & 81.4   & 100.0   & 100.0 \\
Choose\_First\_Of\_5     & 73.8   & 100.0   & 99.1  \\
Choose\_Last\_Of\_3      & 2.9    & 99.5    & 100.0 \\
Choose\_Last\_Of\_5      & 1.9    & 97.1    & 100.0 \\
Choose\_Middle\_Of\_3    & 5.7    & 42.9    & 98.6  \\
Choose\_Middle\_Of\_5    & 0.5    & 33.3    & 70.5  \\
Color\_V\_Animal\_3      & 28.6   & 84.3    & 99.1  \\
Color\_V\_Animal\_5      & 17.6   & 85.2    & 99.1  \\
Concept\_V\_Object\_3    & 19.1   & 77.6    & 99.1  \\
Concept\_V\_Object\_5    & 17.1   & 88.1    & 97.1  \\
Conll2003\_Location      & 9.7    & 87.2    & 94.5  \\
Conll2003\_Organization  & 9.3    & 77.1    & 92.0  \\
Conll2003\_Person        & 9.7    & 92.1    & 97.6  \\
Fruit\_V\_Animal\_3      & 29.1   & 87.1    & 98.6  \\
Fruit\_V\_Animal\_5      & 13.3   & 93.3    & 98.6  \\
Object\_V\_Concept\_3    & 27.6   & 81.4    & 98.6  \\
Object\_V\_Concept\_5    & 14.3   & 81.0    & 97.6  \\
Squad\_Val               & 58.4   & 84.9    & 88.9  \\
Verb\_V\_Adjective\_3    & 24.3   & 67.6    & 97.1  \\
Verb\_V\_Adjective\_5    & 9.1    & 80.0    & 98.1  \\ \midrule
Average                  & 22.9   & \underline{73.5}    & \textbf{91.1}         \\
\bottomrule
\end{tabular}}
\caption{Extractive task results}
\end{subtable}
\caption{\textbf{Task-wise performance on 57 FV tasks using Mistral-7B-v0.3.} Our method is evaluated along with 0-shot and 10-shot ICL baselines. (a) Results on 29 abstractive tasks. (b) Results on 28 extractive tasks. The best results are shown in \textbf{bold}, and the second-best results are \underline{underlined}.}
\label{tab:main_results_mistral-7b-v0.3}
\end{table*}

\clearpage
\begin{table*}[ht]
\centering
\setlength{\tabcolsep}{4.5pt}
\begin{subtable}[t]{0.495\textwidth}
\centering
\scriptsize
\resizebox{0.99\textwidth}{!}{\begin{tabular}{lccc}
\toprule
\textbf{Task Name}                  & \textbf{0-shot} & \textbf{10-shot} & \textbf{Ours}  \\ \midrule
AG\_News                   & 0.0    & 79.5    & 88.4  \\
Antonym                    & 1.2    & 69.8    & 70.0  \\
Capitalize                 & 31.8   & 99.4    & 100.0 \\
Capitalize\_First\_Letter  & 30.6   & 98.2    & 99.4  \\
Capitalize\_Last\_Letter   & 0.6    & 30.4    & 90.6  \\
Capitalize\_Second\_Letter & 1.8    & 26.1    & 95.8  \\
Commonsense\_QA            & 24.0   & 71.8    & 66.1  \\
Country-Capital            & 4.8    & 88.1    & 88.1  \\
Country-Currency           & 0.0    & 78.6    & 71.4  \\
English-French             & 0.4    & 82.4    & 79.3  \\
English-German             & 1.5    & 75.5    & 60.5  \\
English-Spanish            & 0.3    & 85.4    & 80.0  \\
Landmark-Country           & 0.0    & 84.6    & 80.6  \\
Lowercase\_First\_Letter   & 0.0    & 97.7    & 100.0 \\
Lowercase\_Last\_Letter    & 0.0    & 42.7    & 95.3  \\
National\_Parks            & 1.1    & 80.0    & 75.8  \\
Next\_Capital\_Letter      & 0.0    & 4.1     & 97.7  \\
Next\_Item                 & 0.0    & 97.9    & 95.7  \\
Park-Country               & 0.0    & 84.7    & 80.9  \\
Person-Instrument          & 1.9    & 71.0    & 75.7  \\
Person-Occupation          & 0.6    & 54.7    & 66.3  \\
Person-Sport               & 0.0    & 92.5    & 94.0  \\
Present-Past               & 3.3    & 98.4    & 100.0 \\
Prev\_Item                 & 4.3    & 89.4    & 97.9  \\
Product-Company            & 0.0    & 79.8    & 81.7  \\
Sentiment                  & 0.0    & 94.3    & 94.3  \\
Singular-Plural            & 7.0    & 100.0   & 97.7  \\
Synonym                    & 1.2    & 53.2    & 49.5  \\
Word\_Length               & 0.6    & 53.2    & 62.6  \\ \midrule
Average                    & 4.0    & \underline{74.6}    & \textbf{84.0}   \\
\bottomrule
\end{tabular}}
\caption{Abstractive task results}
\end{subtable}
\hfill
\begin{subtable}[t]{0.495\textwidth}
\centering
\scriptsize
\resizebox{0.99\textwidth}{!}{\begin{tabular}{lccccc}
\toprule
\textbf{Task Name}                & \textbf{0-shot} & \textbf{10-shot} & \textbf{Ours}  \\ \midrule
Adjective\_V\_Verb\_3    & 14.8   & 78.6    & 99.1  \\
Adjective\_V\_Verb\_5    & 1.0    & 83.8    & 98.1  \\
Alphabetically\_First\_3 & 16.7   & 31.9    & 45.2  \\
Alphabetically\_First\_5 & 4.8    & 24.8    & 88.6  \\
Alphabetically\_Last\_3  & 11.4   & 31.4    & 48.1  \\
Alphabetically\_Last\_5  & 6.7    & 22.4    & 42.4  \\
Animal\_V\_Object\_3     & 10.5   & 84.3    & 97.1  \\
Animal\_V\_Object\_5     & 4.3    & 94.8    & 98.1  \\
Choose\_First\_Of\_3     & 41.4   & 99.5    & 99.1  \\
Choose\_First\_Of\_5     & 8.6    & 96.2    & 99.1  \\
Choose\_Last\_Of\_3      & 5.2    & 88.1    & 100.0 \\
Choose\_Last\_Of\_5      & 1.0    & 87.1    & 100.0 \\
Choose\_Middle\_Of\_3    & 5.7    & 45.2    & 96.2  \\
Choose\_Middle\_Of\_5    & 1.9    & 41.0    & 96.7  \\
Color\_V\_Animal\_3      & 14.3   & 82.4    & 100.0 \\
Color\_V\_Animal\_5      & 1.9    & 91.4    & 99.5  \\
Concept\_V\_Object\_3    & 8.1    & 88.1    & 97.6  \\
Concept\_V\_Object\_5    & 5.2    & 90.5    & 99.1  \\
Conll2003\_Location      & 5.6    & 84.0    & 95.2  \\
Conll2003\_Organization  & 9.7    & 79.6    & 92.8  \\
Conll2003\_Person        & 22.3   & 89.9    & 97.7  \\
Fruit\_V\_Animal\_3      & 13.8   & 94.3    & 99.1  \\
Fruit\_V\_Animal\_5      & 1.0    & 97.6    & 100.0 \\
Object\_V\_Concept\_3    & 15.2   & 78.6    & 98.1  \\
Object\_V\_Concept\_5    & 3.3    & 84.3    & 98.1  \\
Squad\_Val               & 60.4   & 87.5    & 88.1  \\
Verb\_V\_Adjective\_3    & 10.0   & 68.1    & 96.7  \\
Verb\_V\_Adjective\_5    & 2.9    & 79.5    & 97.6  \\ \midrule
Average                  & 11.0   & \underline{75.2}    & \textbf{91.7}        \\
\bottomrule
\end{tabular}}
\caption{Extractive task results}
\end{subtable}
\caption{\textbf{Task-wise performance on 57 FV tasks using Mistral-7B-Instruct-v0.3.} Our method is evaluated along with 0-shot and 10-shot ICL baselines. (a) Results on 29 abstractive tasks. (b) Results on 28 extractive tasks. The best results are shown in \textbf{bold}, and the second-best results are \underline{underlined}.}
\label{tab:main_results_mistral-7b-instruct-v0.3}
\end{table*}

\clearpage
\begin{table*}[ht]
\centering
\setlength{\tabcolsep}{4.5pt}
\begin{subtable}[t]{0.495\textwidth}
\centering
\scriptsize
\resizebox{0.99\textwidth}{!}{\begin{tabular}{lccc}
\toprule
\textbf{Task Name} & \multicolumn{1}{l}{\textbf{0-shot}} & \multicolumn{1}{l}{\textbf{10-shot}} & \multicolumn{1}{l}{\textbf{Ours}} \\
\midrule
AG\_News                   & 0.4  & 79.4  & 86.8  \\
Antonym                    & 0.0  & 69.9  & 69.1 \\
Capitalize                 & 5.3  & 99.6  & 100.0   \\
Capitalize\_First\_Letter  & 10.0 & 99.2  & 99.8  \\
Capitalize\_Last\_Letter   & 1.2  & 20.9  & 94.0 \\
Capitalize\_Second\_Letter & 1.2  & 32.1  & 97.4 \\
Commonsense\_QA            & 40.3 & 72.4  & 63.5 \\
Country-Capital            & 4.8  & 96.0  & 92.9 \\
Country-Currency           & 0.0  & 80.2  & 81.7 \\
English-French             & 0.5  & 81.4  & 80.5 \\
English-German             & 1.2  & 76.1  & 68.6 \\
English-Spanish            & 0.2  & 84.4  & 83.9 \\
Landmark-Country           & 0.0  & 91.1  & 86.7 \\
Lowercase\_First\_Letter   & 0.0  & 99.6  & 100.0   \\
Lowercase\_Last\_Letter    & 0.0  & 35.7  & 95.9 \\
National\_Parks            & 0.0  & 84.6  & 80.4 \\
Next\_Capital\_Letter      & 0.6  & 5.5   & 99.2 \\
Next\_Item                 & 2.1  & 97.2  & 97.9 \\
Park-Country               & 0.0  & 88.8  & 83.9 \\
Person-Instrument          & 0.0  & 85.4  & 88.5 \\
Person-Occupation          & 0.0  & 65.9  & 80.6 \\
Person-Sport               & 0.0  & 95.0  & 96.5 \\
Present-Past               & 3.3  & 99.5  & 100.0   \\
Prev\_Item                 & 2.1  & 95.7  & 95.7 \\
Product-Company            & 0.0  & 87.8  & 89.3  \\
Sentiment                  & 0.0  & 95.9  & 96.1 \\
Singular-Plural            & 2.3  & 99.2 & 98.4 \\
Synonym                    & 1.8  & 50.4  & 52.9 \\
Word\_Length               & 0.0  & 33.7  & 83.0 \\
\midrule
Average                    & 2.7  & \underline{75.9}  & \textbf{87.7}  \\
\bottomrule
\end{tabular}}
\caption{Abstractive task results}
\end{subtable}
\hfill
\begin{subtable}[t]{0.495\textwidth}
\centering
\scriptsize
\resizebox{0.99\textwidth}{!}{\begin{tabular}{lccccc}
\toprule
\textbf{Task Name}       & \textbf{0-shot} & \textbf{10-shot} & \textbf{Ours} \\
\midrule
Adjective\_V\_Verb\_3    & 14.3            & 86.7             & 99.7         \\
Adjective\_V\_Verb\_5    & 9.1             & 86.8             & 97.9         \\
Alphabetically\_First\_3 & 21.9            & 42.1             & 56.2         \\
Alphabetically\_First\_5 & 16.7            & 21.1             & 89.7         \\
Alphabetically\_Last\_3  & 16.2            & 34.4             & 48.1          \\
Alphabetically\_Last\_5  & 10.5            & 20.8             & 53.0         \\
Animal\_V\_Object\_3     & 12.4            & 81.3             & 99.2         \\
Animal\_V\_Object\_5     & 19.1            & 80.2             & 98.3         \\
Choose\_First\_Of\_3     & 52.9            & 98.9             & 100.0           \\
Choose\_First\_Of\_5     & 52.4            & 98.3             & 100.0           \\
Choose\_Last\_Of\_3      & 1.0             & 97.0             & 100.0           \\
Choose\_Last\_Of\_5      & 3.8             & 95.9             & 100.0           \\
Choose\_Middle\_Of\_3    & 2.9             & 53.7             & 98.6         \\
Choose\_Middle\_Of\_5    & 4.3             & 33.5             & 91.4         \\
Color\_V\_Animal\_3      & 16.7            & 94.9             & 99.5         \\
Color\_V\_Animal\_5      & 15.7            & 91.3             & 99.2         \\
Concept\_V\_Object\_3    & 14.3            & 83.7             & 99.8         \\
Concept\_V\_Object\_5    & 17.6            & 86.0             & 94.6          \\
Conll2003\_Location      & 21.8            & 87.7             & 94.3         \\
Conll2003\_Organization  & 39.3            & 78.0             & 91.3         \\
Conll2003\_Person        & 12.4            & 92.6             & 97.7         \\
Fruit\_V\_Animal\_3      & 23.3            & 81.9             & 98.6         \\
Fruit\_V\_Animal\_5      & 10.0            & 79.8             & 99.5         \\
Object\_V\_Concept\_3    & 17.6            & 96.7             & 100.0           \\
Object\_V\_Concept\_5    & 5.7             & 94.8             & 98.6         \\
Squad\_Val               & 39.4            & 85.6             & 86.7         \\
Verb\_V\_Adjective\_3    & 11.4            & 95.1             & 98.1          \\
Verb\_V\_Adjective\_5    & 1.9             & 94.6             & 98.3         \\
\midrule
Average                  & 17.3            & \underline{77.6}             & \textbf{92.4}          \\
\bottomrule
\end{tabular}}
\caption{Extractive task results}
\end{subtable}
\caption{\textbf{Task-wise performance on 57 FV tasks using Llama-3.1-8B.} Our method is evaluated along with 0-shot and 10-shot ICL baselines. (a) Results on 29 abstractive tasks. (b) Results on 28 extractive tasks. The best results are shown in \textbf{bold}, and the second-best results are \underline{underlined}.}
\label{tab:main_results_llama-3.1-8b}
\end{table*}

\clearpage
\begin{table*}[ht]
\centering
\setlength{\tabcolsep}{4.5pt}
\begin{subtable}[t]{0.495\textwidth}
\centering
\scriptsize
\resizebox{0.99\textwidth}{!}{\begin{tabular}{lccc}
\toprule
\textbf{Task Name}                  & \textbf{0-shot} & \textbf{10-shot} & \textbf{Ours}  \\ \midrule
AG\_News                   & 0.0    & 77.6    & 90.0  \\
Antonym                    & 0.4    & 70.8    & 71.6  \\
Capitalize                 & 0.6    & 99.4    & 99.4  \\
Capitalize\_First\_Letter  & 2.4    & 100.0   & 100.0 \\
Capitalize\_Last\_Letter   & 2.3    & 49.7    & 95.3  \\
Capitalize\_Second\_Letter & 1.8    & 49.7    & 100.0 \\
Commonsense\_QA            & 71.3   & 74.0    & 72.0  \\
Country-Capital            & 2.4    & 90.5    & 90.5  \\
Country-Currency           & 0.0    & 81.0    & 85.7  \\
English-French             & 0.7    & 83.1    & 82.0  \\
English-German             & 0.7    & 76.7    & 70.1  \\
English-Spanish            & 0.2    & 84.8    & 84.3  \\
Landmark-Country           & 0.0    & 88.0    & 82.9  \\
Lowercase\_First\_Letter   & 0.0    & 100.0   & 99.4  \\
Lowercase\_Last\_Letter    & 0.0    & 60.2    & 97.1  \\
National\_Parks            & 1.1    & 86.3    & 75.8  \\
Next\_Capital\_Letter      & 1.2    & 2.9     & 99.4  \\
Next\_Item                 & 0.0    & 97.9    & 97.9  \\
Park-Country               & 1.3    & 89.2    & 84.1  \\
Person-Instrument          & 0.0    & 82.2    & 88.8  \\
Person-Occupation          & 0.0    & 65.1    & 77.3  \\
Person-Sport               & 0.0    & 95.5    & 97.0  \\
Present-Past               & 1.6    & 100.0   & 100.0 \\
Prev\_Item                 & 4.3    & 91.5    & 95.7  \\
Product-Company            & 1.8    & 84.4    & 84.4  \\
Sentiment                  & 0.0    & 94.7    & 97.1  \\
Singular-Plural            & 4.7    & 100.0   & 95.4  \\
Synonym                    & 32.1   & 53.3    & 55.1  \\
Word\_Length               & 0.0    & 74.3    & 83.6  \\ \midrule
Average                    & 4.5    & \underline{79.4}    & \textbf{88.0}       \\
\bottomrule
\end{tabular}}
\caption{Abstractive task results}
\end{subtable}
\hfill
\begin{subtable}[t]{0.495\textwidth}
\centering
\scriptsize
\resizebox{0.99\textwidth}{!}{\begin{tabular}{lccccc}
\toprule
\textbf{Task Name}                & \textbf{0-shot} & \textbf{10-shot} & \textbf{Ours}  \\ \midrule
Adjective\_V\_Verb\_3    & 14.3   & 87.6    & 100.0 \\
Adjective\_V\_Verb\_5    & 13.3   & 91.0    & 97.1  \\
Alphabetically\_First\_3 & 26.7   & 37.6    & 51.4  \\
Alphabetically\_First\_5 & 18.6   & 22.4    & 92.4  \\
Alphabetically\_Last\_3  & 16.2   & 37.6    & 49.1  \\
Alphabetically\_Last\_5  & 11.0   & 28.6    & 74.3  \\
Animal\_V\_Object\_3     & 21.9   & 95.7    & 99.5  \\
Animal\_V\_Object\_5     & 6.7    & 96.7    & 100.0 \\
Choose\_First\_Of\_3     & 19.5   & 97.6    & 100.0 \\
Choose\_First\_Of\_5     & 9.1    & 97.1    & 100.0 \\
Choose\_Last\_Of\_3      & 29.5   & 93.3    & 100.0 \\
Choose\_Last\_Of\_5      & 26.2   & 94.3    & 100.0 \\
Choose\_Middle\_Of\_3    & 15.2   & 53.3    & 99.1  \\
Choose\_Middle\_Of\_5    & 9.5    & 33.8    & 96.7  \\
Color\_V\_Animal\_3      & 29.5   & 99.5    & 100.0 \\
Color\_V\_Animal\_5      & 8.6    & 97.6    & 100.0 \\
Concept\_V\_Object\_3    & 26.2   & 91.9    & 99.5  \\
Concept\_V\_Object\_5    & 17.1   & 95.2    & 97.1  \\
Conll2003\_Location      & 10.2   & 90.1    & 94.7  \\
Conll2003\_Organization  & 33.6   & 80.6    & 93.7  \\
Conll2003\_Person        & 22.9   & 93.4    & 97.5  \\
Fruit\_V\_Animal\_3      & 4.8    & 99.5    & 99.1  \\
Fruit\_V\_Animal\_5      & 0.5    & 98.1    & 99.5  \\
Object\_V\_Concept\_3    & 26.7   & 94.8    & 98.6  \\
Object\_V\_Concept\_5    & 17.1   & 94.8    & 99.5  \\
Squad\_Val               & 76.6   & 87.8    & 90.1  \\
Verb\_V\_Adjective\_3    & 10.5   & 96.7    & 98.6  \\
Verb\_V\_Adjective\_5    & 8.6    & 97.1    & 99.1  \\ \midrule
Average                  & 18.9   & \underline{81.6}    & \textbf{93.8}                \\
\bottomrule
\end{tabular}}
\caption{Extractive task results}
\end{subtable}
\caption{\textbf{Task-wise performance on 57 FV tasks using Llama-3.1-8B-Instruct.} Our method is evaluated along with 0-shot and 10-shot ICL baselines. (a) Results on 29 abstractive tasks. (b) Results on 28 extractive tasks. The best results are shown in \textbf{bold}, and the second-best results are \underline{underlined}.}
\label{tab:main_results_llama-3.1-8b-instruct}
\end{table*}

\clearpage
\begin{table*}[ht]
\centering
\setlength{\tabcolsep}{4.5pt}
\begin{subtable}[t]{0.495\textwidth}
\centering
\scriptsize
\resizebox{0.99\textwidth}{!}{\begin{tabular}{lccc}
\toprule
\textbf{Task Name}                  & \textbf{0-shot} & \textbf{10-shot} & \textbf{Ours}  \\ \midrule
AG\_News                   & 0.0    & 77.8    & 87.7  \\
Antonym                    & 0.0    & 69.4    & 66.5  \\
Capitalize                 & 19.4   & 99.4    & 98.8  \\
Capitalize\_First\_Letter  & 16.5   & 97.7    & 100.0 \\
Capitalize\_Last\_Letter   & 4.1    & 24.6    & 93.0  \\
Capitalize\_Second\_Letter & 11.5   & 30.3    & 97.6  \\
Commonsense\_QA            & 42.2   & 80.8    & 79.8  \\
Country-Capital            & 4.8    & 88.1    & 88.1  \\
Country-Currency           & 0.0    & 81.0    & 64.3  \\
English-French             & 0.3    & 82.3    & 69.4  \\
English-German             & 0.7    & 73.3    & 57.5  \\
English-Spanish            & 0.2    & 84.4    & 73.9  \\
Landmark-Country           & 0.0    & 81.1    & 77.7  \\
Lowercase\_First\_Letter   & 0.0    & 98.8    & 100.0 \\
Lowercase\_Last\_Letter    & 0.0    & 60.2    & 97.7  \\
National\_Parks            & 0.0    & 79.0    & 70.5  \\
Next\_Capital\_Letter      & 0.6    & 8.8     & 92.4  \\
Next\_Item                 & 0.0    & 95.7    & 95.7  \\
Park-Country               & 0.0    & 80.9    & 70.1  \\
Person-Instrument          & 0.0    & 62.6    & 62.6  \\
Person-Occupation          & 0.0    & 46.5    & 60.5  \\
Person-Sport               & 0.0    & 89.6    & 97.0  \\
Present-Past               & 1.6    & 100.0   & 98.4  \\
Prev\_Item                 & 4.3    & 97.9    & 93.6  \\
Product-Company            & 0.0    & 77.1    & 80.7  \\
Sentiment                  & 0.0    & 95.1    & 95.9  \\
Singular-Plural            & 4.7    & 97.7    & 90.7  \\
Synonym                    & 0.2    & 49.0    & 48.3  \\
Word\_Length               & 0.0    & 67.8    & 73.7  \\ \midrule
Average                    & 3.8    & \underline{75.1}    & \textbf{82.1}   \\
\bottomrule
\end{tabular}}
\caption{Abstractive task results}
\end{subtable}
\hfill
\begin{subtable}[t]{0.495\textwidth}
\centering
\scriptsize
\resizebox{0.99\textwidth}{!}{\begin{tabular}{lccccc}
\toprule
\textbf{Task Name}                & \textbf{0-shot} & \textbf{10-shot} & \textbf{Ours}  \\ \midrule
Adjective\_V\_Verb\_3    & 0.5    & 77.6    & 98.1  \\
Adjective\_V\_Verb\_5    & 0.0    & 87.1    & 98.6  \\
Alphabetically\_First\_3 & 5.2    & 31.0    & 42.4  \\
Alphabetically\_First\_5 & 2.4    & 20.5    & 89.5  \\
Alphabetically\_Last\_3  & 2.9    & 38.1    & 42.4  \\
Alphabetically\_Last\_5  & 2.4    & 20.0    & 80.0  \\
Animal\_V\_Object\_3     & 1.4    & 96.7    & 95.7  \\
Animal\_V\_Object\_5     & 0.5    & 98.1    & 96.7  \\
Choose\_First\_Of\_3     & 5.7    & 95.7    & 100.0 \\
Choose\_First\_Of\_5     & 1.0    & 94.8    & 100.0 \\
Choose\_Last\_Of\_3      & 2.4    & 91.4    & 100.0 \\
Choose\_Last\_Of\_5      & 1.4    & 66.2    & 100.0 \\
Choose\_Middle\_Of\_3    & 2.9    & 62.9    & 99.5  \\
Choose\_Middle\_Of\_5    & 0.5    & 31.4    & 98.1  \\
Color\_V\_Animal\_3      & 2.9    & 99.5    & 100.0 \\
Color\_V\_Animal\_5      & 4.3    & 98.6    & 100.0 \\
Concept\_V\_Object\_3    & 1.0    & 91.9    & 100.0 \\
Concept\_V\_Object\_5    & 0.0    & 90.5    & 95.7  \\
Conll2003\_Location      & 7.7    & 91.0    & 95.1  \\
Conll2003\_Organization  & 15.1   & 83.6    & 90.7  \\
Conll2003\_Person        & 26.8   & 94.0    & 96.6  \\
Fruit\_V\_Animal\_3      & 6.7    & 100.0   & 100.0 \\
Fruit\_V\_Animal\_5      & 5.2    & 99.5    & 99.5  \\
Object\_V\_Concept\_3    & 0.0    & 91.4    & 97.1  \\
Object\_V\_Concept\_5    & 0.0    & 93.8    & 96.7  \\
Squad\_Val               & 38.1   & 90.7    & 88.3  \\
Verb\_V\_Adjective\_3    & 1.9    & 91.9    & 99.5  \\
Verb\_V\_Adjective\_5    & 0.0    & 97.1    & 96.2  \\ \midrule
Average                  & 5.0    & \underline{79.5}    & \textbf{92.7}    \\
\bottomrule
\end{tabular}}
\caption{Extractive task results}
\end{subtable}
\caption{\textbf{Task-wise performance on 57 FV tasks using Qwen3-8B.} Our method is evaluated along with 0-shot and 10-shot ICL baselines. (a) Results on 29 abstractive tasks. (b) Results on 28 extractive tasks. The best results are shown in \textbf{bold}, and the second-best results are \underline{underlined}.}
\label{tab:main_results_qwen3-8b}
\end{table*}
\clearpage
\begin{table*}[ht]
\centering
\setlength{\tabcolsep}{4.5pt}
\begin{subtable}[t]{0.495\textwidth}
\centering
\scriptsize
\resizebox{0.99\textwidth}{!}{\begin{tabular}{lccc}
\toprule
\textbf{Task Name}                  & \textbf{0-shot} & \textbf{10-shot} & \textbf{Ours}  \\ \midrule
AG\_News                   & 0.1    & 81.8    & 88.2  \\
Antonym                    & 8.5    & 67.1    & 65.7  \\
Capitalize                 & 4.7    & 93.5    & 99.4  \\
Capitalize\_First\_Letter  & 6.5    & 97.7    & 100.0 \\
Capitalize\_Last\_Letter   & 7.0    & 24.0    & 95.3  \\
Capitalize\_Second\_Letter & 3.0    & 29.1    & 97.0  \\
Commonsense\_QA            & 38.9   & 86.2    & 84.5  \\
Country-Capital            & 7.1    & 76.2    & 90.5  \\
Country-Currency           & 0.0    & 81.0    & 81.0  \\
English-French             & 0.4    & 80.2    & 77.2  \\
English-German             & 0.5    & 76.1    & 66.2  \\
English-Spanish            & 0.0    & 82.9    & 80.1  \\
Landmark-Country           & 0.0    & 84.0    & 81.1  \\
Lowercase\_First\_Letter   & 0.6    & 89.5    & 99.4  \\
Lowercase\_Last\_Letter    & 0.6    & 39.8    & 95.3  \\
National\_Parks            & 0.0    & 83.2    & 79.0  \\
Next\_Capital\_Letter      & 2.3    & 2.9     & 98.3  \\
Next\_Item                 & 0.0    & 85.1    & 95.7  \\
Park-Country               & 0.0    & 84.1    & 77.7  \\
Person-Instrument          & 0.0    & 56.1    & 71.0  \\
Person-Occupation          & 0.0    & 36.6    & 69.2  \\
Person-Sport               & 0.0    & 88.1    & 95.5  \\
Present-Past               & 0.0    & 83.6    & 98.4  \\
Prev\_Item                 & 0.0    & 78.7    & 97.9  \\
Product-Company            & 0.0    & 83.5    & 84.4  \\
Sentiment                  & 2.0    & 93.9    & 92.2  \\
Singular-Plural            & 2.3    & 97.7    & 95.4  \\
Synonym                    & 1.2    & 47.5    & 50.2  \\
Word\_Length               & 0.0    & 77.8    & 73.1  \\ \midrule
Average                    & 3.0    & \underline{72.0}    & \textbf{85.5}   \\
\bottomrule
\end{tabular}}
\caption{Abstractive task results}
\end{subtable}
\hfill
\begin{subtable}[t]{0.495\textwidth}
\centering
\scriptsize
\resizebox{0.99\textwidth}{!}{\begin{tabular}{lccccc}
\toprule
\textbf{Task Name}                & \textbf{0-shot} & \textbf{10-shot} & \textbf{Ours}  \\ \midrule
Adjective\_V\_Verb\_3    & 6.2    & 78.1    & 99.5  \\
Adjective\_V\_Verb\_5    & 4.8    & 81.0    & 98.1  \\
Alphabetically\_First\_3 & 7.1    & 38.1    & 97.6  \\
Alphabetically\_First\_5 & 3.3    & 22.9    & 92.9  \\
Alphabetically\_Last\_3  & 5.2    & 36.2    & 43.3  \\
Alphabetically\_Last\_5  & 2.4    & 17.6    & 71.0  \\
Animal\_V\_Object\_3     & 11.0   & 97.6    & 99.1  \\
Animal\_V\_Object\_5     & 11.0   & 98.6    & 98.6  \\
Choose\_First\_Of\_3     & 14.8   & 95.7    & 100.0 \\
Choose\_First\_Of\_5     & 6.7    & 95.2    & 100.0 \\
Choose\_Last\_Of\_3      & 1.4    & 92.9    & 100.0 \\
Choose\_Last\_Of\_5      & 2.9    & 85.7    & 100.0 \\
Choose\_Middle\_Of\_3    & 1.4    & 56.2    & 100.0 \\
Choose\_Middle\_Of\_5    & 0.0    & 32.9    & 98.1  \\
Color\_V\_Animal\_3      & 4.8    & 96.2    & 100.0 \\
Color\_V\_Animal\_5      & 8.6    & 97.6    & 99.5  \\
Concept\_V\_Object\_3    & 1.9    & 88.1    & 99.1  \\
Concept\_V\_Object\_5    & 1.0    & 97.1    & 98.1  \\
Conll2003\_Location      & 2.1    & 89.4    & 95.7  \\
Conll2003\_Organization  & 8.7    & 83.8    & 92.3  \\
Conll2003\_Person        & 15.1   & 93.8    & 96.8  \\
Fruit\_V\_Animal\_3      & 5.2    & 100.0   & 100.0 \\
Fruit\_V\_Animal\_5      & 6.2    & 100.0   & 99.5  \\
Object\_V\_Concept\_3    & 4.3    & 96.2    & 99.5  \\
Object\_V\_Concept\_5    & 2.9    & 94.3    & 98.6  \\
Squad\_Val               & 31.6   & 90.8    & 90.5  \\
Verb\_V\_Adjective\_3    & 0.5    & 91.4    & 100.0 \\
Verb\_V\_Adjective\_5    & 1.4    & 98.6    & 99.5  \\ \midrule
Average                  & 6.2    & \underline{80.2}    & \textbf{95.3}  \\
\bottomrule
\end{tabular}}
\caption{Extractive task results}
\end{subtable}
\caption{\textbf{Task-wise performance on 57 FV tasks using Qwen3-32B.} Our method is evaluated along with 0-shot and 10-shot ICL baselines. (a) Results on 29 abstractive tasks. (b) Results on 28 extractive tasks. The best results are shown in \textbf{bold}, and the second-best results are \underline{underlined}.}
\label{tab:main_results_qwen3-32b}
\end{table*}
\clearpage
\begin{table*}[ht]
\centering
\setlength{\tabcolsep}{4.5pt}
\begin{subtable}[t]{0.495\textwidth}
\centering
\scriptsize
\resizebox{0.99\textwidth}{!}{\begin{tabular}{lccc}
\toprule
\textbf{Task Name}                  & \textbf{0-shot} & \textbf{10-shot} & \textbf{Ours}  \\ \midrule
AG\_News                   & 0.3  & 81.5  & 89.9  \\
Antonym                    & 3.8  & 70.4  & 67.3  \\
Capitalize                 & 8.8  & 99.4  & 100.0 \\
Capitalize\_First\_Letter  & 8.2  & 97.7  & 100.0 \\
Capitalize\_Last\_Letter   & 0.6  & 39.2  & 91.8  \\
Capitalize\_Second\_Letter & 0.0  & 32.1  & 95.8  \\
Commonsense\_QA            & 39.5 & 73.9  & 61.8  \\
Country-Capital            & 4.8  & 90.5  & 85.7  \\
Country-Currency           & 0.0  & 83.3  & 83.3  \\
English-French             & 0.2  & 84.3  & 82.0  \\
English-German             & 0.8  & 78.2  & 74.4  \\
English-Spanish            & 0.2  & 86.4  & 87.6  \\
Landmark-Country           & 0.0  & 90.3  & 85.1  \\
Lowercase\_First\_Letter   & 0.0  & 93.6  & 100.0 \\
Lowercase\_Last\_Letter    & 0.0  & 46.2  & 94.7  \\
National\_Parks            & 1.1  & 85.3  & 79.0  \\
Next\_Capital\_Letter      & 1.8  & 5.3   & 98.3  \\
Next\_Item                 & 0.0  & 97.9  & 97.9  \\
Park-Country               & 0.0  & 91.7  & 87.9  \\
Person-Instrument          & 0.0  & 84.1  & 87.9  \\
Person-Occupation          & 0.0  & 77.9  & 82.6  \\
Person-Sport               & 0.0  & 95.5  & 98.5  \\
Present-Past               & 1.6  & 100.0 & 100.0 \\
Prev\_Item                 & 2.1  & 97.9  & 97.9  \\
Product-Company            & 0.0  & 89.9  & 88.1  \\
Sentiment                  & 0.0  & 96.3  & 95.1  \\
Singular-Plural            & 2.3  & 100.0 & 100.0 \\
Synonym                    & 0.7  & 54.6  & 49.5  \\
Word\_Length               & 0.0  & 75.4  & 74.3  \\ \midrule
Average                    & 2.6  & \underline{79.3}  & \textbf{87.5}   \\
\bottomrule
\end{tabular}}
\caption{Abstractive task results}
\end{subtable}
\hfill
\begin{subtable}[t]{0.495\textwidth}
\centering
\scriptsize
\resizebox{0.99\textwidth}{!}{\begin{tabular}{lccccc}
\toprule
\textbf{Task Name}                & \textbf{0-shot} & \textbf{10-shot} & \textbf{Ours}  \\ \midrule
Adjective\_V\_Verb\_3    & 26.2   & 84.3    & 99.5  \\
Adjective\_V\_Verb\_5    & 18.1   & 83.8    & 97.6  \\
Alphabetically\_First\_3 & 28.1   & 37.1    & 46.2  \\
Alphabetically\_First\_5 & 20.5   & 22.9    & 89.1  \\
Alphabetically\_Last\_3  & 16.2   & 39.1    & 51.0  \\
Alphabetically\_Last\_5  & 13.8   & 20.5    & 46.2  \\
Animal\_V\_Object\_3     & 21.4   & 94.3    & 97.1  \\
Animal\_V\_Object\_5     & 24.8   & 91.9    & 98.1  \\
Choose\_First\_Of\_3     & 65.7   & 99.1    & 100.0 \\
Choose\_First\_Of\_5     & 73.3   & 99.1    & 100.0 \\
Choose\_Last\_Of\_3      & 2.9    & 99.5    & 100.0 \\
Choose\_Last\_Of\_5      & 2.4    & 95.7    & 99.5  \\
Choose\_Middle\_Of\_3    & 4.3    & 50.5    & 98.1  \\
Choose\_Middle\_Of\_5    & 1.4    & 28.1    & 89.5  \\
Color\_V\_Animal\_3      & 22.9   & 97.6    & 100.0 \\
Color\_V\_Animal\_5      & 9.1    & 97.1    & 99.5  \\
Concept\_V\_Object\_3    & 18.6   & 76.2    & 99.1  \\
Concept\_V\_Object\_5    & 13.8   & 86.7    & 95.7  \\
Conll2003\_Location      & 6.8    & 88.6    & 93.7  \\
Conll2003\_Organization  & 12.3   & 77.2    & 92.3  \\
Conll2003\_Person        & 8.1    & 93.8    & 97.9  \\
Fruit\_V\_Animal\_3      & 31.0   & 97.1    & 99.1  \\
Fruit\_V\_Animal\_5      & 6.7    & 97.1    & 99.5  \\
Object\_V\_Concept\_3    & 22.9   & 91.4    & 99.1  \\
Object\_V\_Concept\_5    & 12.9   & 86.7    & 97.1  \\
Squad\_Val               & 58.9   & 86.2    & 87.5  \\
Verb\_V\_Adjective\_3    & 8.1    & 70.5    & 96.7  \\
Verb\_V\_Adjective\_5    & 4.3    & 90.0    & 98.6  \\ \midrule
Average                  & 19.8   & \underline{77.9}    & \textbf{91.7}  \\
\bottomrule
\end{tabular}}
\caption{Extractive task results}
\end{subtable}
\caption{\textbf{Task-wise performance on 57 FV tasks using Mixtral-8x7B-v0.1.} Our method is evaluated along with 0-shot and 10-shot ICL baselines. (a) Results on 29 abstractive tasks. (b) Results on 28 extractive tasks. The best results are shown in \textbf{bold}, and the second-best results are \underline{underlined}.}
\label{tab:main_results_mixtral-8x7B-v0.1}
\end{table*}
\clearpage
\begin{table*}[ht]
\centering
\setlength{\tabcolsep}{4.5pt}
\begin{subtable}[t]{0.495\textwidth}
\centering
\scriptsize
\resizebox{0.99\textwidth}{!}{\begin{tabular}{lccc}
\toprule
\textbf{Task Name}                  & \textbf{0-shot} & \textbf{10-shot} & \textbf{Ours}  \\ \midrule
AG\_News                   & 0.0    & 81.3    & 89.2  \\
Antonym                    & 0.0    & 72.4    & 67.7  \\
Capitalize                 & 4.7    & 99.4    & 100.0 \\
Capitalize\_First\_Letter  & 8.2    & 99.4    & 100.0 \\
Capitalize\_Last\_Letter   & 0.6    & 25.7    & 87.1  \\
Capitalize\_Second\_Letter & 0.6    & 27.9    & 92.1  \\
Commonsense\_QA            & 56.8   & 73.9    & 67.5  \\
Country-Capital            & 4.8    & 90.5    & 83.3  \\
Country-Currency           & 0.0    & 73.8    & 81.0  \\
English-French             & 0.0    & 84.7    & 82.0  \\
English-German             & 0.2    & 77.2    & 71.5  \\
English-Spanish            & 0.1    & 85.8    & 85.4  \\
Landmark-Country           & 0.0    & 92.0    & 82.3  \\
Lowercase\_First\_Letter   & 0.0    & 97.1    & 100.0 \\
Lowercase\_Last\_Letter    & 0.0    & 48.0    & 97.7  \\
National\_Parks            & 2.1    & 80.0    & 76.8  \\
Next\_Capital\_Letter      & 0.6    & 5.3     & 99.4  \\
Next\_Item                 & 0.0    & 97.9    & 95.7  \\
Park-Country               & 0.0    & 90.5    & 87.9  \\
Person-Instrument          & 0.0    & 84.1    & 88.8  \\
Person-Occupation          & 0.0    & 74.4    & 82.6  \\
Person-Sport               & 0.0    & 95.5    & 98.5  \\
Present-Past               & 0.0    & 100.0   & 100.0 \\
Prev\_Item                 & 0.0    & 93.6    & 95.7  \\
Product-Company            & 0.0    & 91.7    & 88.1  \\
Sentiment                  & 0.0    & 94.7    & 93.9  \\
Singular-Plural            & 0.0    & 100.0   & 100.0 \\
Synonym                    & 0.2    & 51.7    & 47.9  \\
Word\_Length               & 0.0    & 58.5    & 65.5  \\ \midrule
Average                    & 2.7    & \underline{77.5}    & \textbf{86.5}   \\
\bottomrule
\end{tabular}}
\caption{Abstractive task results}
\end{subtable}
\hfill
\begin{subtable}[t]{0.495\textwidth}
\centering
\scriptsize
\resizebox{0.99\textwidth}{!}{\begin{tabular}{lccccc}
\toprule
\textbf{Task Name}                & \textbf{0-shot} & \textbf{10-shot} & \textbf{Ours}  \\ \midrule
Adjective\_V\_Verb\_3    & 30.0   & 87.6    & 97.1  \\
Adjective\_V\_Verb\_5    & 17.6   & 89.5    & 98.1  \\
Alphabetically\_First\_3 & 21.9   & 39.1    & 39.1  \\
Alphabetically\_First\_5 & 14.3   & 24.8    & 77.6  \\
Alphabetically\_Last\_3  & 14.8   & 30.5    & 48.6  \\
Alphabetically\_Last\_5  & 8.6    & 24.3    & 42.9  \\
Animal\_V\_Object\_3     & 12.9   & 93.3    & 97.6  \\
Animal\_V\_Object\_5     & 8.1    & 95.7    & 98.6  \\
Choose\_First\_Of\_3     & 59.5   & 98.1    & 98.6  \\
Choose\_First\_Of\_5     & 35.7   & 94.8    & 99.1  \\
Choose\_Last\_Of\_3      & 5.7    & 97.1    & 100.0 \\
Choose\_Last\_Of\_5      & 6.7    & 95.7    & 100.0 \\
Choose\_Middle\_Of\_3    & 1.9    & 53.3    & 97.1  \\
Choose\_Middle\_Of\_5    & 2.4    & 28.1    & 79.1  \\
Color\_V\_Animal\_3      & 20.0   & 98.6    & 100.0 \\
Color\_V\_Animal\_5      & 1.0    & 96.7    & 99.1  \\
Concept\_V\_Object\_3    & 16.7   & 86.2    & 99.1  \\
Concept\_V\_Object\_5    & 7.1    & 92.9    & 96.7  \\
Conll2003\_Location      & 6.5    & 89.4    & 94.1  \\
Conll2003\_Organization  & 7.6    & 80.6    & 93.9  \\
Conll2003\_Person        & 20.8   & 92.6    & 96.6  \\
Fruit\_V\_Animal\_3      & 17.1   & 97.1    & 97.1  \\
Fruit\_V\_Animal\_5      & 1.0    & 97.6    & 98.6  \\
Object\_V\_Concept\_3    & 13.8   & 92.4    & 97.1  \\
Object\_V\_Concept\_5    & 6.7    & 87.6    & 96.2  \\
Squad\_Val               & 59.3   & 84.8    & 87.3  \\
Verb\_V\_Adjective\_3    & 14.8   & 71.4    & 97.6  \\
Verb\_V\_Adjective\_5    & 2.4    & 85.7    & 100.0 \\ \midrule
Average                  & 15.5   & \underline{78.8}    & \textbf{90.2}   \\
\bottomrule
\end{tabular}}
\caption{Extractive task results}
\end{subtable}
\caption{\textbf{Task-wise performance on 57 FV tasks using Mixtral-8x7B-Instruct-v0.1.} Our method is evaluated along with 0-shot and 10-shot ICL baselines. (a) Results on 29 abstractive tasks. (b) Results on 28 extractive tasks. The best results are shown in \textbf{bold}, and the second-best results are \underline{underlined}.}
\label{tab:main_results_mixtral-8x7B-instruct-v0.1}
\end{table*}
\clearpage
\begin{table*}[ht]
\centering
\setlength{\tabcolsep}{4.5pt}
\begin{subtable}[t]{0.495\textwidth}
\centering
\scriptsize
\resizebox{0.99\textwidth}{!}{\begin{tabular}{lccc}
\toprule
\textbf{Task Name}                  & \textbf{0-shot} & \textbf{10-shot} & \textbf{Ours}  \\ \midrule
AG\_News                   & 0.4    & 84.3    & 91.0  \\
Antonym                    & 16.7   & 71.6    & 71.4  \\
Capitalize                 & 0.0    & 99.4    & 100.0 \\
Capitalize\_First\_Letter  & 0.6    & 100.0   & 100.0 \\
Capitalize\_Last\_Letter   & 0.0    & 35.1    & 97.7  \\
Capitalize\_Second\_Letter & 0.0    & 37.6    & 98.2  \\
Commonsense\_QA            & 31.1   & 78.7    & 73.9  \\
Country-Capital            & 4.8    & 92.9    & 92.9  \\
Country-Currency           & 0.0    & 78.6    & 83.3  \\
English-French             & 0.3    & 85.5    & 85.6  \\
English-German             & 1.0    & 81.5    & 80.0  \\
English-Spanish            & 0.3    & 89.8    & 89.2  \\
Landmark-Country           & 0.0    & 89.1    & 84.0  \\
Lowercase\_First\_Letter   & 0.0    & 98.8    & 100.0 \\
Lowercase\_Last\_Letter    & 0.0    & 42.7    & 99.4  \\
National\_Parks            & 20.0   & 81.1    & 75.8  \\
Next\_Capital\_Letter      & 0.6    & 9.4     & 100.0 \\
Next\_Item                 & 4.3    & 95.7    & 95.7  \\
Park-Country               & 48.4   & 91.7    & 86.0  \\
Person-Instrument          & 0.0    & 79.4    & 83.2  \\
Person-Occupation          & 0.0    & 66.9    & 83.7  \\
Person-Sport               & 0.0    & 97.0    & 98.5  \\
Present-Past               & 1.6    & 100.0   & 100.0 \\
Prev\_Item                 & 2.1    & 97.9    & 97.9  \\
Product-Company            & 1.8    & 90.8    & 88.1  \\
Sentiment                  & 0.0    & 98.0    & 96.3  \\
Singular-Plural            & 2.3    & 100.0   & 97.7  \\
Synonym                    & 2.7    & 55.6    & 60.3  \\
Word\_Length               & 0.0    & 77.2    & 87.1  \\ \midrule
Average                    & 4.8    & \underline{79.5}    & \textbf{89.5}    \\
\bottomrule
\end{tabular}}
\caption{Abstractive task results}
\end{subtable}
\hfill
\begin{subtable}[t]{0.495\textwidth}
\centering
\scriptsize
\resizebox{0.99\textwidth}{!}{\begin{tabular}{lccccc}
\toprule
\textbf{Task Name}                & \textbf{0-shot} & \textbf{10-shot} & \textbf{Ours}  \\ \midrule
Adjective\_V\_Verb\_3    & 29.1   & 89.1    & 100.0 \\
Adjective\_V\_Verb\_5    & 17.1   & 86.7    & 100.0 \\
Alphabetically\_First\_3 & 30.0   & 37.1    & 98.6  \\
Alphabetically\_First\_5 & 22.4   & 30.5    & 96.7  \\
Alphabetically\_Last\_3  & 23.3   & 34.8    & 62.9  \\
Alphabetically\_Last\_5  & 12.9   & 23.8    & 93.8  \\
Animal\_V\_Object\_3     & 22.4   & 98.6    & 97.6  \\
Animal\_V\_Object\_5     & 21.9   & 97.1    & 99.1  \\
Choose\_First\_Of\_3     & 81.9   & 100.0   & 100.0 \\
Choose\_First\_Of\_5     & 88.6   & 100.0   & 100.0 \\
Choose\_Last\_Of\_3      & 0.0    & 94.8    & 100.0 \\
Choose\_Last\_Of\_5      & 0.0    & 99.1    & 100.0 \\
Choose\_Middle\_Of\_3    & 0.5    & 68.1    & 98.6  \\
Choose\_Middle\_Of\_5    & 0.0    & 36.2    & 98.1  \\
Color\_V\_Animal\_3      & 26.7   & 99.5    & 100.0 \\
Color\_V\_Animal\_5      & 13.3   & 98.6    & 100.0 \\
Concept\_V\_Object\_3    & 20.5   & 93.3    & 99.1  \\
Concept\_V\_Object\_5    & 18.1   & 91.9    & 97.6  \\
Conll2003\_Location      & 20.6   & 92.3    & 96.8  \\
Conll2003\_Organization  & 36.9   & 85.1    & 93.6  \\
Conll2003\_Person        & 13.7   & 95.0    & 98.7  \\
Fruit\_V\_Animal\_3      & 17.6   & 100.0   & 99.5  \\
Fruit\_V\_Animal\_5      & 6.7    & 99.5    & 99.5  \\
Object\_V\_Concept\_3    & 21.4   & 98.6    & 99.1  \\
Object\_V\_Concept\_5    & 15.7   & 89.1    & 99.1  \\
Squad\_Val               & 48.8   & 88.4    & 90.4  \\
Verb\_V\_Adjective\_3    & 22.4   & 90.5    & 100.0 \\
Verb\_V\_Adjective\_5    & 12.9   & 96.7    & 99.5  \\ \midrule
Average                  & 23.0   & \underline{82.7}    & \textbf{97.1}  \\
\bottomrule
\end{tabular}}
\caption{Extractive task results}
\end{subtable}
\caption{\textbf{Task-wise performance on 57 FV tasks using Llama-3.1-70B.} Our method is evaluated along with 0-shot and 10-shot ICL baselines. (a) Results on 29 abstractive tasks. (b) Results on 28 extractive tasks. The best results are shown in \textbf{bold}, and the second-best results are \underline{underlined}.}
\label{tab:main_results_llama-3.1-70b}
\end{table*}
\clearpage
\begin{table*}[ht]
\centering
\setlength{\tabcolsep}{4.5pt}
\begin{subtable}[t]{0.495\textwidth}
\centering
\scriptsize
\resizebox{0.99\textwidth}{!}{\begin{tabular}{lccc}
\toprule
\textbf{Task Name}       & \textbf{0-shot} & \textbf{10-shot} & \textbf{Ours} \\
\midrule
AG\_News                   & 0.4    & 83.6    & 91.4  \\
Antonym                    & 23.4   & 70.6    & 71.0  \\
Capitalize                 & 4.1    & 100.0   & 99.4  \\
Capitalize\_First\_Letter  & 3.5    & 100.0   & 100.0 \\
Capitalize\_Last\_Letter   & 0.6    & 54.4    & 97.7  \\
Capitalize\_Second\_Letter & 0.6    & 41.8    & 98.8  \\
Commonsense\_QA            & 76.0   & 80.8    & 77.6  \\
Country-Capital            & 7.1    & 90.5    & 88.1  \\
Country-Currency           & 0.0    & 76.2    & 85.7  \\
English-French             & 0.3    & 86.2    & 86.3  \\
English-German             & 1.4    & 81.2    & 80.4  \\
English-Spanish            & 0.5    & 88.6    & 88.2  \\
Landmark-Country           & 1.7    & 88.0    & 84.6  \\
Lowercase\_First\_Letter   & 0.0    & 100.0   & 100.0 \\
Lowercase\_Last\_Letter    & 0.0    & 77.8    & 98.3  \\
National\_Parks            & 6.3    & 83.2    & 76.8  \\
Next\_Capital\_Letter      & 1.2    & 18.1    & 98.8  \\
Next\_Item                 & 6.4    & 97.9    & 95.7  \\
Park-Country               & 3.8    & 89.8    & 84.1  \\
Person-Instrument          & 0.0    & 76.6    & 81.3  \\
Person-Occupation          & 0.0    & 63.4    & 72.7  \\
Person-Sport               & 0.0    & 97.0    & 98.5  \\
Present-Past               & 3.3    & 93.4    & 100.0 \\
Prev\_Item                 & 8.5    & 97.9    & 97.9  \\
Product-Company            & 1.8    & 87.2    & 88.1  \\
Sentiment                  & 0.0    & 94.3    & 95.9  \\
Singular-Plural            & 4.7    & 100.0   & 100.0 \\
Synonym                    & 9.4    & 55.0    & 57.1  \\
Word\_Length               & 0.0    & 87.1    & 83.6  \\ \midrule
Average                    & 5.7    & \underline{81.4}    & \textbf{88.9}     \\
\bottomrule
\end{tabular}}
\caption{Abstractive task results}
\end{subtable}
\hfill
\begin{subtable}[t]{0.495\textwidth}
\centering
\scriptsize
\resizebox{0.99\textwidth}{!}{\begin{tabular}{lccccc}
\toprule
\textbf{Task Name}       & \textbf{0-shot} & \textbf{10-shot} & \textbf{Ours} \\
\midrule
Adjective\_V\_Verb\_3    & 26.7   & 94.8    & 99.5  \\
Adjective\_V\_Verb\_5    & 8.1    & 95.7    & 99.5  \\
Alphabetically\_First\_3 & 18.1   & 37.6    & 96.7  \\
Alphabetically\_First\_5 & 6.2    & 25.7    & 96.2  \\
Alphabetically\_Last\_3  & 14.3   & 38.6    & 93.3  \\
Alphabetically\_Last\_5  & 0.5    & 21.0    & 78.6  \\
Animal\_V\_Object\_3     & 5.2    & 99.5    & 99.1  \\
Animal\_V\_Object\_5     & 1.0    & 100.0   & 99.5  \\
Choose\_First\_Of\_3     & 27.6   & 98.6    & 100.0 \\
Choose\_First\_Of\_5     & 17.6   & 98.6    & 100.0 \\
Choose\_Last\_Of\_3      & 5.2    & 94.8    & 100.0 \\
Choose\_Last\_Of\_5      & 2.9    & 92.4    & 100.0 \\
Choose\_Middle\_Of\_3    & 1.4    & 60.5    & 100.0 \\
Choose\_Middle\_Of\_5    & 1.4    & 31.9    & 99.5  \\
Color\_V\_Animal\_3      & 3.8    & 99.1    & 100.0 \\
Color\_V\_Animal\_5      & 0.5    & 98.1    & 100.0 \\
Concept\_V\_Object\_3    & 3.8    & 96.2    & 100.0 \\
Concept\_V\_Object\_5    & 1.4    & 97.1    & 99.1  \\
Conll2003\_Location      & 24.3   & 92.1    & 97.1  \\
Conll2003\_Organization  & 40.9   & 81.4    & 94.2  \\
Conll2003\_Person        & 21.9   & 96.9    & 98.1  \\
Fruit\_V\_Animal\_3      & 5.2    & 100.0   & 99.5  \\
Fruit\_V\_Animal\_5      & 1.4    & 100.0   & 99.5  \\
Object\_V\_Concept\_3    & 13.3   & 99.1    & 99.1  \\
Object\_V\_Concept\_5    & 4.8    & 94.8    & 100.0 \\
Squad\_Val               & 66.8   & 88.9    & 91.4  \\
Verb\_V\_Adjective\_3    & 11.9   & 94.3    & 99.1  \\
Verb\_V\_Adjective\_5    & 4.3    & 97.1    & 99.5  \\ \midrule
Average                  & 12.2   & \underline{83.0}    & \textbf{97.8}   \\
\bottomrule
\end{tabular}}
\caption{Extractive task results}
\end{subtable}
\caption{\textbf{Task-wise performance on 57 FV tasks using Llama-3.1-70B-Instruct.} Our method is evaluated along with 0-shot and 10-shot ICL baselines. (a) Results on 29 abstractive tasks. (b) Results on 28 extractive tasks. The best results are shown in \textbf{bold}, and the second-best results are \underline{underlined}.}
\label{tab:main_results_llama-3.1-70b-instruct}
\end{table*}
\clearpage

\section{Extended results on optimized soft head-selection values}
\label{sec:Extended analysis of task-relevant head identification}

\subsection{Optimized soft head-selection values for all 57 FV tasks}
\label{subsec:Optimized soft head-selection values for all 57 tasks}
\begin{figure*}[h]
\begin{center}
    \includegraphics[width=0.97\linewidth]{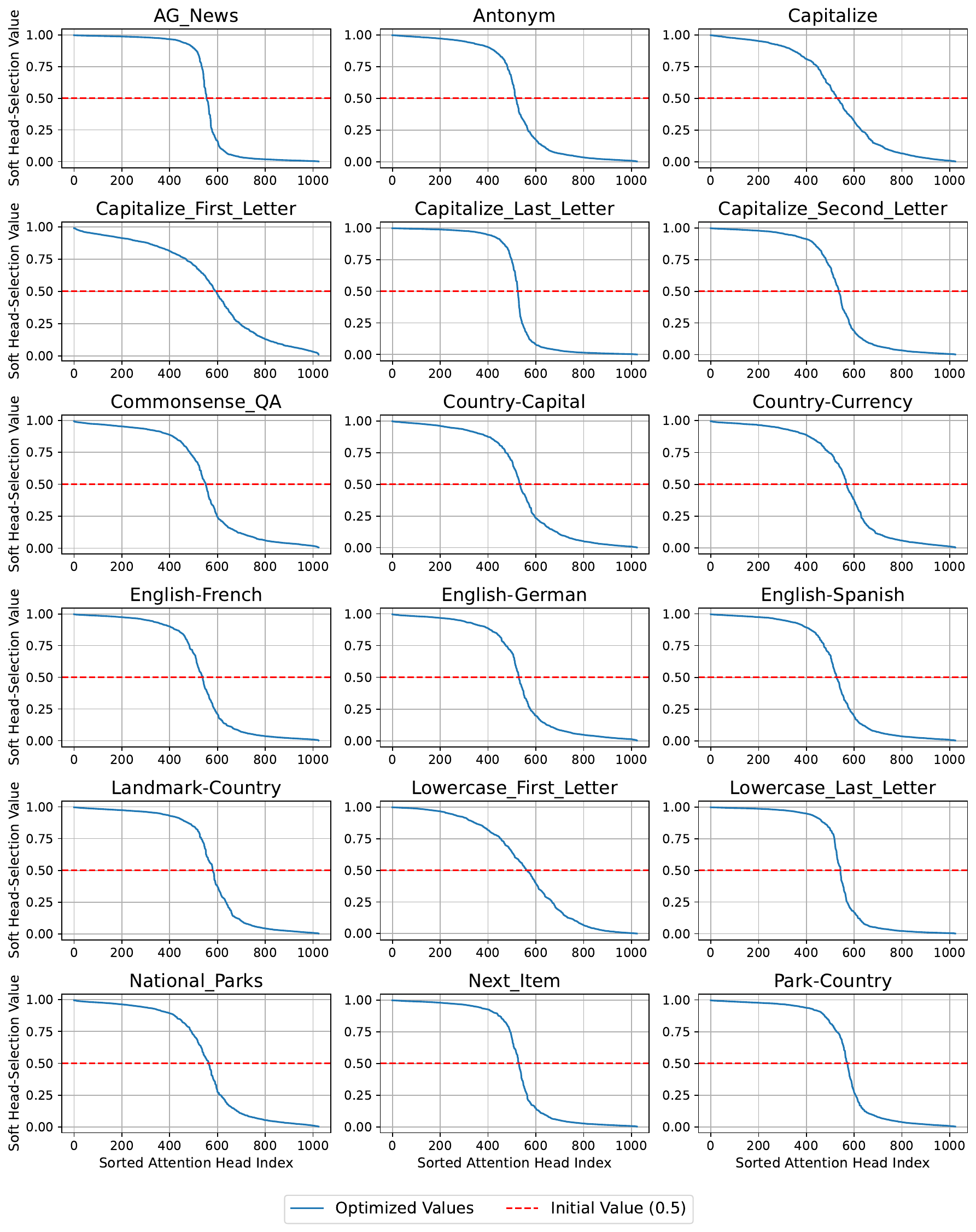}
\end{center}
\vspace{-2mm}
\caption{\textbf{Optimized values of the soft head-selection parameters for 57 FV tasks~(Part 1 of 3).} Each plot shows the optimized values of the soft head-selection parameters for all 1024 attention heads in Llama-3.1-8B, sorted in descending order. Dashed lines indicate the initial value of 0.5 assigned to all selection parameters at the start of training. Plots for the remaining tasks are provided in Figures~\ref{fig:analysis_head_visualization_line_plot_grid_2}-\ref{fig:analysis_head_visualization_line_plot_grid_3}.
}
\label{fig:analysis_head_visualization_line_plot_grid_1}
\end{figure*}

\begin{figure*}[h]
\begin{center}
    \includegraphics[width=0.97\linewidth]{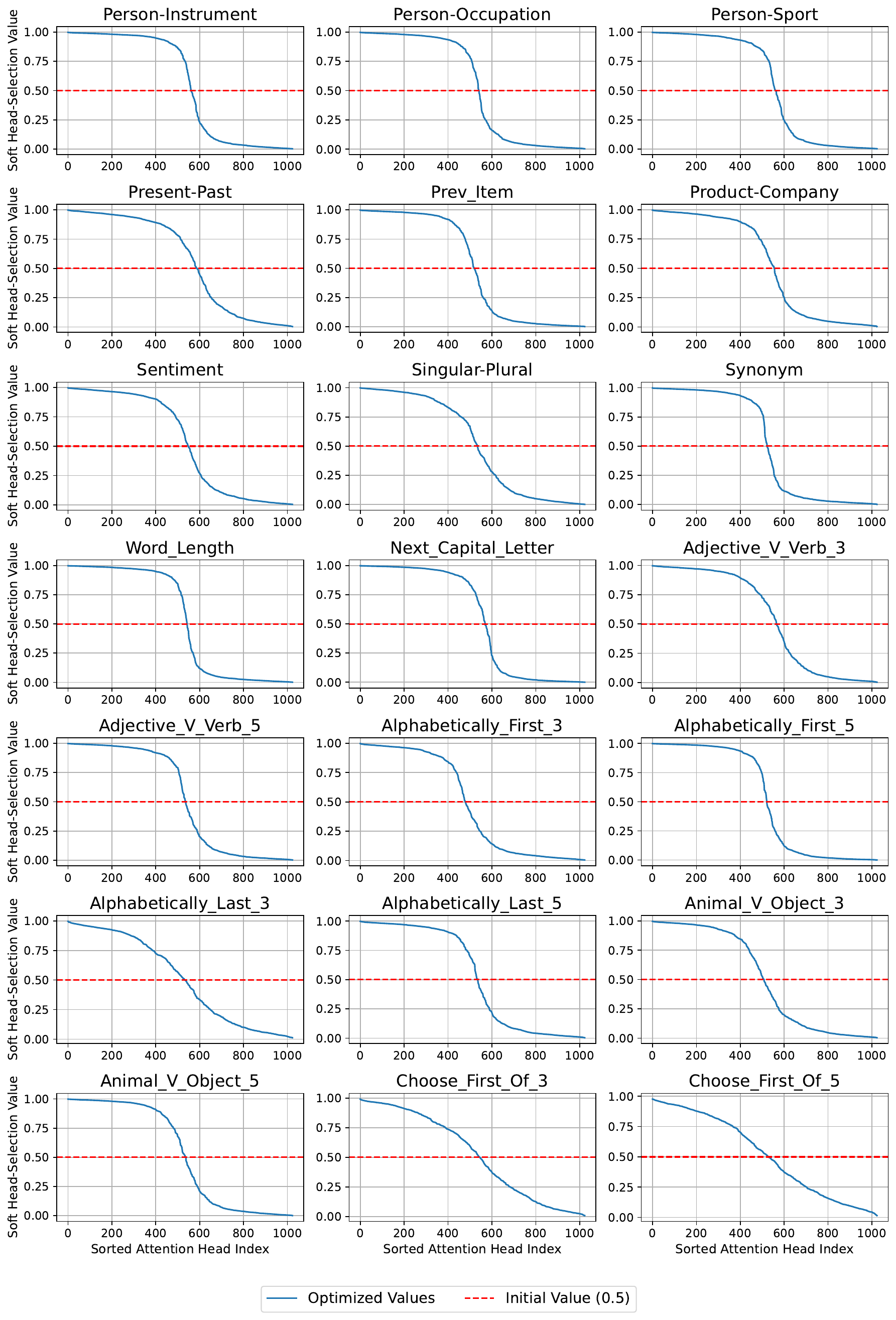}
\end{center}
\vspace{-2mm}
\caption{\textbf{Optimized values of the soft head-selection parameters for 57 FV tasks~(Part 2 of 3).} This figure continues from Figure~\ref{fig:analysis_head_visualization_line_plot_grid_1}. Each plot shows the optimized values of the soft head-selection parameters for all 1024 attention heads in Llama-3.1-8B, sorted in descending order. Dashed lines indicate the initial value of 0.5 assigned to all selection parameters at the start of training. Plots for the remaining tasks are provided in Figure~\ref{fig:analysis_head_visualization_line_plot_grid_3}.
}
\label{fig:analysis_head_visualization_line_plot_grid_2}
\end{figure*}

\begin{figure*}[h]
\begin{center}
    \includegraphics[width=0.97\linewidth]{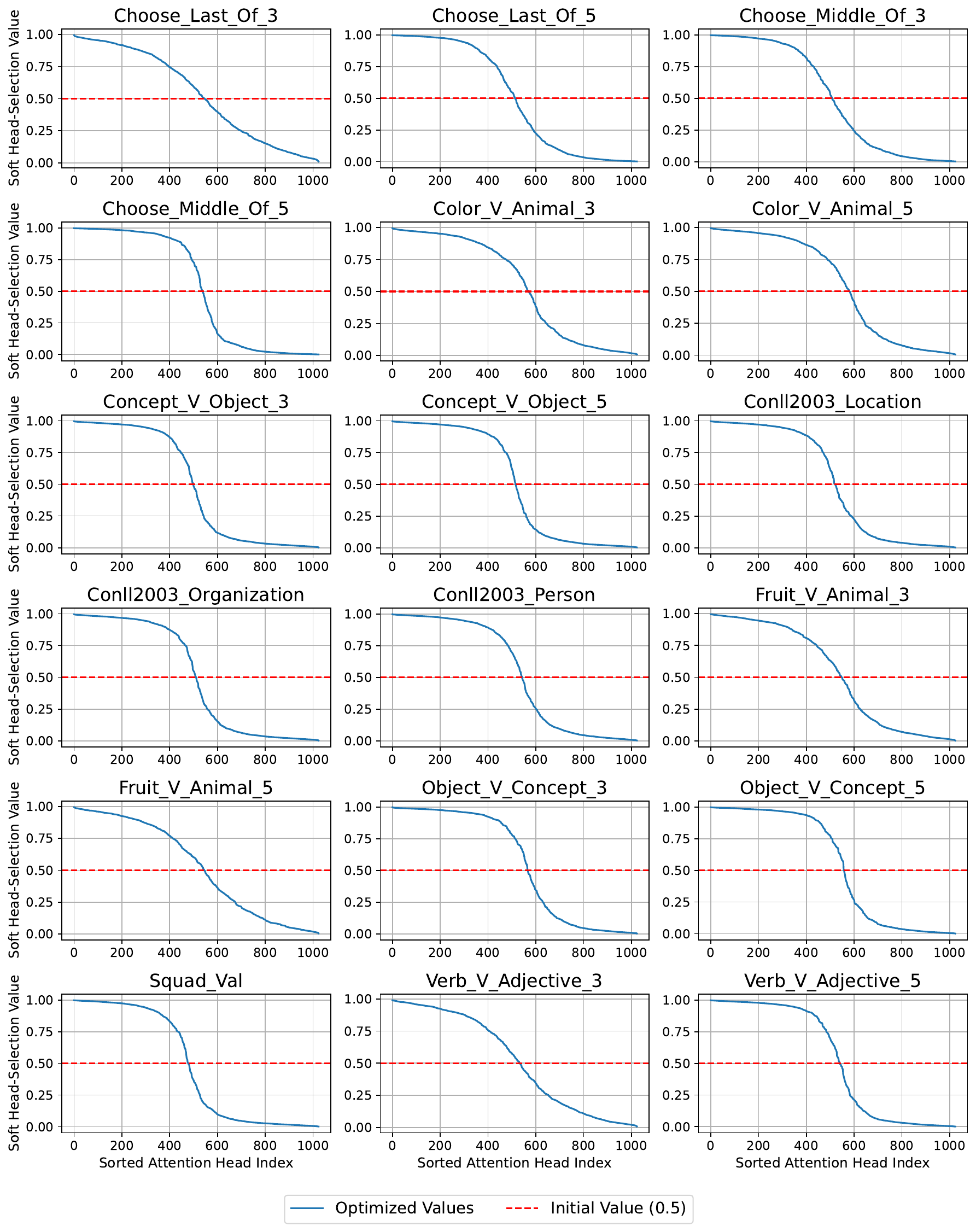}
\end{center}
\vspace{-2mm}
\caption{\textbf{Optimized values of the soft head-selection parameters for 57 FV tasks~(Part 3 of 3).} This figure concludes the series from Figures~\ref{fig:analysis_head_visualization_line_plot_grid_1}-\ref{fig:analysis_head_visualization_line_plot_grid_2}. Each plot shows the optimized values of the soft head-selection parameters for all 1024 attention heads in Llama-3.1-8B, sorted in descending order. Dashed lines indicate the initial value of 0.5 assigned to all selection parameters at the start of training. 
}
\label{fig:analysis_head_visualization_line_plot_grid_3}
\end{figure*}

In this section, we present extended results of Figure~\ref{fig:analysis_optimized_head_selection_line_plot} in Section~\ref{subsec:identification of task-relevant attention heads}, showing the optimized values of the soft head-selection parameters for all 57 FV tasks using Llama-3.1-8B. The full set of results is provided in
Figures~\ref{fig:analysis_head_visualization_line_plot_grid_1}-\ref{fig:analysis_head_visualization_line_plot_grid_3}
\clearpage

\clearpage

\clearpage

\subsection{Optimized soft head-selection values for larger language models}
\label{subsec:Optimized soft head-selection values for larger language models}
\begin{figure*}[h]
\begin{center}
    \includegraphics[width=0.8\linewidth]{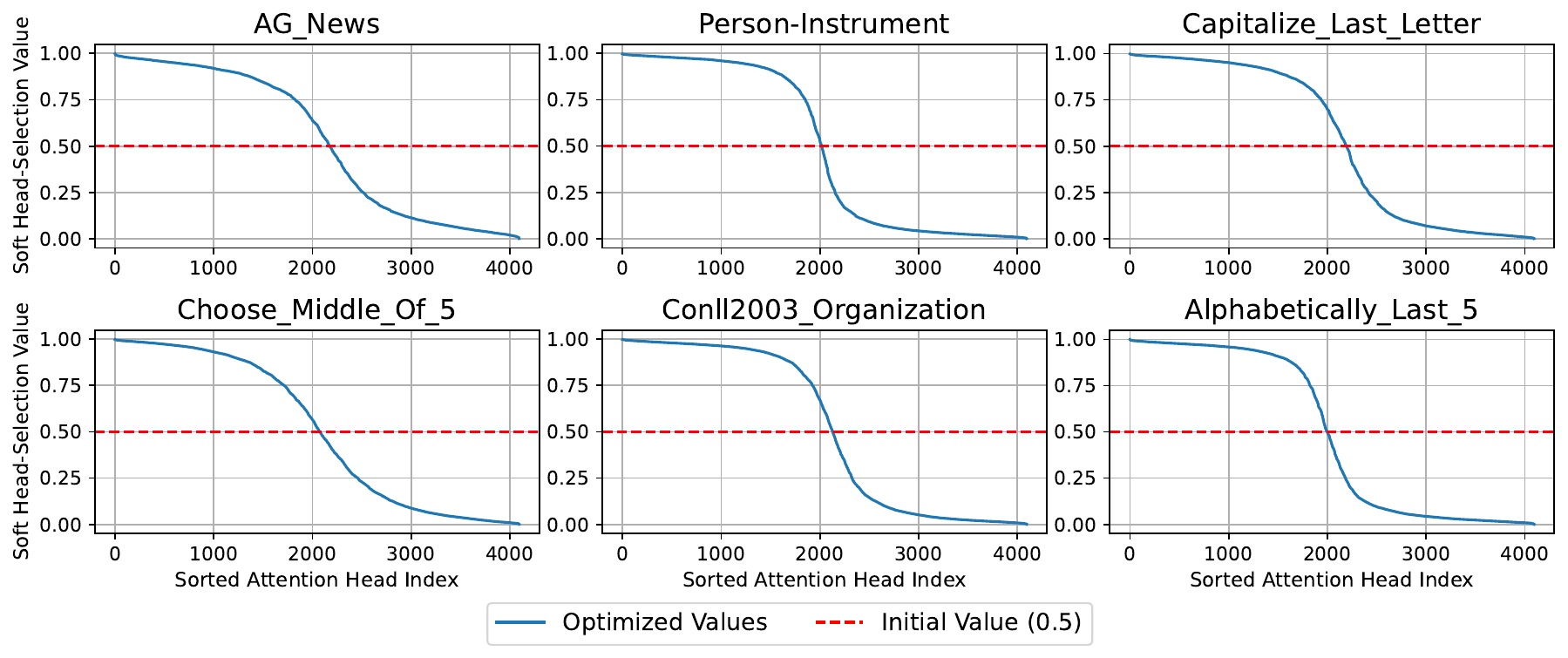}
\end{center}
\vspace{-3mm}
\caption{\textbf{Optimized values of the soft head-selection parameters for six FV tasks using Qwen3-32B.} Each plot shows the optimized values of the soft head-selection parameters for all 4096 attention heads in Qwen3-32B, sorted in descending order. Dashed lines indicate the initial value of 0.5 assigned to all selection parameters at the start of training.  
}
\label{fig:analysis_head_visualization_line_plot_Qwen3-32B_combined_6}
\end{figure*}

\begin{figure*}[h]
\begin{center}
    \includegraphics[width=0.8\linewidth]{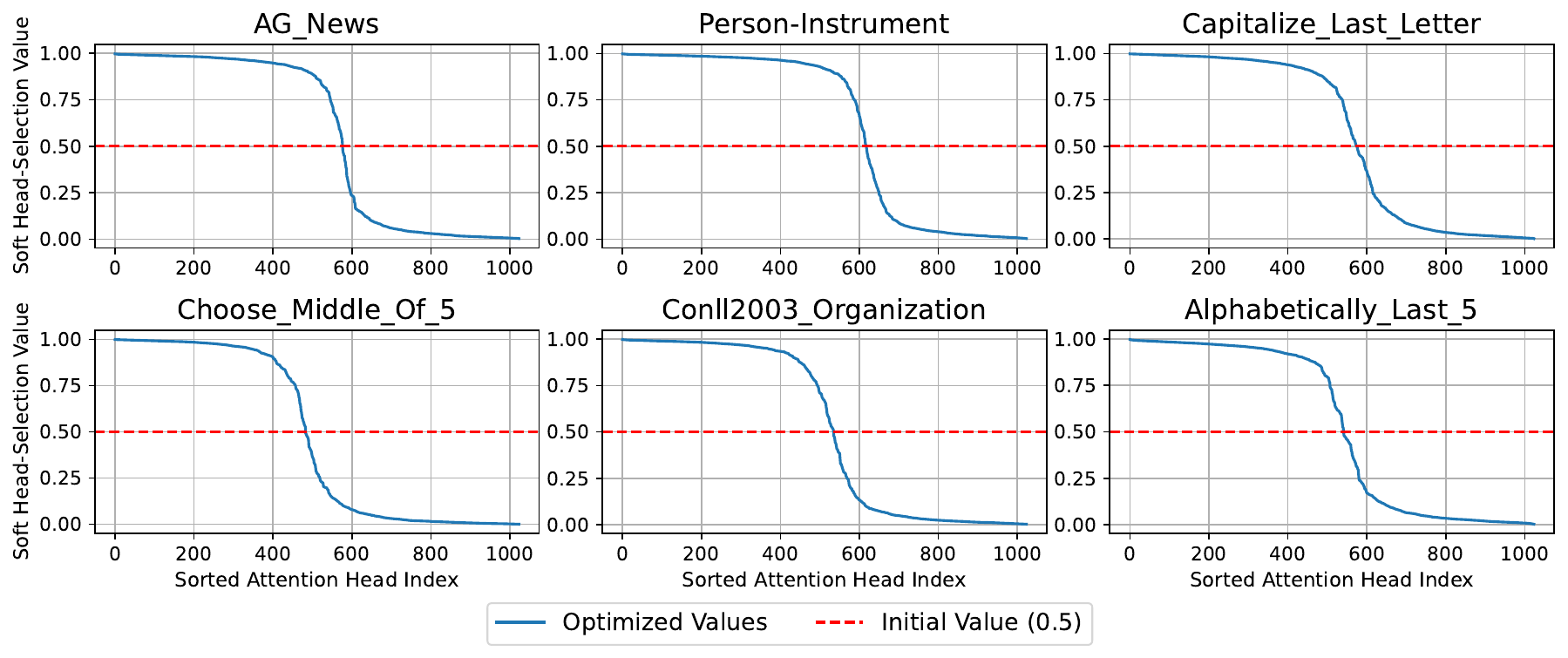}
\end{center}
\vspace{-3mm}
\caption{\textbf{Optimized values of the soft head-selection parameters for six FV tasks using Mixtral-8x7B-v0.1.} Each plot shows the optimized values of the soft head-selection parameters for all 1024 attention heads in Mixtral-8x7B-v0.1, sorted in descending order. Dashed lines indicate the initial value of 0.5 assigned to all selection parameters at the start of training. 
}
\label{fig:analysis_head_visualization_line_plot_Mixtral-8x7B-v0.1_combined_6}
\end{figure*}

\begin{figure*}[h]
\begin{center}
    \includegraphics[width=0.8\linewidth]{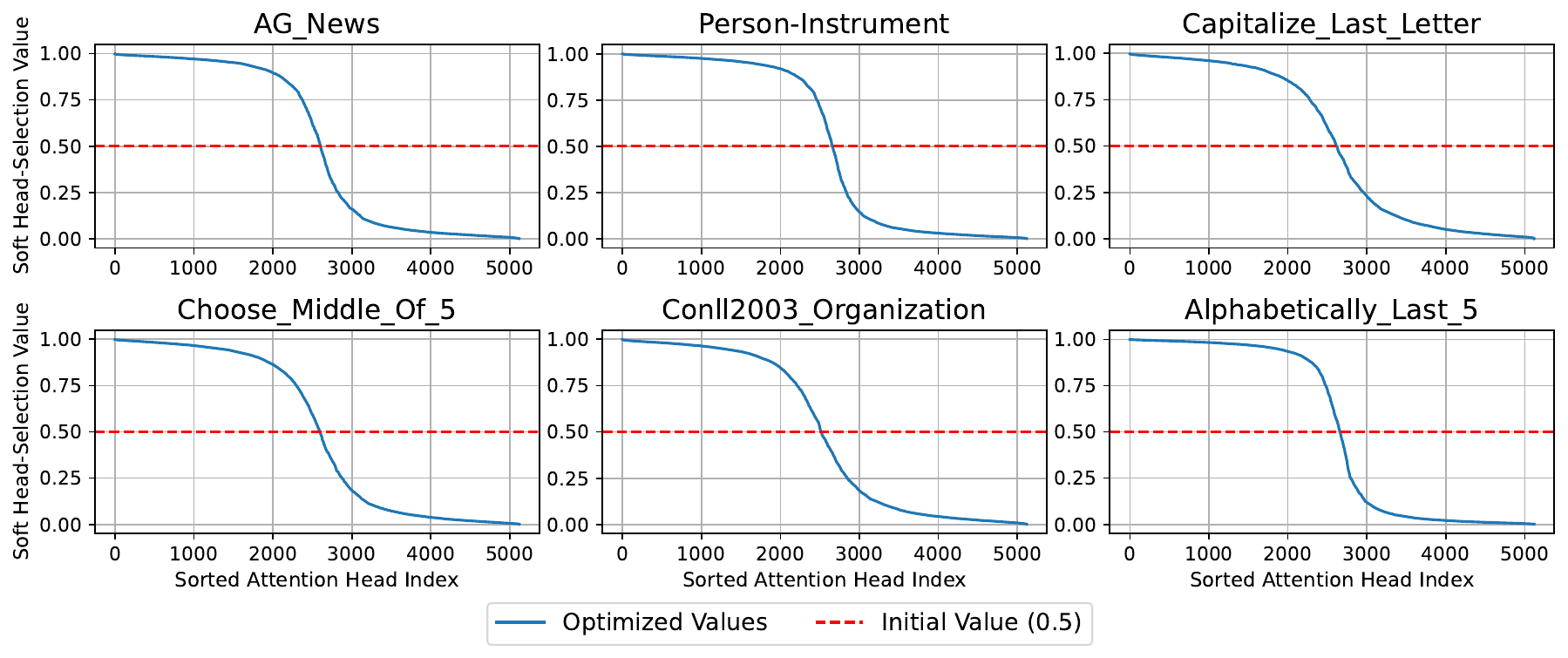}
\end{center}
\vspace{-3mm}
\caption{\textbf{Optimized values of the soft head-selection parameters for six FV tasks using Llama-3.1-70B.} Each plot shows the optimized values of the soft head-selection parameters for all 5120 attention heads in Llama-3.1-70B, sorted in descending order. Dashed lines indicate the initial value of 0.5 assigned to all selection parameters at the start of training. 
}
\label{fig:analysis_head_visualization_line_plot_Llama-3.1-70B_combined_6}
\end{figure*}

Figures~\ref{fig:analysis_head_visualization_line_plot_Qwen3-32B_combined_6}-\ref{fig:analysis_head_visualization_line_plot_Llama-3.1-70B_combined_6} present the optimized values of the soft head-selection parameters for the larger models Qwen3-32B, Mixtral-8x7B-v0.1, and Llama-3.1-70B across six selected tasks from the FV benchmark. The plots show consistent overall patterns, closely matching those observed for Llama-3.1-8B in Section~\ref{subsec:identification of task-relevant attention heads}.

\vspace{-1mm}

\vspace{-1mm}
\clearpage

\section{Additional results on inter-task activation patching analysis}
\label{sec:additional results on cross-task analysis}
\subsection{Additional results on inter-task activation patching analysis for Llama-3.1-8B}
\label{subsec:addtional results for Llama-3.1-8B}
In Table~\ref{tab:Appendix_additional_cross-task_accuracy}, we present the results of the inter-task activation patching analysis for 12 additional FV tasks using Llama-3.1-8B, following the procedure described in Section~\ref{subsec:Task-specific vs. task-agnostic: which better explains head roles?}. 
As explained in Section~\ref{subsec:Task-specific vs. task-agnostic: which better explains head roles?}, 
we fix both the task of zero-shot inference and the task embeddings used for activation patching, and vary only the head-selection parameters across tasks. 
Specifically, we choose a single \emph{evaluation task}, from which both the zero-shot inference input and the task embeddings are derived, and apply head-selection parameters learned from different tasks in the FV benchmark.
For each evaluation task, we report the top-3 and bottom-3 head-selection tasks ranked by post-patching performance. 
The overall trends are consistent with those reported in Table~\ref{tab:Analysis_cross-task_accuracy} of Section~\ref{subsec:Task-specific vs. task-agnostic: which better explains head roles?}.


\begin{table*}[ht]
\centering
\resizebox{\textwidth}{!}{\begin{tabular}{l|l|l|l}
\toprule
\textbf{Evaluation Task}          & \textbf{Task Description}                                                                                                     & \textbf{\begin{tabular}[c]{@{}r@{}}Top-3 Head-Selection Tasks\\~(Accuracy, \%) \end{tabular}}                                                                                                     & \textbf{\begin{tabular}[c]{@{}r@{}}Bottom-3 Head-Selection Tasks\\~(Accuracy, \%) \end{tabular}}                                                                                               \\ \toprule
Adjective\_V\_Verb\_3     & \begin{tabular}[c]{@{}l@{}}Select the adjective \\ from a list of 3 words \\ (1 adjective, 2 verbs)\end{tabular}                       & \begin{tabular}[c]{@{}l@{}}Adjective\_V\_Verb\_5 (99.0)\\ Adjective\_V\_Verb\_3 (97.1)\\ Fruit\_V\_Animal\_5 (89.0)\end{tabular}       & \begin{tabular}[c]{@{}l@{}}Verb\_V\_Adjective\_3 (1.0)\\ Verb\_V\_Adjective\_5 (7.6)\\ Squad\_Val (15.7)\end{tabular}                \\ \midrule
Verb\_V\_Adjective\_3     & \begin{tabular}[c]{@{}l@{}}Select the verb \\ from a list of 3 words \\ (1 verb, 2 adjectives)\end{tabular}                            & \begin{tabular}[c]{@{}l@{}}Verb\_V\_Adjective\_5 (99.5)\\ Verb\_V\_Adjective\_3 (97.1)\\ Color\_V\_Animal\_5 (80.5)\end{tabular}       & \begin{tabular}[c]{@{}l@{}}Adjective\_V\_Verb\_3 (0.5)\\ Adjective\_V\_Verb\_5 (4.8)\\ Synonym (14.8)\end{tabular}                   \\ \midrule
Alphabetically\_First\_3  & \begin{tabular}[c]{@{}l@{}}Select the word that comes\\ first in alphabetical order\\ from a list of 3 words\end{tabular}              & \begin{tabular}[c]{@{}l@{}}Alphabetically\_First\_5 (86.7)\\ Alphabetically\_First\_3 (52.9)\\ Next\_Item (45.2)\end{tabular}          & \begin{tabular}[c]{@{}l@{}}Person\_Occupation (23.3)\\ Alphabetically\_Last\_5 (23.3)\\ Alphabetically\_Last\_3 (25.2)\end{tabular}  \\ \midrule
Alphabetically\_Last\_3   & \begin{tabular}[c]{@{}l@{}}Select the word that comes\\ last in alphabetical order\\ from a list of 3 words\end{tabular}               & \begin{tabular}[c]{@{}l@{}}Alphabetically\_Last\_5 (50.0)\\ Alphabetically\_Last\_3 (46.2)\\ Concept\_V\_Object\_5 (36.7)\end{tabular} & \begin{tabular}[c]{@{}l@{}}Alphabetically\_First\_5 (1.0)\\ Alphabetically\_First\_3 (15.7)\\ Person\_Occupation (22.4)\end{tabular} \\ \midrule
Concept\_V\_Object\_3     & \begin{tabular}[c]{@{}l@{}}Select the concept from\\ a list of 3 words (1 abstract\\ concept, 2 concrete entities)\end{tabular}        & \begin{tabular}[c]{@{}l@{}}Concept\_V\_Object\_3 (99.0)\\ Concept\_V\_Object\_5 (98.1)\\ Animal\_V\_Object\_5 (71.9)\end{tabular}      & \begin{tabular}[c]{@{}l@{}}Object\_V\_Concept\_5 (0.5)\\ Object\_V\_Concept\_3 (2.9)\\ Park\_Country (13.3)\end{tabular}             \\ \midrule
Concept\_V\_Object\_5     & \begin{tabular}[c]{@{}l@{}}Select the concept from\\ a list of 5 words (1 abstract\\ concept, 4 concrete entities)\end{tabular}        & \begin{tabular}[c]{@{}l@{}}Concept\_V\_Object\_3 (95.2)\\ Concept\_V\_Object\_5 (91.9)\\ Animal\_V\_Object\_5 (73.3)\end{tabular}      & \begin{tabular}[c]{@{}l@{}}Object\_V\_Concept\_5 (2.4)\\ Park\_Country (6.2)\\ Object\_V\_Concept\_3 (6.2)\end{tabular}              \\ \midrule
Object\_V\_Concept\_3     & \begin{tabular}[c]{@{}l@{}}Select the concrete entity from\\ a list of 3 words (1 concrete\\ entity, 2 abstract concepts)\end{tabular} & \begin{tabular}[c]{@{}l@{}}Object\_V\_Concept\_3 (100.0)\\ Object\_V\_Concept\_5 (99.0)\\ Color\_V\_Animal\_3 (95.7)\end{tabular}      & \begin{tabular}[c]{@{}l@{}}Concept\_V\_Object\_3 (5.7)\\ Concept\_V\_Object\_5 (8.6)\\ Squad\_Val (15.2)\end{tabular}                \\ \midrule
Object\_V\_Concept\_5     & \begin{tabular}[c]{@{}l@{}}Select the concrete entity from\\ a list of 5 words (1 concrete\\ entity, 4 abstract concepts)\end{tabular} & \begin{tabular}[c]{@{}l@{}}Object\_V\_Concept\_5 (98.1)\\ Object\_V\_Concept\_3 (96.2)\\ Fruit\_V\_Animal\_3 (92.4)\end{tabular}       & \begin{tabular}[c]{@{}l@{}}Concept\_V\_Object\_3 (3.3)\\ Concept\_V\_Object\_5 (3.8)\\ Squad\_Val (10.5)\end{tabular}                \\ \midrule
Capitalize\_First\_Letter & \begin{tabular}[c]{@{}l@{}}Generate the first letter\\ of a given word \\ in captial form\end{tabular}                                 & \begin{tabular}[c]{@{}l@{}}Capitalize (100.0)\\ Capitalize\_First\_Letter (100.0)\\ Lowercase\_First\_Letter (100.0)\end{tabular}      & \begin{tabular}[c]{@{}l@{}}Conll2003\_Organization (0.0)\\ Choose\_Middle\_Of\_3 (0.0)\\ Next\_Item (0.0)\end{tabular}               \\ \midrule
Lowercase\_First\_Letter  & \begin{tabular}[c]{@{}l@{}}Generate the first letter\\ of a given word \\ in lowercase\end{tabular}                                    & \begin{tabular}[c]{@{}l@{}}Capitalize (100.0)\\ Capitalize\_First\_Letter (100.0)\\ English\_French (100.0)\end{tabular}               & \begin{tabular}[c]{@{}l@{}}Conll2003\_Organization (0.0)\\ Conll2003\_Location (0.0)\\ Person\_Occupation (0.0)\end{tabular}         \\ \midrule
Capitalize\_Last\_Letter  & \begin{tabular}[c]{@{}l@{}}Generate the last letter\\ of a given word \\ in captial form\end{tabular}                                  & \begin{tabular}[c]{@{}l@{}}Capitalize\_Last\_Letter (87.7)\\ Lowercase\_Last\_Letter (86.0)\\ Country\_Currency (42.7)\end{tabular}    & \begin{tabular}[c]{@{}l@{}}Choose\_First\_Of\_5 (0.0)\\ Choose\_Middle\_Of\_3 (0.0)\\ Conll2003\_Organization (0.6)\end{tabular}     \\ \midrule
Lowercase\_Last\_Letter   & \begin{tabular}[c]{@{}l@{}}Generate the last letter\\ of a given word \\ in lowercase\end{tabular}                                     & \begin{tabular}[c]{@{}l@{}}Lowercase\_Last\_Letter (94.7)\\ Capitalize\_Last\_Letter (93.6)\\ National\_Parks (50.3)\end{tabular}      & \begin{tabular}[c]{@{}l@{}}Conll2003\_Organization (0.0)\\ Alphabetically\_Last\_3 (0.0)\\ Prev\_Item (0.0)\end{tabular}            \\
\bottomrule
\end{tabular}}
\caption{
\textbf{Inter-task activation patching analysis for 12 additional FV tasks using Llama-3.1-8B.} 
For each evaluation task, we report performance after patching high-$\alpha$ attention heads, where task embeddings are fixed to the evaluation task and head-selection parameters are derived from different head-selection tasks. Among the 57 FV tasks, we report the top-3 and bottom-3 head-selection tasks ranked by post-patching accuracy.
}
\label{tab:Appendix_additional_cross-task_accuracy}
\end{table*}
\clearpage

\subsection{Results on inter-task activation patching analysis for larger language models}
\label{subsec:cross-task results for larger language models}
Tables~\ref{tab:Appendix_cross-task_accuracy_qwen3-32b}-\ref{tab:Appendix_cross-task_accuracy_llama-3.1-70b} present the results of the inter-task activation patching analysis for larger models: Qwen3-32B, Mixtral-8x7B-v0.1, and Llama-3.1-70B, across 19 FV tasks. The overall trends are consistent with those reported for Llama-3.1-8B in Table~\ref{tab:Analysis_cross-task_accuracy} of Section~\ref{subsec:Task-specific vs. task-agnostic: which better explains head roles?} and Table~\ref{tab:Appendix_additional_cross-task_accuracy}.


\clearpage
\begin{table*}[ht]
\centering
\resizebox{0.9\textwidth}{!}{\begin{tabular}{l|l|l|l}
\toprule
\textbf{Evaluation Task}          & \textbf{Task Description}                                                                                                     & \textbf{\begin{tabular}[c]{@{}r@{}}Top-3 Head-Selection Tasks\\~(Accuracy, \%) \end{tabular}}                                                                                                     & \textbf{\begin{tabular}[c]{@{}r@{}}Bottom-3 Head-Selection Tasks\\~(Accuracy, \%) \end{tabular}}                                                                                               \\ \toprule
Adjective\_V\_Verb\_3     & \begin{tabular}[c]{@{}l@{}}Select the adjective \\ from a list of 3 words \\ (1 adjective, 2 verbs)\end{tabular}                       & \begin{tabular}[c]{@{}l@{}}Adjective\_V\_Verb\_3 (99.5)\\ Adjective\_V\_Verb\_5 (99.0)\\ Animal\_V\_Object\_3 (85.7)\end{tabular}       & \begin{tabular}[c]{@{}l@{}}Person\_Occupation (3.3)\\ Sentiment (7.6)\\ Verb\_V\_Adjective\_3 (9.5)\end{tabular}                   \\
\midrule
Adjective\_V\_Verb\_5     & \begin{tabular}[c]{@{}l@{}}Select the only adjective \\ from a list of 5 words \\ (1 adjective, 4 verbs)\end{tabular}                  & \begin{tabular}[c]{@{}l@{}}Adjective\_V\_Verb\_5 (98.1)\\ Adjective\_V\_Verb\_3 (97.1)\\ Animal\_V\_Object\_3 (86.2)\end{tabular}       & \begin{tabular}[c]{@{}l@{}}Person\_Occupation (5.2)\\ Park\_Country (10.0)\\ Sentiment (11.0)\end{tabular}                         \\
\midrule
Verb\_V\_Adjective\_3     & \begin{tabular}[c]{@{}l@{}}Select the verb \\ from a list of 3 words \\ (1 verb, 2 adjectives)\end{tabular}                            & \begin{tabular}[c]{@{}l@{}}Verb\_V\_Adjective\_3 (99.5)\\ Verb\_V\_Adjective\_5 (95.7)\\ Singular\_Plural (81.9)\end{tabular}           & \begin{tabular}[c]{@{}l@{}}Ag\_News (1.4)\\ Person\_Occupation (2.9)\\ Adjective\_V\_Verb\_3 (5.2)\end{tabular}                    \\
\midrule
Verb\_V\_Adjective\_5     & \begin{tabular}[c]{@{}l@{}}Select the only verb \\ from a list of 5 words \\ (1 verb, 4 adjectives)\end{tabular}                       & \begin{tabular}[c]{@{}l@{}}Verb\_V\_Adjective\_5 (99.5)\\ Verb\_V\_Adjective\_3 (99.0)\\ Concept\_V\_Object\_5 (91.4)\end{tabular}      & \begin{tabular}[c]{@{}l@{}}Person\_Occupation (3.8)\\ Ag\_News (5.7)\\ Sentiment (6.2)\end{tabular}                                \\
\midrule
Alphabetically\_First\_3  & \begin{tabular}[c]{@{}l@{}}Select the word that comes\\ first in alphabetical order\\ from a list of 3 words\end{tabular}              & \begin{tabular}[c]{@{}l@{}}Alphabetically\_First\_3 (96.7)\\ Alphabetically\_First\_5 (94.8)\\ Animal\_V\_Object\_3 (41.0)\end{tabular} & \begin{tabular}[c]{@{}l@{}}Person\_Occupation (0.5)\\ Sentiment (2.4)\\ Park\_Country (6.2)\end{tabular}                           \\
\midrule
Alphabetically\_First\_5  & \begin{tabular}[c]{@{}l@{}}Choose the word that comes \\ first in alphabetical order \\ from a list of 5 words\end{tabular}            & \begin{tabular}[c]{@{}l@{}}Alphabetically\_First\_5 (88.1)\\ Alphabetically\_First\_3 (85.2)\\ Antonym (27.6)\end{tabular}              & \begin{tabular}[c]{@{}l@{}}Sentiment (1.9)\\ Person\_Occupation (6.2)\\ Next\_Capital\_Letter (7.1)\end{tabular}                   \\
\midrule
Alphabetically\_Last\_3   & \begin{tabular}[c]{@{}l@{}}Select the word that comes\\ last in alphabetical order\\ from a list of 3 words\end{tabular}               & \begin{tabular}[c]{@{}l@{}}Alphabetically\_Last\_5 (44.8)\\ Alphabetically\_Last\_3 (42.9)\\ Color\_V\_Animal\_3 (38.6)\end{tabular}    & \begin{tabular}[c]{@{}l@{}}Alphabetically\_First\_5 (1.0)\\ Alphabetically\_First\_3 (1.4)\\ Person\_Occupation (3.3)\end{tabular} \\
\midrule
Alphabetically\_Last\_5   & \begin{tabular}[c]{@{}l@{}}Choose the word that comes \\ last in alphabetical order \\ from a list of 5 words\end{tabular}             & \begin{tabular}[c]{@{}l@{}}Alphabetically\_Last\_5 (39.5)\\ Alphabetically\_Last\_3 (29.0)\\ Choose\_Middle\_Of\_5 (25.2)\end{tabular}  & \begin{tabular}[c]{@{}l@{}}Alphabetically\_First\_3 (0.0)\\ Alphabetically\_First\_5 (0.5)\\ Sentiment (5.7)\end{tabular}          \\
\midrule
Concept\_V\_Object\_3     & \begin{tabular}[c]{@{}l@{}}Select the concept from\\ a list of 3 words (1 abstract\\ concept, 2 concrete entities)\end{tabular}        & \begin{tabular}[c]{@{}l@{}}Concept\_V\_Object\_3 (99.0)\\ Concept\_V\_Object\_5 (97.1)\\ Fruit\_V\_Animal\_3 (82.9)\end{tabular}        & \begin{tabular}[c]{@{}l@{}}Sentiment (3.8)\\ Ag\_News (4.8)\\ Person\_Occupation (7.6)\end{tabular}                                \\
\midrule
Concept\_V\_Object\_5     & \begin{tabular}[c]{@{}l@{}}Select the concept from\\ a list of 5 words (1 abstract\\ concept, 4 concrete entities)\end{tabular}        & \begin{tabular}[c]{@{}l@{}}Concept\_V\_Object\_3 (97.1)\\ Concept\_V\_Object\_5 (97.1)\\ Fruit\_V\_Animal\_3 (82.9)\end{tabular}        & \begin{tabular}[c]{@{}l@{}}Sentiment (7.6)\\ Person\_Occupation (11.4)\\ Ag\_News (11.9)\end{tabular}                              \\
\midrule
Object\_V\_Concept\_3     & \begin{tabular}[c]{@{}l@{}}Select the concrete entity from\\ a list of 3 words (1 concrete\\ entity, 2 abstract concepts)\end{tabular} & \begin{tabular}[c]{@{}l@{}}Object\_V\_Concept\_3 (100.0)\\ Object\_V\_Concept\_5 (98.6)\\ Country\_Capital (90.5)\end{tabular}          & \begin{tabular}[c]{@{}l@{}}Person\_Occupation (4.3)\\ Park\_Country (4.8)\\ Ag\_News (7.6)\end{tabular}                            \\
\midrule
Object\_V\_Concept\_5     & \begin{tabular}[c]{@{}l@{}}Select the concrete entity from\\ a list of 5 words (1 concrete\\ entity, 4 abstract concepts)\end{tabular} & \begin{tabular}[c]{@{}l@{}}Object\_V\_Concept\_5 (98.1)\\ Object\_V\_Concept\_3 (97.6)\\ Adjective\_V\_Verb\_5 (86.7)\end{tabular}      & \begin{tabular}[c]{@{}l@{}}Person\_Occupation (1.9)\\ Park\_Country (5.7)\\ Concept\_V\_Object\_3 (8.6)\end{tabular}               \\
\midrule
English\_French           & \begin{tabular}[c]{@{}l@{}}Translate the given \\ English word into French\end{tabular}                                                & \begin{tabular}[c]{@{}l@{}}English\_German (77.4)\\ English\_French (77.3)\\ English\_Spanish (75.5)\end{tabular}                       & \begin{tabular}[c]{@{}l@{}}Alphabetically\_Last\_5 (1.1)\\ Commonsense\_Qa (2.7)\\ Person\_Instrument (4.9)\end{tabular}           \\
\midrule
English\_German           & \begin{tabular}[c]{@{}l@{}}Translate the given \\ English word into German\end{tabular}                                                & \begin{tabular}[c]{@{}l@{}}English\_French (67.8)\\ English\_German (65.0)\\ English\_Spanish (63.3)\end{tabular}                       & \begin{tabular}[c]{@{}l@{}}Adjective\_V\_Verb\_3 (0.9)\\ Person\_Instrument (2.1)\\ Alphabetically\_Last\_5 (3.0)\end{tabular}     \\
\midrule
English\_Spanish          & \begin{tabular}[c]{@{}l@{}}Translate the given \\ English word into Spanish\end{tabular}                                               & \begin{tabular}[c]{@{}l@{}}English\_French (83.2)\\ English\_German (82.3)\\ English\_Spanish (80.2)\end{tabular}                       & \begin{tabular}[c]{@{}l@{}}Alphabetically\_Last\_5 (4.5)\\ Person\_Instrument (8.0)\\ Commonsense\_Qa (15.1)\end{tabular}          \\
\midrule
Capitalize\_First\_Letter & \begin{tabular}[c]{@{}l@{}}Generate the first letter\\ of a given word \\ in captial form\end{tabular}                                 & \begin{tabular}[c]{@{}l@{}}Capitalize (100.0)\\ Capitalize\_First\_Letter (100.0)\\ Country\_Capital (100.0)\end{tabular}               & \begin{tabular}[c]{@{}l@{}}Person\_Occupation (3.5)\\ Prev\_Item (4.1)\\ Next\_Item (7.6)\end{tabular}                             \\
\midrule
Lowercase\_First\_Letter  & \begin{tabular}[c]{@{}l@{}}Generate the first letter\\ of a given word \\ in lowercase\end{tabular}                                    & \begin{tabular}[c]{@{}l@{}}Capitalize\_First\_Letter (100.0)\\ English\_German (100.0)\\ Lowercase\_First\_Letter (100.0)\end{tabular}  & \begin{tabular}[c]{@{}l@{}}Prev\_Item (0.0)\\ Conll2003\_Location (0.0)\\ Next\_Item (0.0)\end{tabular}                            \\
\midrule
Capitalize\_Last\_Letter  & \begin{tabular}[c]{@{}l@{}}Generate the last letter\\ of a given word \\ in captial form\end{tabular}                                  & \begin{tabular}[c]{@{}l@{}}Capitalize\_Last\_Letter (90.1)\\ Lowercase\_Last\_Letter (81.3)\\ Next\_Item (49.7)\end{tabular}            & \begin{tabular}[c]{@{}l@{}}Choose\_Last\_Of\_3 (0.6)\\ English\_French (1.2)\\ Prev\_Item (1.2)\end{tabular}                       \\
\midrule
Lowercase\_Last\_Letter   & \begin{tabular}[c]{@{}l@{}}Generate the last letter\\ of a given word \\ in lowercase\end{tabular}                                     & \begin{tabular}[c]{@{}l@{}}Lowercase\_Last\_Letter (95.9)\\ Verb\_V\_Adjective\_5 (82.5)\\ Capitalize\_Last\_Letter (81.9)\end{tabular} & \begin{tabular}[c]{@{}l@{}}Prev\_Item (0.0)\\ Conll2003\_Location (0.0)\\ Next\_Item (0.0)\end{tabular}                           
\\
\bottomrule
\end{tabular}}
\caption{
\textbf{Inter-task activation patching analysis for 19 FV tasks using Qwen3-32B.} 
For each evaluation task, we report performance after patching high-$\alpha$ attention heads, where task embeddings are fixed to the evaluation task and head-selection parameters are derived from different head-selection tasks. Among the 57 FV tasks, we report the top-3 and bottom-3 head-selection tasks ranked by post-patching accuracy.}
\label{tab:Appendix_cross-task_accuracy_qwen3-32b}
\end{table*}
\clearpage
\begin{table*}[ht]
\centering
\resizebox{0.9\textwidth}{!}{\begin{tabular}{l|l|l|l}
\toprule
\textbf{Evaluation Task}          & \textbf{Task Description}                                                                                                     & \textbf{\begin{tabular}[c]{@{}r@{}}Top-3 Head-Selection Tasks\\~(Accuracy, \%) \end{tabular}}                                                                                                     & \textbf{\begin{tabular}[c]{@{}r@{}}Bottom-3 Head-Selection Tasks\\~(Accuracy, \%) \end{tabular}}                                                                                               \\ \toprule
Adjective\_V\_Verb\_3     & \begin{tabular}[c]{@{}l@{}}Select the adjective \\ from a list of 3 words \\ (1 adjective, 2 verbs)\end{tabular}                       & \begin{tabular}[c]{@{}l@{}}Adjective\_V\_Verb\_3 (98.6)\\ Adjective\_V\_Verb\_5 (98.1)\\ Conll2003\_Organization (84.3)\end{tabular}        & \begin{tabular}[c]{@{}l@{}}Verb\_V\_Adjective\_3 (2.9)\\ Verb\_V\_Adjective\_5 (6.7)\\ Product\_Company (19.0)\end{tabular}           \\
\midrule
Adjective\_V\_Verb\_5     & \begin{tabular}[c]{@{}l@{}}Select the only adjective \\ from a list of 5 words \\ (1 adjective, 4 verbs)\end{tabular}                  & \begin{tabular}[c]{@{}l@{}}Adjective\_V\_Verb\_3 (97.6)\\ Adjective\_V\_Verb\_5 (97.6)\\ Animal\_V\_Object\_5 (71.4)\end{tabular}           & \begin{tabular}[c]{@{}l@{}}Verb\_V\_Adjective\_3 (1.4)\\ Verb\_V\_Adjective\_5 (5.7)\\ Product\_Company (19.5)\end{tabular}           \\
\midrule
Verb\_V\_Adjective\_3     & \begin{tabular}[c]{@{}l@{}}Select the verb \\ from a list of 3 words \\ (1 verb, 2 adjectives)\end{tabular}                            & \begin{tabular}[c]{@{}l@{}}Verb\_V\_Adjective\_5 (96.7)\\ Verb\_V\_Adjective\_3 (96.2)\\ Fruit\_V\_Animal\_3 (65.2)\end{tabular}            & \begin{tabular}[c]{@{}l@{}}Adjective\_V\_Verb\_3 (0.0)\\ Adjective\_V\_Verb\_5 (0.0)\\ Ag\_News (12.9)\end{tabular}                   \\
\midrule
Verb\_V\_Adjective\_5     & \begin{tabular}[c]{@{}l@{}}Select the only verb \\ from a list of 5 words \\ (1 verb, 4 adjectives)\end{tabular}                       & \begin{tabular}[c]{@{}l@{}}Verb\_V\_Adjective\_5 (98.1)\\ Verb\_V\_Adjective\_3 (96.2)\\ Animal\_V\_Object\_5 (67.1)\end{tabular}           & \begin{tabular}[c]{@{}l@{}}Adjective\_V\_Verb\_3 (0.0)\\ Adjective\_V\_Verb\_5 (1.0)\\ English\_French (6.2)\end{tabular}             \\
\midrule
Alphabetically\_First\_3  & \begin{tabular}[c]{@{}l@{}}Select the word that comes\\ first in alphabetical order\\ from a list of 3 words\end{tabular}              & \begin{tabular}[c]{@{}l@{}}Alphabetically\_First\_5 (76.7)\\ Alphabetically\_First\_3 (47.1)\\ Lowercase\_First\_Letter (38.6)\end{tabular} & \begin{tabular}[c]{@{}l@{}}Alphabetically\_Last\_5 (16.7)\\ Ag\_News (23.3)\\ Alphabetically\_Last\_3 (23.8)\end{tabular}             \\
\midrule
Alphabetically\_First\_5  & \begin{tabular}[c]{@{}l@{}}Choose the word that comes \\ first in alphabetical order \\ from a list of 5 words\end{tabular}            & \begin{tabular}[c]{@{}l@{}}Alphabetically\_First\_5 (88.6)\\ Alphabetically\_First\_3 (26.7)\\ Adjective\_V\_Verb\_5 (25.2)\end{tabular}    & \begin{tabular}[c]{@{}l@{}}Alphabetically\_Last\_5 (8.6)\\ Alphabetically\_Last\_3 (10.5)\\ Animal\_V\_Object\_3 (14.3)\end{tabular}  \\
\midrule
Alphabetically\_Last\_3   & \begin{tabular}[c]{@{}l@{}}Select the word that comes\\ last in alphabetical order\\ from a list of 3 words\end{tabular}               & \begin{tabular}[c]{@{}l@{}}Alphabetically\_Last\_5 (51.9)\\ Alphabetically\_Last\_3 (50.5)\\ Squad\_Val (41.0)\end{tabular}                 & \begin{tabular}[c]{@{}l@{}}Alphabetically\_First\_5 (9.0)\\ Alphabetically\_First\_3 (25.7)\\ Person\_Instrument (26.2)\end{tabular}  \\
\midrule
Alphabetically\_Last\_5   & \begin{tabular}[c]{@{}l@{}}Choose the word that comes \\ last in alphabetical order \\ from a list of 5 words\end{tabular}             & \begin{tabular}[c]{@{}l@{}}Alphabetically\_Last\_5 (39.0)\\ Alphabetically\_Last\_3 (33.8)\\ Person\_Sport (26.7)\end{tabular}              & \begin{tabular}[c]{@{}l@{}}Alphabetically\_First\_5 (0.5)\\ Alphabetically\_First\_3 (11.9)\\ National\_Parks (12.9)\end{tabular}     \\
\midrule
Concept\_V\_Object\_3     & \begin{tabular}[c]{@{}l@{}}Select the concept from\\ a list of 3 words (1 abstract\\ concept, 2 concrete entities)\end{tabular}        & \begin{tabular}[c]{@{}l@{}}Concept\_V\_Object\_3 (99.0)\\ Concept\_V\_Object\_5 (99.0)\\ Fruit\_V\_Animal\_3 (62.4)\end{tabular}            & \begin{tabular}[c]{@{}l@{}}Object\_V\_Concept\_3 (2.9)\\ Object\_V\_Concept\_5 (8.6)\\ Person\_Instrument (11.0)\end{tabular}         \\
\midrule
Concept\_V\_Object\_5     & \begin{tabular}[c]{@{}l@{}}Select the concept from\\ a list of 5 words (1 abstract\\ concept, 4 concrete entities)\end{tabular}        & \begin{tabular}[c]{@{}l@{}}Concept\_V\_Object\_5 (96.2)\\ Concept\_V\_Object\_3 (93.8)\\ Animal\_V\_Object\_5 (71.0)\end{tabular}           & \begin{tabular}[c]{@{}l@{}}Object\_V\_Concept\_5 (4.8)\\ Object\_V\_Concept\_3 (6.2)\\ Person\_Instrument (10.0)\end{tabular}         \\
\midrule
Object\_V\_Concept\_3     & \begin{tabular}[c]{@{}l@{}}Select the concrete entity from\\ a list of 3 words (1 concrete\\ entity, 2 abstract concepts)\end{tabular} & \begin{tabular}[c]{@{}l@{}}Object\_V\_Concept\_3 (99.0)\\ Object\_V\_Concept\_5 (97.1)\\ Fruit\_V\_Animal\_5 (76.7)\end{tabular}            & \begin{tabular}[c]{@{}l@{}}Concept\_V\_Object\_5 (2.4)\\ Concept\_V\_Object\_3 (2.9)\\ Adjective\_V\_Verb\_3 (21.4)\end{tabular}      \\
\midrule
Object\_V\_Concept\_5     & \begin{tabular}[c]{@{}l@{}}Select the concrete entity from\\ a list of 5 words (1 concrete\\ entity, 4 abstract concepts)\end{tabular} & \begin{tabular}[c]{@{}l@{}}Object\_V\_Concept\_5 (96.7)\\ Object\_V\_Concept\_3 (96.2)\\ Fruit\_V\_Animal\_5 (81.0)\end{tabular}            & \begin{tabular}[c]{@{}l@{}}Concept\_V\_Object\_3 (1.0)\\ Concept\_V\_Object\_5 (1.9)\\ Adjective\_V\_Verb\_3 (10.0)\end{tabular}      \\
\midrule
English\_French           & \begin{tabular}[c]{@{}l@{}}Translate the given \\ English word into French\end{tabular}                                                & \begin{tabular}[c]{@{}l@{}}English\_Spanish (82.3)\\ English\_French (82.2)\\ Capitalize (80.6)\end{tabular}                                & \begin{tabular}[c]{@{}l@{}}Conll2003\_Organization (2.1)\\ Alphabetically\_First\_5 (3.0)\\ Object\_V\_Concept\_3 (3.3)\end{tabular}  \\
\midrule
English\_German           & \begin{tabular}[c]{@{}l@{}}Translate the given \\ English word into German\end{tabular}                                                & \begin{tabular}[c]{@{}l@{}}English\_Spanish (78.4)\\ English\_French (75.6)\\ English\_German (75.1)\end{tabular}                           & \begin{tabular}[c]{@{}l@{}}Conll2003\_Organization (3.4)\\ Alphabetically\_First\_5 (6.2)\\ Animal\_V\_Object\_5 (8.3)\end{tabular}   \\
\midrule
English\_Spanish          & \begin{tabular}[c]{@{}l@{}}Translate the given \\ English word into Spanish\end{tabular}                                               & \begin{tabular}[c]{@{}l@{}}English\_Spanish (87.1)\\ English\_German (85.3)\\ English\_French (84.0)\end{tabular}                           & \begin{tabular}[c]{@{}l@{}}Conll2003\_Organization (4.9)\\ Alphabetically\_First\_5 (25.2)\\ Animal\_V\_Object\_5 (28.6)\end{tabular} \\
\midrule
Capitalize\_First\_Letter & \begin{tabular}[c]{@{}l@{}}Generate the first letter\\ of a given word \\ in captial form\end{tabular}                                 & \begin{tabular}[c]{@{}l@{}}Capitalize\_First\_Letter (100.0)\\ Lowercase\_First\_Letter (100.0)\\ English\_French (99.4)\end{tabular}       & \begin{tabular}[c]{@{}l@{}}Ag\_News (0.0)\\ Park\_Country (0.0)\\ Landmark\_Country (0.6)\end{tabular}                                \\
\midrule
Lowercase\_First\_Letter  & \begin{tabular}[c]{@{}l@{}}Generate the first letter\\ of a given word \\ in lowercase\end{tabular}                                    & \begin{tabular}[c]{@{}l@{}}Capitalize (100.0)\\ Capitalize\_First\_Letter (100.0)\\ English\_French (100.0)\end{tabular}                    & \begin{tabular}[c]{@{}l@{}}Ag\_News (0.0)\\ Park\_Country (0.6)\\ Conll2003\_Organization (1.2)\end{tabular}                          \\
\midrule
Capitalize\_Last\_Letter  & \begin{tabular}[c]{@{}l@{}}Generate the last letter\\ of a given word \\ in captial form\end{tabular}                                  & \begin{tabular}[c]{@{}l@{}}Capitalize\_Last\_Letter (90.1)\\ Lowercase\_Last\_Letter (81.3)\\ Present\_Past (58.5)\end{tabular}             & \begin{tabular}[c]{@{}l@{}}Choose\_First\_Of\_3 (0.0)\\ Choose\_First\_Of\_5 (0.0)\\ Conll2003\_Organization (0.6)\end{tabular}       \\
\midrule
Lowercase\_Last\_Letter   & \begin{tabular}[c]{@{}l@{}}Generate the last letter\\ of a given word \\ in lowercase\end{tabular}                                     & \begin{tabular}[c]{@{}l@{}}Lowercase\_Last\_Letter (93.0)\\ Capitalize\_Last\_Letter (90.6)\\ Present\_Past (70.2)\end{tabular}             & \begin{tabular}[c]{@{}l@{}}Choose\_First\_Of\_3 (0.0)\\ Choose\_Middle\_Of\_5 (0.0)\\ Prev\_Item (0.0)\end{tabular}         
                  \\            
\bottomrule
\end{tabular}}
\caption{\textbf{Inter-task activation patching analysis for 19 FV tasks using Mixtral-8x7B-v0.1.} For each evaluation task, we report performance after patching high-$\alpha$ attention heads, where task embeddings are fixed to the evaluation task and head-selection parameters are derived from different head-selection tasks. Among the 57 FV tasks, we report the top-3 and bottom-3 head-selection tasks ranked by post-patching accuracy.}
\label{tab:Appendix_cross-task_accuracy_mixtral-8x7b-v0.1}
\end{table*}
\clearpage
\begin{table*}[ht]
\centering
\resizebox{0.9\textwidth}{!}{\begin{tabular}{l|l|l|l}
\toprule
\textbf{Evaluation Task}          & \textbf{Task Description}                                                                                                     & \textbf{\begin{tabular}[c]{@{}r@{}}Top-3 Head-Selection Tasks\\~(Accuracy, \%) \end{tabular}}                                                                                                     & \textbf{\begin{tabular}[c]{@{}r@{}}Bottom-3 Head-Selection Tasks\\~(Accuracy, \%) \end{tabular}}                                                                                               \\ \toprule
Adjective\_V\_Verb\_3     & \begin{tabular}[c]{@{}l@{}}Select the adjective \\ from a list of 3 words \\ (1 adjective, 2 verbs)\end{tabular}                       & \begin{tabular}[c]{@{}l@{}}Adjective\_V\_Verb\_5 (99.5)\\ Adjective\_V\_Verb\_3 (99.0)\\ Object\_V\_Concept\_3 (88.1)\end{tabular}       & \begin{tabular}[c]{@{}l@{}}Verb\_V\_Adjective\_5 (3.8)\\ Conll2003\_Organization (10.0)\\ Verb\_V\_Adjective\_3 (10.5)\end{tabular}     \\
\midrule
Adjective\_V\_Verb\_5     & \begin{tabular}[c]{@{}l@{}}Select the only adjective \\ from a list of 5 words \\ (1 adjective, 4 verbs)\end{tabular}                  & \begin{tabular}[c]{@{}l@{}}Adjective\_V\_Verb\_5 (100.0)\\ Adjective\_V\_Verb\_3 (95.2)\\ Color\_V\_Animal\_3 (79.0)\end{tabular}        & \begin{tabular}[c]{@{}l@{}}Choose\_First\_Of\_3 (4.8)\\ Capitalize\_Second\_Letter (8.1)\\ Verb\_V\_Adjective\_5 (10.0)\end{tabular}    \\
\midrule
Verb\_V\_Adjective\_3     & \begin{tabular}[c]{@{}l@{}}Select the verb \\ from a list of 3 words \\ (1 verb, 2 adjectives)\end{tabular}                            & \begin{tabular}[c]{@{}l@{}}Verb\_V\_Adjective\_3 (100.0)\\ Verb\_V\_Adjective\_5 (99.5)\\ Concept\_V\_Object\_5 (89.0)\end{tabular}      & \begin{tabular}[c]{@{}l@{}}Antonym (7.1)\\ Adjective\_V\_Verb\_3 (8.1)\\ Choose\_First\_Of\_3 (11.9)\end{tabular}                       \\
\midrule
Verb\_V\_Adjective\_5     & \begin{tabular}[c]{@{}l@{}}Select the only verb \\ from a list of 5 words \\ (1 verb, 4 adjectives)\end{tabular}                       & \begin{tabular}[c]{@{}l@{}}Verb\_V\_Adjective\_5 (99.5)\\ Verb\_V\_Adjective\_3 (99.0)\\ Concept\_V\_Object\_5 (92.9)\end{tabular}       & \begin{tabular}[c]{@{}l@{}}Antonym (2.4)\\ Choose\_First\_Of\_3 (2.9)\\ Capitalize\_Second\_Letter (3.8)\end{tabular}                   \\
\midrule
Alphabetically\_First\_3  & \begin{tabular}[c]{@{}l@{}}Select the word that comes\\ first in alphabetical order\\ from a list of 3 words\end{tabular}              & \begin{tabular}[c]{@{}l@{}}Alphabetically\_First\_3 (98.1)\\ Alphabetically\_First\_5 (95.7)\\ Commonsense\_Qa (43.3)\end{tabular}       & \begin{tabular}[c]{@{}l@{}}Conll2003\_Organization (1.9)\\ Person\_Instrument (6.7)\\ Choose\_First\_Of\_3 (9.0)\end{tabular}           \\
\midrule
Alphabetically\_First\_5  & \begin{tabular}[c]{@{}l@{}}Choose the word that comes \\ first in alphabetical order \\ from a list of 5 words\end{tabular}            & \begin{tabular}[c]{@{}l@{}}Alphabetically\_First\_5 (97.6)\\ Alphabetically\_First\_3 (96.7)\\ Adjective\_V\_Verb\_3 (31.0)\end{tabular} & \begin{tabular}[c]{@{}l@{}}Alphabetically\_Last\_5 (1.0)\\ Conll2003\_Organization (4.8)\\ Choose\_First\_Of\_3 (6.2)\end{tabular}      \\
\midrule
Alphabetically\_Last\_3   & \begin{tabular}[c]{@{}l@{}}Select the word that comes\\ last in alphabetical order\\ from a list of 3 words\end{tabular}               & \begin{tabular}[c]{@{}l@{}}Alphabetically\_Last\_5 (85.2)\\ Alphabetically\_Last\_3 (61.9)\\ Verb\_V\_Adjective\_5 (45.7)\end{tabular}   & \begin{tabular}[c]{@{}l@{}}Conll2003\_Organization (0.5)\\ Alphabetically\_First\_3 (0.5)\\ Alphabetically\_First\_5 (0.5)\end{tabular} \\
\midrule
Alphabetically\_Last\_5   & \begin{tabular}[c]{@{}l@{}}Choose the word that comes \\ last in alphabetical order \\ from a list of 5 words\end{tabular}             & \begin{tabular}[c]{@{}l@{}}Alphabetically\_Last\_5 (85.2)\\ Alphabetically\_Last\_3 (50.0)\\ Country\_Currency (33.8)\end{tabular}       & \begin{tabular}[c]{@{}l@{}}Alphabetically\_First\_3 (0.0)\\ Alphabetically\_First\_5 (0.0)\\ Choose\_First\_Of\_3 (3.3)\end{tabular}    \\
\midrule
Concept\_V\_Object\_3     & \begin{tabular}[c]{@{}l@{}}Select the concept from\\ a list of 3 words (1 abstract\\ concept, 2 concrete entities)\end{tabular}        & \begin{tabular}[c]{@{}l@{}}Concept\_V\_Object\_5 (99.5)\\ Concept\_V\_Object\_3 (98.6)\\ Animal\_V\_Object\_5 (83.3)\end{tabular}        & \begin{tabular}[c]{@{}l@{}}Conll2003\_Organization (0.0)\\ Person\_Instrument (10.5)\\ Choose\_First\_Of\_3 (11.0)\end{tabular}         \\
\midrule
Concept\_V\_Object\_5     & \begin{tabular}[c]{@{}l@{}}Select the concept from\\ a list of 5 words (1 abstract\\ concept, 4 concrete entities)\end{tabular}        & \begin{tabular}[c]{@{}l@{}}Concept\_V\_Object\_3 (98.6)\\ Concept\_V\_Object\_5 (97.6)\\ Animal\_V\_Object\_5 (82.4)\end{tabular}        & \begin{tabular}[c]{@{}l@{}}Capitalize\_Second\_Letter (4.3)\\ Choose\_First\_Of\_3 (9.0)\\ Word\_Length (10.0)\end{tabular}             \\
\midrule
Object\_V\_Concept\_3     & \begin{tabular}[c]{@{}l@{}}Select the concrete entity from\\ a list of 3 words (1 concrete\\ entity, 2 abstract concepts)\end{tabular} & \begin{tabular}[c]{@{}l@{}}Object\_V\_Concept\_3 (99.0)\\ Object\_V\_Concept\_5 (98.6)\\ Conll2003\_Location (86.7)\end{tabular}         & \begin{tabular}[c]{@{}l@{}}Choose\_First\_Of\_3 (2.9)\\ Conll2003\_Organization (3.8)\\ Person\_Instrument (9.0)\end{tabular}           \\
\midrule
Object\_V\_Concept\_5     & \begin{tabular}[c]{@{}l@{}}Select the concrete entity from\\ a list of 5 words (1 concrete\\ entity, 4 abstract concepts)\end{tabular} & \begin{tabular}[c]{@{}l@{}}Object\_V\_Concept\_5 (98.6)\\ Object\_V\_Concept\_3 (93.3)\\ Animal\_V\_Object\_5 (90.0)\end{tabular}        & \begin{tabular}[c]{@{}l@{}}Choose\_First\_Of\_3 (2.4)\\ Antonym (8.1)\\ Person\_Instrument (14.8)\end{tabular}                          \\
\midrule
English\_French           & \begin{tabular}[c]{@{}l@{}}Translate the given \\ English word into French\end{tabular}                                                & \begin{tabular}[c]{@{}l@{}}English\_German (86.2)\\ English\_French (85.6)\\ English\_Spanish (85.4)\end{tabular}                        & \begin{tabular}[c]{@{}l@{}}Conll2003\_Organization (0.0)\\ Animal\_V\_Object\_5 (23.6)\\ Choose\_First\_Of\_3 (25.6)\end{tabular}       \\
\midrule
English\_German           & \begin{tabular}[c]{@{}l@{}}Translate the given \\ English word into German\end{tabular}                                                & \begin{tabular}[c]{@{}l@{}}English\_Spanish (80.6)\\ English\_French (79.8)\\ English\_German (79.2)\end{tabular}                        & \begin{tabular}[c]{@{}l@{}}Conll2003\_Organization (0.0)\\ Choose\_First\_Of\_5 (13.5)\\ Choose\_First\_Of\_3 (17.2)\end{tabular}       \\
\midrule
English\_Spanish          & \begin{tabular}[c]{@{}l@{}}Translate the given \\ English word into Spanish\end{tabular}                                               & \begin{tabular}[c]{@{}l@{}}English\_Spanish (88.9)\\ English\_French (88.7)\\ English\_German (88.3)\end{tabular}                        & \begin{tabular}[c]{@{}l@{}}Conll2003\_Organization (0.0)\\ Choose\_First\_Of\_5 (16.1)\\ Choose\_First\_Of\_3 (25.5)\end{tabular}       \\
\midrule
Capitalize\_First\_Letter & \begin{tabular}[c]{@{}l@{}}Generate the first letter\\ of a given word \\ in captial form\end{tabular}                                 & \begin{tabular}[c]{@{}l@{}}Capitalize\_First\_Letter (100.0)\\ Lowercase\_First\_Letter (100.0)\\ Capitalize (99.4)\end{tabular}         & \begin{tabular}[c]{@{}l@{}}Choose\_Middle\_Of\_5 (0.0)\\ Prev\_Item (0.0)\\ Person\_Instrument (0.0)\end{tabular}                       \\
\midrule
Lowercase\_First\_Letter  & \begin{tabular}[c]{@{}l@{}}Generate the first letter\\ of a given word \\ in lowercase\end{tabular}                                    & \begin{tabular}[c]{@{}l@{}}Capitalize\_First\_Letter (100.0)\\ Lowercase\_First\_Letter (100.0)\\ Word\_Length (100.0)\end{tabular}      & \begin{tabular}[c]{@{}l@{}}Person\_Instrument (0.0)\\ Prev\_Item (0.0)\\ Next\_Item (0.0)\end{tabular}                                  \\
\midrule
Capitalize\_Last\_Letter  & \begin{tabular}[c]{@{}l@{}}Generate the last letter\\ of a given word \\ in captial form\end{tabular}                                  & \begin{tabular}[c]{@{}l@{}}Capitalize\_Last\_Letter (97.1)\\ Lowercase\_Last\_Letter (94.7)\\ Country\_Currency (52.6)\end{tabular}      & \begin{tabular}[c]{@{}l@{}}Choose\_Middle\_Of\_5 (0.0)\\ Conll2003\_Organization (0.0)\\ Antonym (0.0)\end{tabular}                     \\
\midrule
Lowercase\_Last\_Letter   & \begin{tabular}[c]{@{}l@{}}Generate the last letter\\ of a given word \\ in lowercase\end{tabular}                                     & \begin{tabular}[c]{@{}l@{}}Capitalize\_Last\_Letter (100.0)\\ Lowercase\_Last\_Letter (97.7)\\ Present\_Past (60.8)\end{tabular}         & \begin{tabular}[c]{@{}l@{}}Choose\_Middle\_Of\_5 (0.0)\\ Color\_V\_Animal\_5 (0.0)\\ Antonym (0.0)\end{tabular}                        
\\
\bottomrule
\end{tabular}}
\caption{\textbf{Inter-task activation patching analysis for 19 FV tasks using Llama-3.1-70B.} For each evaluation task, we report performance after patching high-$\alpha$ attention heads, where task embeddings are fixed to the evaluation task and head-selection parameters are derived from different head-selection tasks. Among the 57 FV tasks, we report the top-3 and bottom-3 head-selection tasks ranked by post-patching accuracy.}
\label{tab:Appendix_cross-task_accuracy_llama-3.1-70b}
\end{table*}
\clearpage

\section{Extended results on head-selection training dynamics}
\label{sec:Extended results on head-selection training dynamics}

\subsection{Extended results for all 57 FV tasks}
\label{subsec:training_dynamics-full results for all 57 tasks}
\begin{figure*}[h]
\begin{center}
    \includegraphics[width=0.99\linewidth]{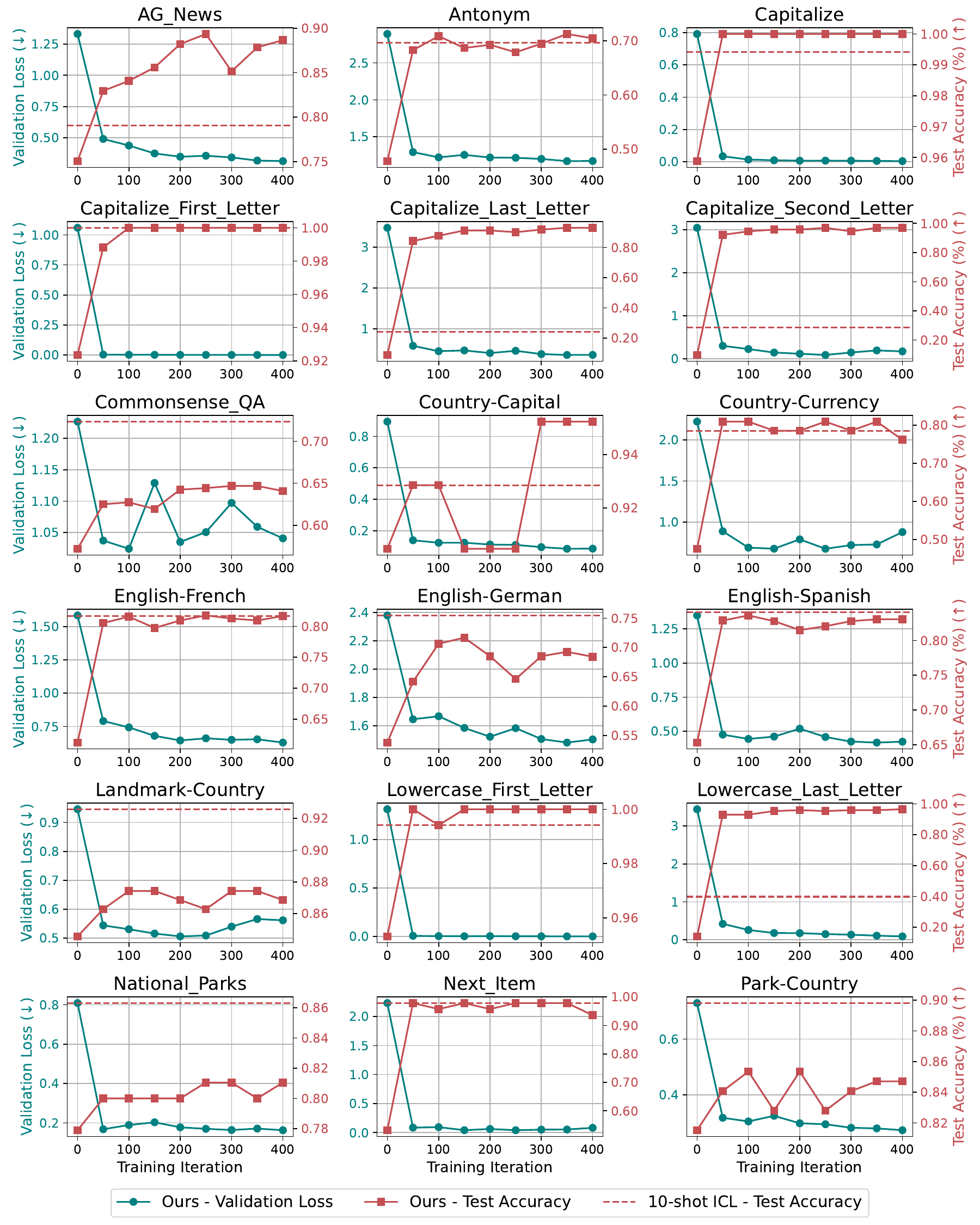}
\end{center}
\vspace{-2mm}
\caption{
\textbf{Training dynamics of soft head-selection parameters for 57 FV tasks~(Part 1 of 3).} 
Validation loss~(left y-axis) and test accuracy~(right y-axis) are plotted over 400 training iterations. Dashed lines indicate the 10-shot ICL accuracies for reference. The results are based on Llama-3.1-8B. Plots for the remaining tasks are provided in Figure~\ref{fig:analysis_training_dynamics_grid_2}-\ref{fig:analysis_training_dynamics_grid_3}.
}
\label{fig:analysis_training_dynamics_grid_1}
\end{figure*}

\begin{figure*}[t]
\begin{center}
    \includegraphics[width=0.99\linewidth]{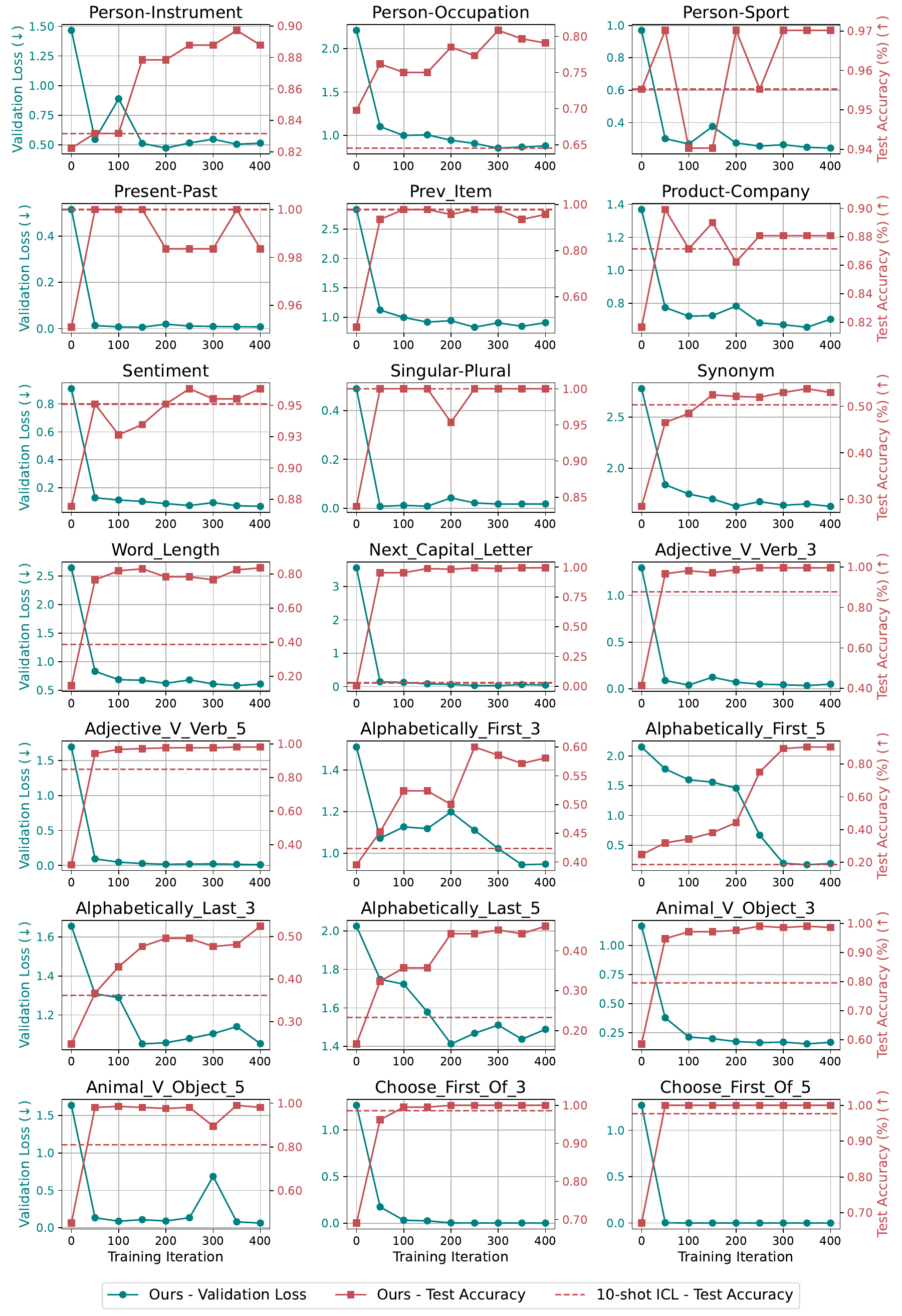}
\end{center}
\vspace{-2mm}
\caption{\textbf{Training dynamics of soft head-selection parameters for 57 FV tasks~(Part 2 of 3).} This figure continues from Figure~\ref{fig:analysis_training_dynamics_grid_1}. Validation loss~(left y-axis) and test accuracy~(right y-axis) are plotted over 400 training iterations. Dashed lines indicate the 10-shot ICL accuracies for reference. The results are based on Llama-3.1-8B. Plots for the remaining tasks are provided in Figure~\ref{fig:analysis_training_dynamics_grid_3}.
}
\label{fig:analysis_training_dynamics_grid_2}
\end{figure*}

\begin{figure*}[t]
\begin{center}
    \includegraphics[width=0.99\linewidth]{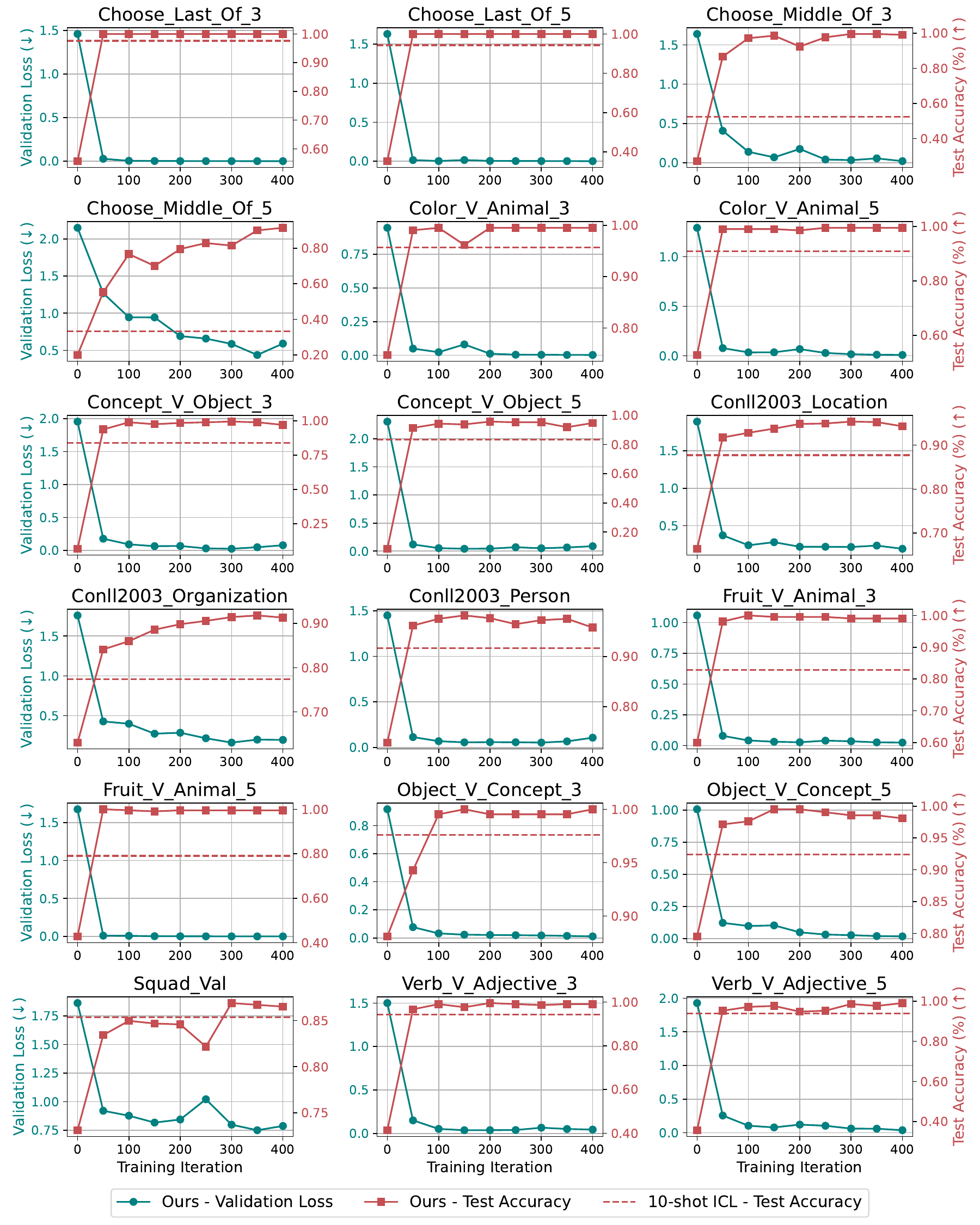}
\end{center}
\vspace{-2mm}
\caption{\textbf{Training dynamics of soft head-selection parameters for 57 FV tasks~(Part 3 of 3).} This figure concludes the series from Figures~\ref{fig:analysis_training_dynamics_grid_1}-\ref{fig:analysis_training_dynamics_grid_2}. Validation loss~(left y-axis) and test accuracy~(right y-axis) are plotted over 400 training iterations. Dashed lines indicate the 10-shot ICL accuracies for reference. The results are based on Llama-3.1-8B. 
}
\label{fig:analysis_training_dynamics_grid_3}
\end{figure*}

In this section, we present extended results corresponding to Figure~\ref{fig:analysis_training_dynamics} in Section~\ref{subsec:training dynamics of head selection}, showing the training dynamics in terms of validation loss and test accuracy for all 57 tasks from the FV benchmark using Llama-3.1-8B. The full set of results is provided in Figures~\ref{fig:analysis_training_dynamics_grid_1}-\ref{fig:analysis_training_dynamics_grid_3}.

\clearpage

\clearpage

\clearpage

\subsection{Results for larger language models}
\label{subsec:training dynamics-results with larger language models}
\begin{figure*}[h]
\begin{center}
    \includegraphics[width=0.82\linewidth]{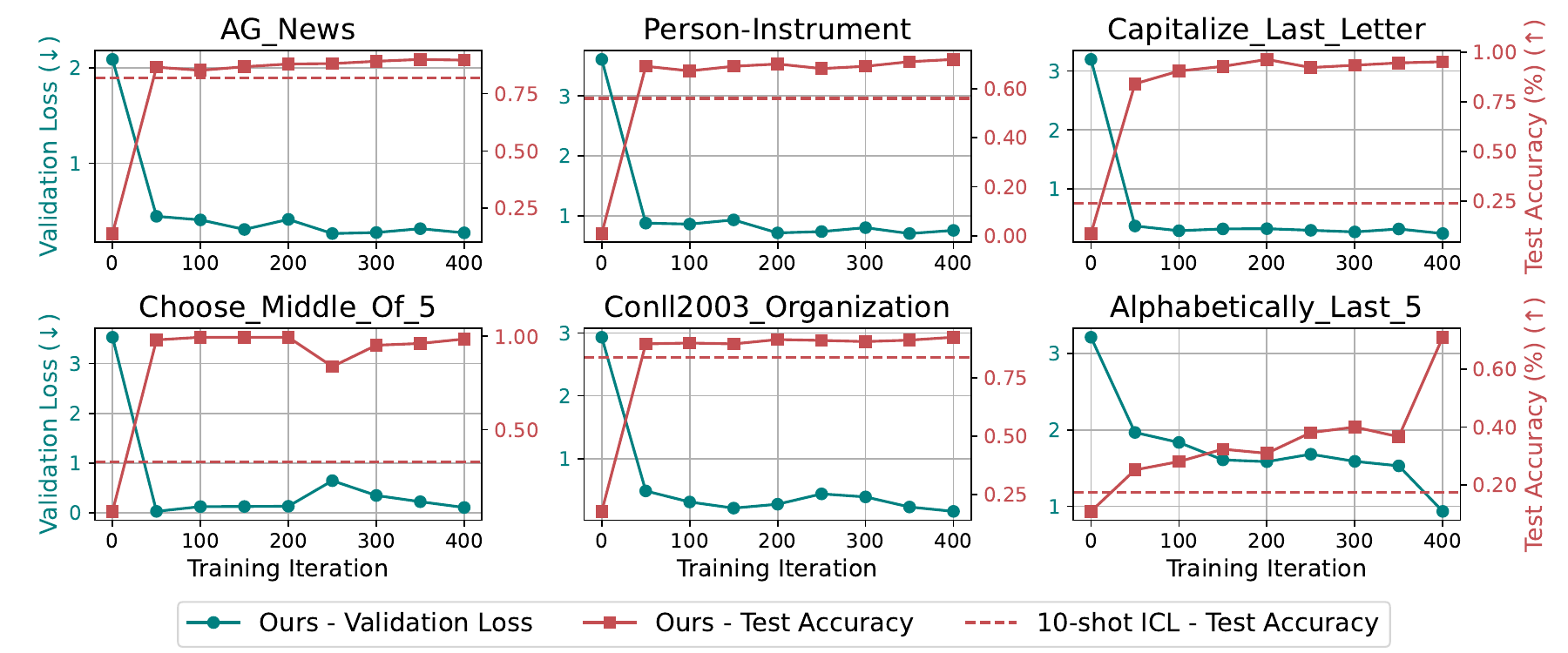}
\end{center}
\vspace{-2mm}
\caption{\textbf{Training dynamics of soft head-selection parameters for six FV tasks using Qwen3-32B.} Validation loss~(left y-axis) and test accuracy~(right y-axis) are plotted over 400 training iterations. Dashed lines indicate the 10-shot ICL accuracies for reference. 
}
\label{fig:analysis_training_dynamics_Qwen3-32B_combined_6}
\end{figure*}

\begin{figure*}[h]
\begin{center}
    \includegraphics[width=0.82\linewidth]{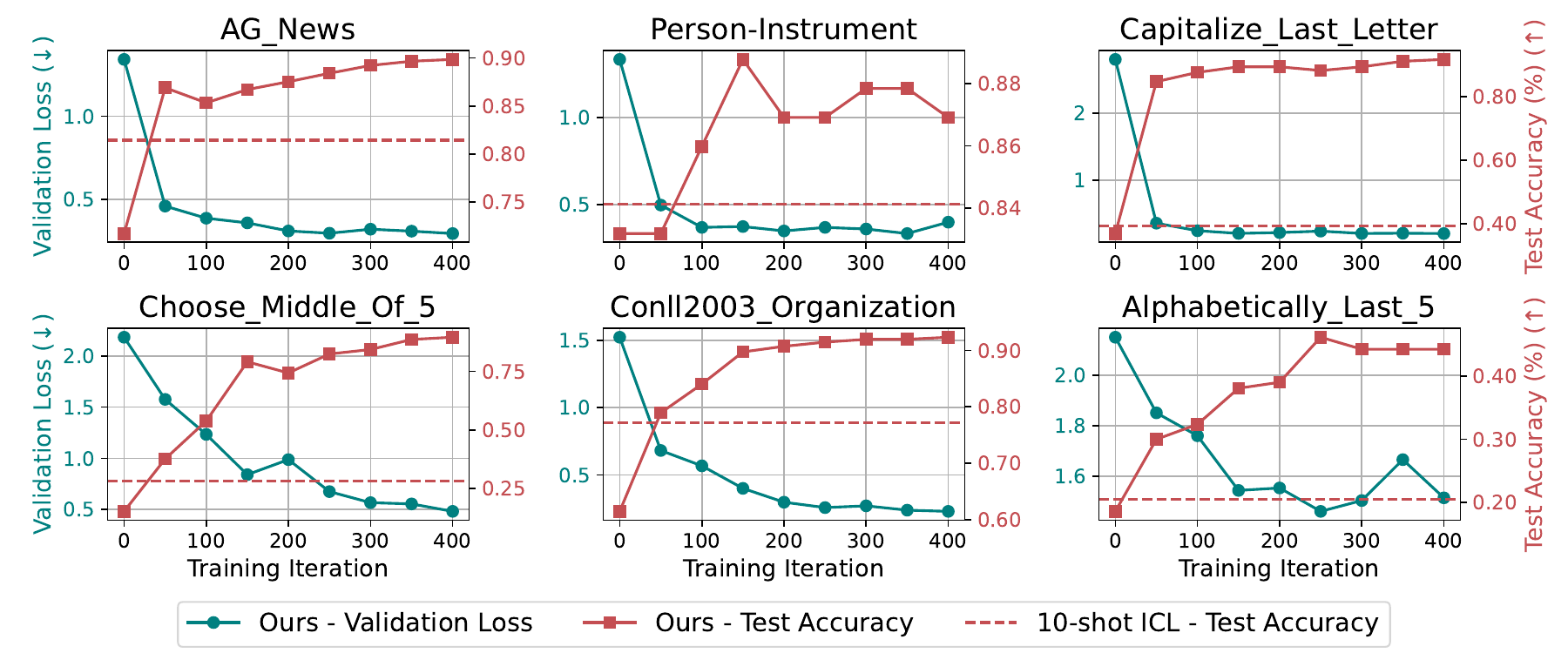}
\end{center}
\vspace{-2mm}
\caption{\textbf{Training dynamics of soft head-selection parameters for six FV tasks using Mixtral-8x7B-v0.1.} Validation loss~(left y-axis) and test accuracy~(right y-axis) are plotted over 400 training iterations. Dashed lines indicate the 10-shot ICL accuracies for reference. 
}
\label{fig:analysis_training_dynamics_Mixtral-8x7B-v0.1_combined_6}
\end{figure*}

\begin{figure*}[h]
\begin{center}
    \includegraphics[width=0.82\linewidth]{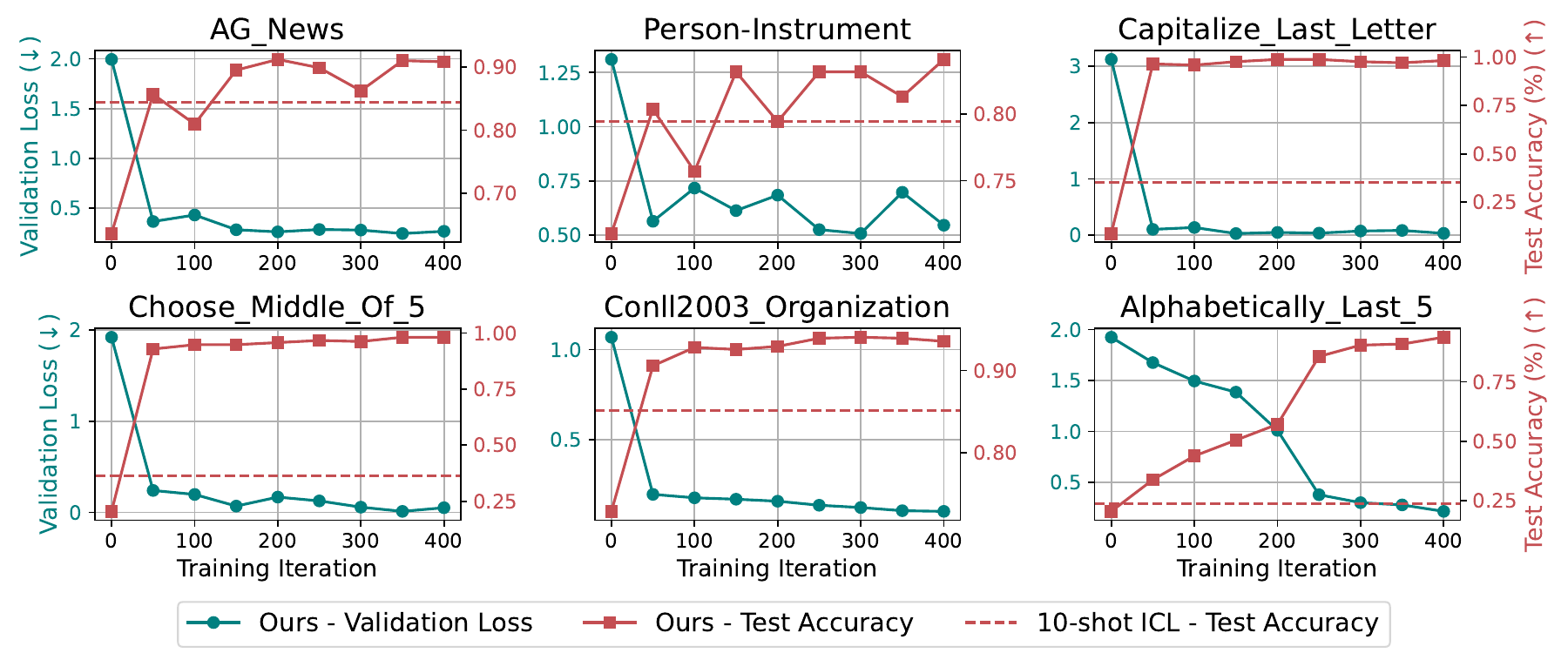}
\end{center}
\vspace{-2mm}
\caption{\textbf{Training dynamics of soft head-selection parameters for six FV tasks using Llama-3.1-70B.} Validation loss~(left y-axis) and test accuracy~(right y-axis) are plotted over 400 training iterations. Dashed lines indicate the 10-shot ICL accuracies for reference. 
}
\label{fig:analysis_training_dynamics_Llama-3.1-70B_combined_6}
\end{figure*}

Figures~\ref{fig:analysis_training_dynamics_Qwen3-32B_combined_6}-\ref{fig:analysis_training_dynamics_Llama-3.1-70B_combined_6} present training dynamics for the larger models Qwen3-32B, Mixtral-8x7B-v0.1, and Llama-3.1-70B across six selected tasks from the FV benchmark. The resulting plots exhibit consistent overall trends, similar to those observed with Llama-3.1-8B in Section~\ref{subsec:training dynamics of head selection}.

\clearpage

\begin{table*}[ht]
\centering
\setlength{\tabcolsep}{8pt}
\resizebox{0.95\textwidth}{!}{\begin{tabular}{l|cccc|c}
\toprule
\textbf{Task Embeddings in Stages 2--3} & \textbf{FV Benchmark} & \textbf{ANLI} & \textbf{MMLU-Pro} & \textbf{BBH} & \textbf{Average} \\
\midrule
Randomly Initialized Vectors & 82.90 & 40.78 & 34.39 & 50.46 & 52.13 \\
ICL-Derived Task Embeddings~(Default) & \textbf{90.02} & \textbf{47.31} & \textbf{38.78} & \textbf{58.04} & \textbf{58.54} \\
\bottomrule
\end{tabular}}
\caption{\textbf{Ablation on ICL-derived task embeddings.} For this ablation, we replace ICL-derived task embeddings with randomly initialized vectors in Stages 2--3 of our method, while keeping the same optimization procedure for the soft head-selection parameters $\mathbf{A}$. This random-vector variant achieves moderate performance but consistently underperforms the version using ICL-derived task embeddings across all four benchmarks.}
\label{tab:random_vs_icl_vector}
\end{table*}
\begin{table*}[ht]
\centering
\setlength{\tabcolsep}{6pt}
\resizebox{0.52\linewidth}{!}{\begin{tabular}{l|c}
\toprule
\textbf{Head Selection Scope} & \textbf{FV Benchmark} \\
\midrule
Ours~(First $1/3$ Layers) & 81.54 \\
Ours~(Middle $1/3$ Layers) & 84.87 \\
Ours~(Last $1/3$ Layers) & 44.43 \\
Ours~(Full Attention Heads; Default) & \textbf{90.02} \\
\bottomrule
\end{tabular}}
\caption{\textbf{Ablation on head selection across layers.} We restrict both the optimization of the soft head-selection parameters and the injection of task embeddings to different layer subsets. Using all attention heads yields the best performance, while restricting to partial layers degrades performance, with a particularly large performance drop when using only the last layers.}
\label{tab:layer_ablation}
\end{table*}


\begin{table*}[ht]
\centering
\setlength{\tabcolsep}{8pt}
\resizebox{0.94\textwidth}{!}{\begin{tabular}{l|ccc}
\toprule
\textbf{Method} & \textbf{\# Trainable Parameters~($\downarrow$)} & \textbf{Training Time~($\downarrow$)} & \textbf{Inference Time~(1000 queries)~($\downarrow$)} \\
\midrule
LoRA & 3407.87K & 6.2 min. & \textbf{16.5 min.} \\
Ours~($M=50$) & \textbf{1.02K} & \textbf{5.3 min.} & 16.7 min. \\
\bottomrule
\end{tabular}}
\caption{\textbf{Efficiency comparison with LoRA.} We report the number of trainable parameters, training time, and inference time for 1000 inference queries. For our method, \emph{Training Time} includes both Stage 1~(task embedding construction) and Stage 2~(optimization of soft head-selection parameters), which are performed once per task. All measurements are conducted on the AG\_News dataset using Llama-3.1-8B under the same hardware setting using a single NVIDIA A6000 GPU.}
\label{tab:efficiency_comparison_with_LoRA}
\end{table*}

\section{Additional analyses}
In this section, we present additional analyses, all conducted using Llama-3.1-8B.

\subsection{Ablation on ICL-derived task embeddings}
In Stage 2, our method optimizes the soft head-selection parameters $\mathbf{A}$ using ICL-derived task embeddings $\{ \mathbf{t}^{(l, h)} \}_{l=1, \dots, L;\, h=1, \dots, H}$ constructed in Stage 1~(see Figure~\ref{fig:method_overview} in Section~\ref{sec:method}). 
To isolate the contribution of these task embeddings, we replace them with randomly initialized vectors~(Xavier normal initialization) and optimize only $\mathbf{A}$ under the same training procedure. As shown in Table~\ref{tab:random_vs_icl_vector}, this random-vector variant achieves moderate performance but consistently underperforms the version using ICL-derived task embeddings across all four benchmarks.

One possible interpretation is that the random-vector variant acts as a lightweight parameter-efficient adaptation mechanism that injects trainable, bias-like signals into attention outputs, where only their scaling is learned via $\mathbf{A}$. In this sense, it resembles approaches such as BitFit~\citep{zaken2022bitfit}, which adapt models by tuning bias terms and typically achieve moderate but suboptimal performance compared to more expressive PEFT methods such as LoRA. This explains why the random-vector variant provides meaningful gains, yet remains inferior to the ICL-derived version. 

Together with the activation patching analyses in Section~\ref{sec:analysis}, these results highlight that both the ICL-derived task embeddings and the learned soft head-selection parameters are crucial for the effectiveness of our method.

\subsection{Ablation on head selection across layers}
Our method defines soft head-selection parameters $\mathbf{A} \in \mathbb{R}^{L \times H}$ over all attention heads, where $L$ and $H$ denote the number of attention layers and attention heads per layer, respectively. To analyze how task-relevant information is distributed across layers, we perform an ablation by restricting the definition of $\mathbf{A}$ and the injection of task embeddings to subsets of layers. Specifically, we partition the attention layers of Llama-3.1-8B into three groups: the first~(layers 0--9), middle~(10--20), and last~(21--31) thirds. For each group, we define and optimize $\mathbf{A}$ only over the selected layers and inject task embeddings into the corresponding attention heads, while keeping all other components unchanged.

Table~\ref{tab:layer_ablation} shows the results on the FV benchmark. Restricting to the first and middle thirds yields accuracies of $81.54\%$ and $84.87\%$, respectively, which remain competitive with using all attention heads~($90.02\%$). In contrast, using only the last third results in substantially lower performance of $44.43\%$. These results suggest that task-relevant signals are distributed across layers, as using all heads achieves the best performance. Notably, restricting to the last layers alone leads to a substantial performance drop, indicating that late-layer representations alone are insufficient to effectively support task embedding injection.

\subsection{Efficiency comparison with LoRA}
We compare the training and inference efficiency of our method with LoRA. The LoRA configuration used in this study is described in Section~\ref{subsec:implementation details for FV, MTV, and SITE}. As shown in Table~\ref{tab:efficiency_comparison_with_LoRA}, LoRA requires approximately 3.4M trainable parameters, whereas our method optimizes only the soft head-selection parameters, requiring only 1.02K trainable parameters. In terms of time cost, both methods exhibit comparable training and inference times.

While our method shows slightly lower performance than LoRA on complex reasoning benchmarks such as MMLU-Pro and BBH, it outperforms LoRA on the FV benchmark and ANLI while using substantially fewer trainable parameters and maintaining similar training and inference time. These results highlight the efficiency of ICL-driven embedding-based adaptation~(or ICL-driven activation steering) for task-specific adaptation, and suggest that improving its performance on complex reasoning tasks is a promising direction for future work.

\section{Use of LLMs in this work} 
We used chat-based LLMs solely for sentence-level editing to check grammar and improve clarity during paper writing. All edits were reviewed and verified by the authors. No scientific ideas, methods, analyses, or results were produced by LLMs; all conceptual contributions and experimental work are solely by the authors.

\end{document}